\documentclass[journal]{IEEEtran}
\usepackage{here}
\usepackage{graphicx}

\makeatletter
\let\MYcaption\@makecaption
\makeatother
\usepackage[font=footnotesize]{subcaption}
\makeatletter
\let\@makecaption\MYcaption
\makeatother

\usepackage{cite}
\usepackage{bm}
\usepackage{amsmath}
\usepackage{amssymb}
\usepackage{booktabs}
\usepackage{threeparttable}
\usepackage{bbm}
\usepackage{amsthm,amsmath,amssymb}
\usepackage{mathrsfs}
\usepackage{color}
\usepackage{siunitx}
\usepackage{multirow}
\usepackage{cases}
\usepackage{stfloats}
\usepackage{float}
\usepackage{url}

\DeclareMathOperator{\tr}{tr}
\DeclareMathOperator{\diag}{diag}
\DeclareMathOperator{\vectorize}{vec}

\IEEEoverridecommandlockouts                              
\title{A Complementary Framework for Human-Robot Collaboration with a Mixed AR-Haptic Interface}
\author{Xiangjie~Yan, Yongpeng~Jiang, Chen~Chen, Leiliang~Gong, Ming~Ge,~\IEEEmembership{Senior~Member,~IEEE}, Tao~Zhang,~\IEEEmembership{Senior~Member,~IEEE}, and Xiang~Li,~\IEEEmembership{Member,~IEEE}%
\thanks{X.~Yan, Y.~Jiang, C.~Chen, T.~Zhang, and X.~Li are with the Department of Automation, Tsinghua University, Beijing, China (e-mail: \protect\url{yanxj20@mails.tsinghua.edu.cn}; \protect\url{jyp19@mails.tsinghua.edu.cn}; \protect\url{chen-che20@mails.tsinghua.edu.cn}; \protect\url{taozhang@tsinghua.edu.cn}; \protect\url{xiangli@tsinghua.edu.cn}).}%
\thanks{L. Gong and M. Ge are with Hong Kong Productivity Council (HKPC), Hong Kong, China. (e-mail: \protect\url{leiliang@hkpc.org}; \protect\url{mingge@hkpc.org}.)}%
\thanks{This work was supported in part by the Science and Technology Innovation
2030-Key Project under grants 2021ZD0201404, in part by the National Natural Science
Foundation of China under grants U21A20517 and 52075290, in part by Guangdong-Hong Kong-Macao Innovation Center, Guangzhou through Special Foundation for Applied Research under grants ITPRD-2021-279, and in part by Beijing National Research Center for Information Science and Technology under grants 20201880382. Corresponding author: Xiang Li (xiangli@tsinghua.edu.cn)}}

\begin{document}
\markboth{IEEE TRANSACTIONS ON CONTROL SYSTEMS TECHNOLOGY}%
{Shell \MakeLowercase{\textit{et al.}}: A Sample Article Using IEEEtran.cls for IEEE Journals}


\maketitle



\begin{abstract}
There is invariably a trade-off between safety and efficiency for collaborative robots (cobots) in human-robot collaborations. Robots that interact minimally with humans can work with high speed and accuracy but cannot adapt to new tasks or respond to unforeseen changes, whereas robots that work closely with humans can but only by becoming passive to humans, meaning that their main tasks suspended and efficiency compromised. Accordingly, this paper proposes a new complementary framework for human-robot collaboration that balances the safety of humans and the efficiency of robots. In this framework, the robot carries out given tasks using a vision-based adaptive controller, and the human expert collaborates with the robot in the null space. Such a decoupling drives the robot to deal with existing issues in task space (e.g., uncalibrated camera, limited field of view) and in null space (e.g., joint limits) by itself while allowing the expert to adjust the configuration of the robot body to respond to unforeseen changes (e.g., sudden invasion, change of environment) without affecting the robot's main task. Additionally, the robot can simultaneously learn the expert's demonstration in task space and null space beforehand with dynamic movement primitives (DMP). Therefore, an expert's knowledge and a robot's capability are both explored and complementary. Human demonstration and involvement are enabled via a mixed interaction interface, i.e., augmented reality (AR) and haptic devices. The stability of the closed-loop system is rigorously proved with Lyapunov methods. Experimental results in various scenarios are presented to illustrate the performance of the proposed method.
\end{abstract}


\section{Introduction}
A cobot usually shares its workspace with humans, and thus directly (i.e., physically) or indirectly interacts with humans. It usually has two features: enhanced safety and ease of programming \cite{weiss_21}, so that it can operate near humans and be deployed flexibly on various tasks. The development of cobots in recent decades has increased production efficiency in manufacturing industries - which had previously reached a near-maximum level - to a new higher level.
\begin{figure}[!t]
\centering
\includegraphics[width=7cm]{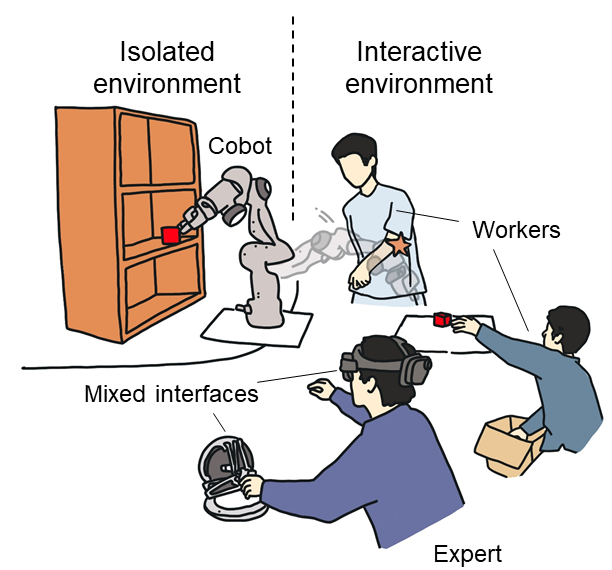}
\caption{An illustrative scenario for human-robot collaboration, where the robot needs to carry out tasks and transfer items in both an interactive environment and an isolated environment. 
An expert can also intervene or collaborate with the robot via the mixed AR-haptic interface.}\label{mixedScenario}
\end{figure}

The trade-off between safety and efficiency is always an open issue for cobots. To guarantee the safety of humans, cobots must typically suspend the ongoing task of end effector \cite{tcst16_li} and become passive to a human's control efforts, regardless of whether the human intervenes intentionally (i.e.,is an expert who wishes to lead the task) or unintentionally. Not until the human ceases intervening can the robot continue its task. This kind of operational process may affect task efficiency because the robot needs to transit between different working modes. 

To address the open issue, this paper proposes a new complementary framework for human-robot collaboration. Specifically, an illustrative scenario is considered in Fig.~\ref{mixedScenario}. First, a cobot grasps the target object in an interactive environment that also contains human workers. Then, it transfers the object to a desired position in an isolated setting, which excludes workers for reasons of safety or cleanliness. 
Such a scenario is commonly seen in factories, such as chemical factories \cite{hentout2019human}, food-processing factories \cite{iqbal2017prospects} and flat-panel-displays factories \cite{sanderson2019intelligent}. Cobots in these factories must often overcome one or several of the following challenges:
\begin{enumerate}
    \item [-] human workers invading their workspace;
    \item [-] an inexactly known environment, such that the relationship between their workspace and the sensory space is uncalibrated;
    \item [-] joint angles that are subject to several limits (e.g., singularity or constrained environment) and features that leave the field of view (FOV) during large displacements;
    \item [-] the unavailability of some exact task information (i.e., desired position or reference trajectory) until the task begins.
\end{enumerate}

This paper considers the aforementioned illustrative scenario and proposes a new framework for human-robot collaboration. The main novelty is its complementarity, which enables effective exploitation of the capabilities of a human expert (i.e., fast responses and smart decision-making) and those of a robot (i.e., high repetition and continuous working) and hence achieves a better balance between safety and efficiency. The contributions of this paper can be summarized as follows:
\begin{enumerate}
    \item [1)] For a co-existing environment, a new vision-based adaptive controller is proposed to ensure a robot's global stability 
    within the whole workspace; 
    a null-space damping model is also formulated to allow a human expert to get involved at any time without affecting the main task.
    \item [2)] For an isolated environment, a DMP-based planning scheme is developed to drive a robot to learn from expert demonstration and also migrate to new tasks; a mixed AR-haptic interface is also constructed such that a human expert can bi-manually demonstrate in both task space and redundant joint space.
    \item [3)] The stability of closed-loop system in both task space and null space is rigorously proved, with consideration of transition between multiple regional feedback. In addition, experimental results in different scenarios are presented to validate the performance of the proposed method.
\end{enumerate}

This work is an extension of our previous conference paper \cite{icra_22}, and the improvement includes:  
1) considering the whole pipeline of ``grasping $\rightarrow$ human interaction $\rightarrow$ transferring $\rightarrow$ obstacle avoidance $\rightarrow$ placement''; 2) dealing with the joint limits and limited FOV during large displacements and developing a model-free method for online estimation of image Jacobian matrix; 3) learning skills from expert demonstration of both redundant joint and robot end effector; 4) building a mixed interface for more illustrative and intuitive interactions between human and robot to better complement each other; 5) carrying out more ablation studies and real-world experiments.

\section{Related Works}
This section reviews related works on cobot control and learning.\\

\noindent\textbf{Task-Space Control}: 
Task-space control directly specifies a feature or goal in task space, e.g., Cartesian space or vision space. This eliminates the need to solve an inverse kinematic problem, and thus task-space control has now become a standard method applied to robot manipulators. When a robot working in task space is subjected to a large displacement, its global stability is commonly limited by several open issues, i.e., joint limits, limited FOV, and uncalibrated camera. 

First, a robot's joint angles may be subjected to several limitations due to singular configurations and constraints limits. Many studies have developed methods to keep the robot away from these limits, e.g., by replanning the trajectory beforehand \cite{handbook},  exploring the kinematic redundancy \cite{tro16_Gosselin}, or damping the robot's motion when it is near the limit \cite{ijrr17_Carmichael}. 

Second, the problem of limited FOV occurs when the visual feature leaves it during the task. In \cite{tro04_chesi}, a switching approach was proposed to switch the control input between a backward motion outside the FOV and a visual servoing method within it. In \cite{tro05_garcia-aracil}, a new weighted feature was proposed for vision-based control to allow some features to leave the FOV during manipulation. In \cite{tro11_gans}, multiple visual features were kept within the FOV by regulating both the mean and also the variance of multiple features. The visual servoing scheme in \cite{tro19_Bechlioulis} set the FOV as visibility constraints in the predefined performance bound, and the robot was controlled to achieve the desired transient response and hence stay within the FOV. 

Third, the parameters of a camera deployed in a task-space control system may be unknown, for lacking prior calibration or being subject to adjustment (e.g., changes in focal length) when undertaking different tasks.  
In \cite{li2022hybrid}, Li {\em et al.} proposed a series of adaptive laws to  estimate the parameters of uncalibrated cameras and robot dynamics concurrently. Without estimating the unknown camera parameters, 
\cite{liang2020purely} used three feature points distributed in a particular pattern, such that only pixel feedback from a fixed uncalibrated camera was needed to perform stabilization control for a nonholonomic mobile robot.

To address the aforementioned issues together, \cite{automatica13_li} proposed an adaptive task-space controller with the feedback switching among joint space, Cartesian space and vision space, to achieve the global stability within the whole workspace. Nevertheless, this and other existing task-space control schemes are commonly applicable to isolated environments, or their global stability is affected by the issues of joint limits, limited FOV, or uncalibrated sensors (e.g., \cite{ijrr17_Carmichael}). \\

\noindent\textbf{Human-Robot Collaboration}: The scenario of human-robot collaboration (HRC) can be found in some manufacturing applications, where humans and robots perform a task together \cite{weiss_21,Hjorth_22,Inkulu_21,simoes_22}. 

As the robot co-exists with human, it is very important to guarantee the safety. 
The safety standard for HRC systems is defined in ISO 10218-1 and ISO/TS 15066, 
which are now used as guidelines for many real-world applications. In \cite{Kanazawa_21}, an objective-switching method was adopted in an assembly task, which balanced the safety and time efficiency when the robot was approaching and avoiding the co-workers respectively. In \cite{Palleschi_21}, an optimization-based trajectory planning framework with iterative online safety module was proposed for HRC.  
A model recovering human-exerted forces was developed for dyadic cooperative object manipulation in \cite{noohi_16}, so that only human-applied force was measured to control the robot while the safety was guaranteed. 
However, most of the existing works have assumed that the perception of humans or obstacles is fully reliable, lacking the ability to deal with suddenly appearing or unforeseen changes.

Various HRC interfaces have also been developed for human involvement.
In \cite{wang_18}, EMG and IMU sensors were adopted to assess the human motion intention during physical interaction. Face and gesture recognition were integrated in a collaborative system for assembly tasks \cite{makrini_17}. Moreover, some fluency evaluation methods were proposed in \cite{hoffman_19}. Nevertheless, the aforementioned works are commonly limited to specific and predefined tasks. 
A general HRC interface in industrial setting is the teach pendant; 
However, it was found that this kind of interface has 
decreased the efficiency of and experience bained by humans \cite{bogue_16}. Therefore, there is a demand for a human-oriented, intuitive, and general interface for HRC that facilitates convenient human involvement in robot-assisted tasks.\\

\noindent\textbf{DMP for Robot Learning}: 
Among various techniques of learning-from-demonstration (LfD), DMP has been proven to be an effective and efficient approach
\cite{argall2009survey,ravichandar2020recent,saveriano2021dynamic}. In DMP, movement is modelled using a spring-damper system, with the addition of a nonlinear forcing term to encode and modulate learning skills \cite{argall2009survey}. The DMP method has also been extended to Cartesian orientation \cite{ude2014orientation}, force adaptation \cite{gams2014coupling}, and arbitrary via-point adaptation \cite{zhou2019learning} variations. 

A typical LfD setting is to construct a teleoperation system, such that skills from the human expert (the master side) can be transferred to the robot (the slave side) and then adjusted via DMP according to the given task. As the master side is usually mechanically different from the slave side, several issues on how to record and reproduce the demonstration for better execution performance were raised \cite{argall2009survey,ravichandar2020recent}. To better teach the robot via DMP, \cite{beik2020model} developed a simulation environment where the human expert demonstrates a task using an AR device, and then transfers the demonstrated skills to the robot using highly transparent feedback.  
In another approach \cite{peternel2018robotic}, a teleoperation control interface was developed for bilateral teleoperation, which consists of a three degrees-of-freedom (DOFs) HapticMaster robot and a stiffness control handle, which allows human-in-the-loop teaching and hence results in a better trajectory encoding.

In summary, DMP methods which mainly or solely focus on the task of the robot end effector have been developed in existing works. However, for a complex task such as that illustrated in Fig.~\ref{mixedScenario}, it is necessary to regulate both the robot end effector and the robot's body shape to suit the position of the collaborating human expert, avoid collisions, and so on.

\section{Preliminaries}
This paper considers a cobot with redundant joints, 
whose forward kinematic model can be described as
\begin{equation}
\bm r=\bm h(\bm q),
\end{equation}
where $\bm r\hspace{-0.05cm}\in\hspace{-0.05cm}\Re^6$ denotes the position and the orientation of the robot end effector in Cartesian space, $\bm q\hspace{-0.05cm}\in\hspace{-0.05cm}\Re^n$ is the vector of joint angles, $n>6$ is the number of DOFs, and $\bm h(\cdot)\hspace{-0.05cm}\in\hspace{-0.05cm}\Re^n\rightarrow \Re^6$ is a nonlinear function. 


Then, the velocity of the robot end effector in Cartesian space is related to the joint-space velocity as follows \cite{sensoryfeedbackbook}:
\begin{equation}
\dot{\bm r}=\bm J(\bm q)\dot{\bm q},
\end{equation}
where $\bm J(\bm q)\hspace{-0.05cm}\in\hspace{-0.05cm}\Re^{6\times n}$ is the Jacobian matrix from joint space to Cartesian space.

The pseudo-inverse matrix is defined as
\begin{equation}
\bm J^+(\bm q)\triangleq\bm J^T(\bm q)(\bm J(\bm q)\bm J^T(\bm q))^{-1}\in\Re^{n\times 6}
\end{equation}
such that $\bm J(\bm q)\bm J^+(\bm q)=\bm I_6$, where $\bm I_6\in \Re^{6\times6}$ is an identity matrix. 
Consequently, the null-space matrix can be introduced as follows \cite{tro14_sadeghian}:
\begin{equation}
\bm N(\bm q)\triangleq\bm I_n-\bm J^+(\bm q)\bm J(\bm q)\in\Re^{n\times n},\label{nulldefine}
\end{equation}
where $\bm I_n\hspace{-0.05cm}\in\hspace{-0.05cm}\Re^{n\times n}$ represents an identity matrix. Equation (\ref{nulldefine}) means that $\bm J(\bm q)\bm N(\bm q)\hspace{-0.05cm}=\hspace{-0.05cm}\bm 0$, $\bm N(\bm q)\bm J^+(\bm q)\hspace{-0.05cm}=\hspace{-0.05cm}\bm 0$, and $\bm N^2(\bm q)\hspace{-0.05cm}=\hspace{-0.05cm}\bm N(\bm q)$, which implies that the null-space matrix $\bm N(\bm q)$ is orthogonal to the Jacobian matrix $\bm J(\bm q)$.

Because cobots are typically lightweight and operate at relatively low speeds, the control input can be specified at kinematic level, such that
\begin{equation}
\dot{\bm q}=\bm u,\label{controlInput}
\end{equation}
where $\bm u\hspace{-0.05cm}\in\hspace{-0.05cm}\Re^n$ denotes the control input corresponding to the joint-space velocity.

When a camera is used to measure the robot end effector in vision space, the feature's velocity in vision space is related to the end effector's velocity in Cartesian space \cite{slotine}, i.e.,
\begin{equation}
\dot{\bm x}=\bm J_s(\bm r)\dot{\bm r},\label{cameraModel}
\end{equation}
where $\bm x$ denotes the feature's position (which is the position of the robot end effector) in vision space, and $\bm J_s(\bm r)\hspace{-0.05cm}\in\hspace{-0.05cm}\Re^{2\times 6}$ is the image Jacobian matrix.  
Due to the limited FOV, the visual feature is not available when it is initially outside the FOV or when it temporarily leaves the FOV during manipulation. 

\begin{center}
\begin{figure}[!htb]
\centering
\includegraphics[width=2.2in]{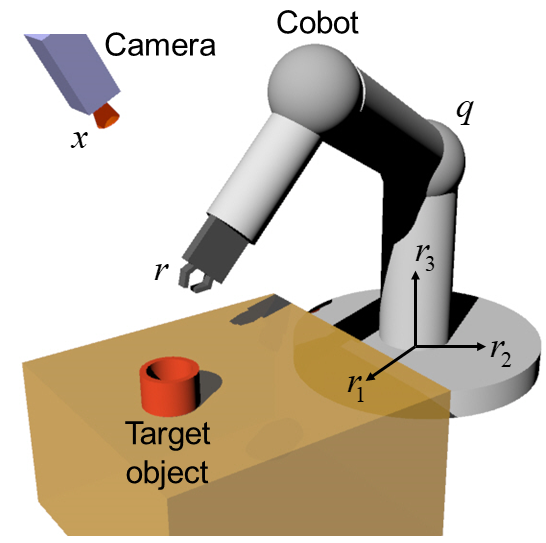}
\caption{A vision-based robot manipulator is controlled to grasp a target object, where $\bm r\in\Re^6$ denotes the position and orientation of the robot end effector in Cartesian space, $\bm q\in\Re^n$ is a vector of joint angles, and $\bm x\in\Re^2$ represents the feature's position in vision space.}\label{cobotFig}
\end{figure}
\end{center}

If the camera is not calibrated beforehand or if its parameters are adjusted to suit new tasks (e.g., camera autofocus, depth variation), the exact knowledge of the camera parameters may not be available, and hence the image Jacobian matrix is also unknown and is denoted as $\hat{\bm J}_s(\bm r)$. An example of the vision-based cobot is illustrated in Fig.~\ref{cobotFig}.\\
\vspace{-0.2cm}
\begin{center}
\begin{figure*}[!t]
\centering
\hspace{-0.3cm}
\includegraphics[width=4.5in]{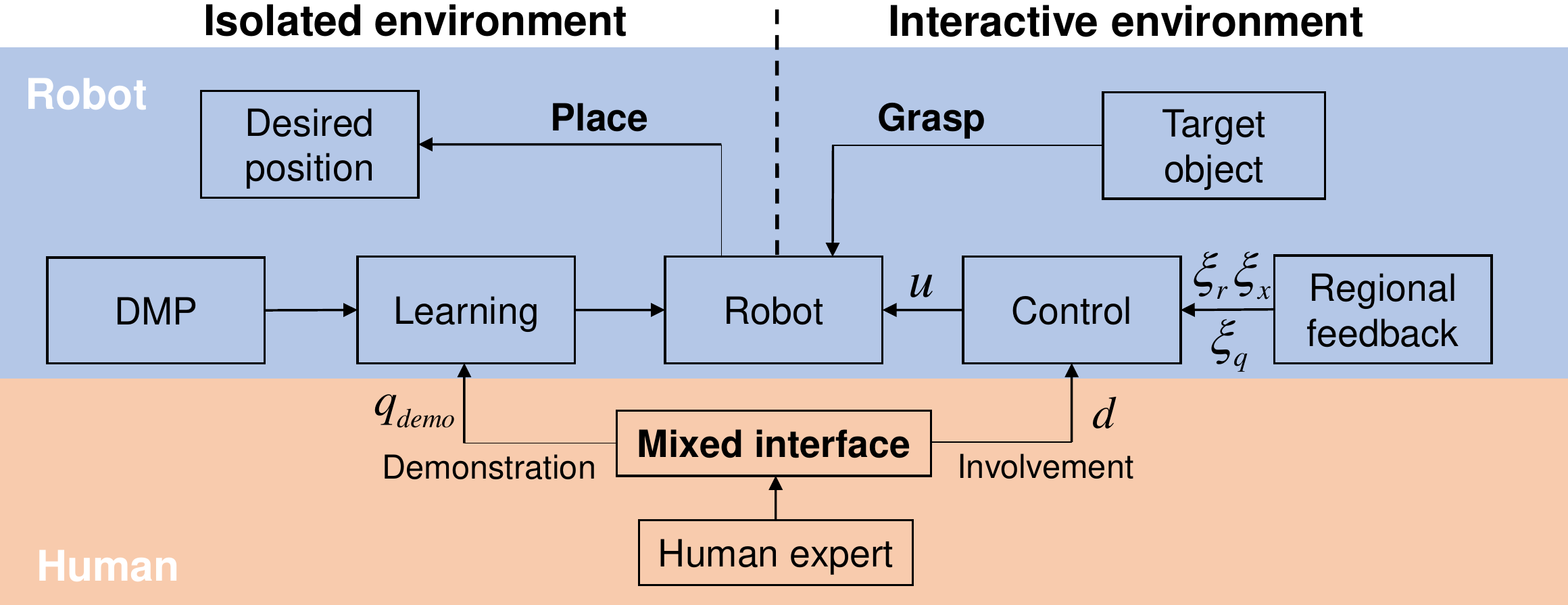}
\vspace{-0.2cm}
\caption{The overall structure of complementary collaboration framework, where $\bm u$ is the control input, $\bm\xi_q, \bm\xi_r, \bm\xi_x$ are regional feedback vectors, $\bm d$ denotes the human control efforts, and $\bm q_{demo}$ is the demonstration trajectory in joint space. The robot carries out the task of ``grasping - placing'', which involves the transition from an interactive environment (where the target object is located) to an isolated environment (where the desired position is located). The human expert can become involved to deal with unforeseen changes that occur during the grasping task, or to demonstrate a movement, which enables his/her expertise to be transferred to the robot before the placement task. Thus the ability of human and robot are complementary. The involvement and demonstration are achieved via a mixed interface, which is detailed in Section VII.}\label{weight}
\end{figure*}
\end{center}
\vspace{-0.5cm}

\noindent\textbf{Problem Formulation}:
\emph{The aim of this study is to design the control input (\ref{controlInput}) to guarantee the global stability of the robot and the convergence of task-space error to zero, in the presence of joint limits, uncalibrated camera, limited FOV and human involvement.}


\section{Multiple Regional Feedback}
This paper considers the scenario illustrated in Fig.~\ref{mixedScenario}. The overall structure of complementary collaboration under such a scenario is shown in Fig.~\ref{weight}. 
That is, the human expert interacts with the robot via the mixed interface. When the robot is controlled to grasp the target object in the co-existing environment, the human expert exerts control efforts in the null space to avoid potential collisions with workers. When the robot learns to place the object at the desired position, the human expert simultaneously demonstrates the reference trajectory in both Cartesian space and null space. Note that the proposed structure can also be extended to many other scenarios involving human-robot interaction.

This section presents the regional feedback \cite{automatica13_li} for the grasping operation, which is used to solve problems (e.g., joint limits and limited FOV) that may arise during large-displacement transfers. Thus, a series of regional feedbacks is formulated for the whole workspace, and 
the combination of regional feedback ensures the performance of robot in a global sense.
\\

\noindent\textbf{Joint-Space Feedback}: 
The joint-space feedback is exploited to keep the robot away from the
joint limits, which exist due to singularity or constrained workspace. 
Given that there are $m$ limited configurations, the region function enclosing the $i^{th}$ ($i\hspace{-0.05cm}=\hspace{-0.05cm}1, 2, \cdots, m$)  configuration is specified as
\begin{equation}
f_{i}(\bm q)\leq0, \label{psx2}
\end{equation}
and the robot is away from this configuration when $f_{i}(\bm q)\hspace{-0.1cm}>\hspace{-0.1cm}0$. 
For example, the joint-space regions for a 2-DOF planar robot \cite{sensoryfeedbackbook} can be specified as: $f_{1}(\bm q)\hspace{-0.1cm}=\hspace{-0.1cm}q_2^2\hspace{-0.05cm}-\hspace{-0.05cm}R_1^2\hspace{-0.05cm}\leq\hspace{-0.05cm}0$ and $f_{2}(\bm q)\hspace{-0.05cm}=\hspace{-0.05cm}R_2^2\hspace{-0.05cm}-\hspace{-0.05cm}(q_2\hspace{-0.05cm}-\hspace{-0.05cm}\pi)^2\hspace{-0.05cm}\leq\hspace{-0.05cm} 0$, where $R_1, R_2$ specify the region size, and $q_2=0, q_2=\pi$ are singular configurations.

Then, the potential energy function for the joint-space regions is proposed as
\begin{equation}
P_s(\bm q) = \sum_{i=1}^m \left\{\frac{k_{qi}}{2}[\min(0,f_{i}(\bm q))]^2 + \frac{k_{ri}}{2}[\min(0,f_{ri}(\bm q))]^2\right\}, \label{jointxp}
\end{equation}
where $k_{qi}$ and $k_{ri}$ are positive constants, and $f_{ri}(\bm q)\hspace{-0.05cm}\leq\hspace{-0.05cm}0$ is a reference region enclosing $f_{i}(\bm q)\hspace{-0.05cm}\leq\hspace{-0.05cm}0$.

\begin{center}
\begin{figure}[!h]
\centering
\includegraphics[width=3in]{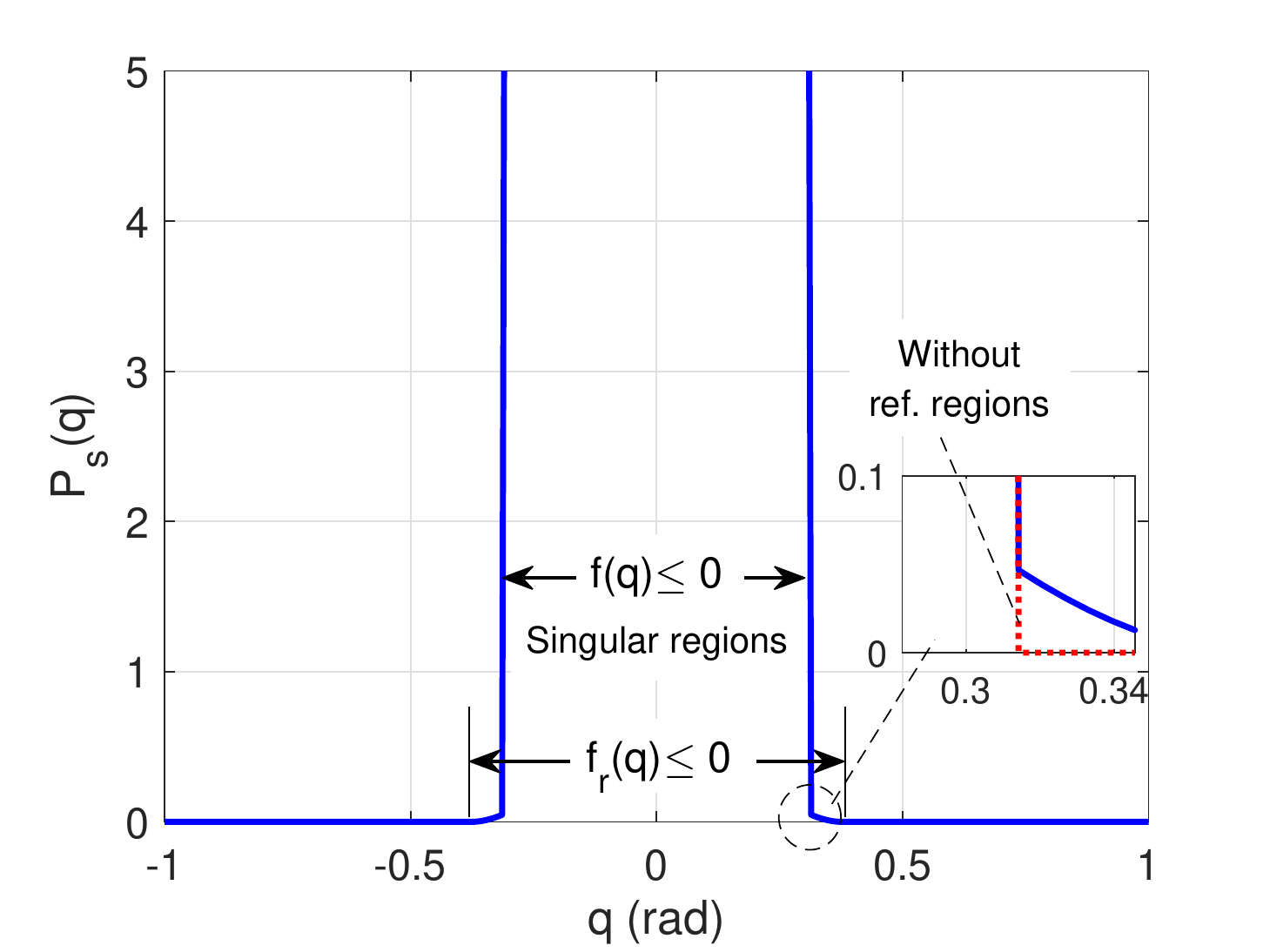}
\vspace{-0.2cm}
\caption{An illustration of the potential energy function $P_s(\bm q)$. If there is no reference region, the steep gradient of the potential energy may cause oscillatory movement of the robot (red dashed line).}\label{singular}
\end{figure}
\end{center}

An illustration of the potential energy function is shown in Fig.~\ref{singular}. 
The first term in (\ref{jointxp}) is to create a high potential energy barrier, such that the robot does not have enough kinematic energy to approach the limited configurations. Hence, $k_{qi}$ is set large to make the gradient of $P_s(\bm q)$ steep (see Fig.~\ref{singular}). However, the steep gradient would cause oscillatory movement of the robot if it is very close to the region boundary $f_{i}(\bm q)\hspace{-0.05cm}=\hspace{-0.05cm}0$. Hence, the second term in (\ref{jointxp}) is to decelerate the robot in advance and hence to alleviate the potential oscillation, where $k_{ri}$ is relatively small.

Now, a regional feedback vector can be specified in joint space as
\begin{equation}
\bm \xi_q\triangleq\frac{\partial P_s(\bm q)}{\partial \bm q},
\end{equation}
which can be treated as a repulsive force to keep the robot away from the limited configurations. The vector automatically reduces to zero when the robot is outside the joint-space region.
\vspace{-0.5cm}
\begin{figure}[!ht]
  \centering
  \begin{subfigure}[b]{.443\linewidth}
    \centering
    \includegraphics[width=\linewidth]{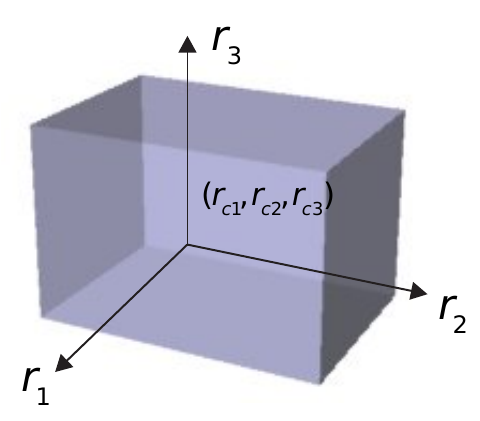}
    \caption{}
    \label{CaRe} 
  \end{subfigure}
  \begin{subfigure}[b]{.543\linewidth}
    \centering
    \includegraphics[width=\linewidth]{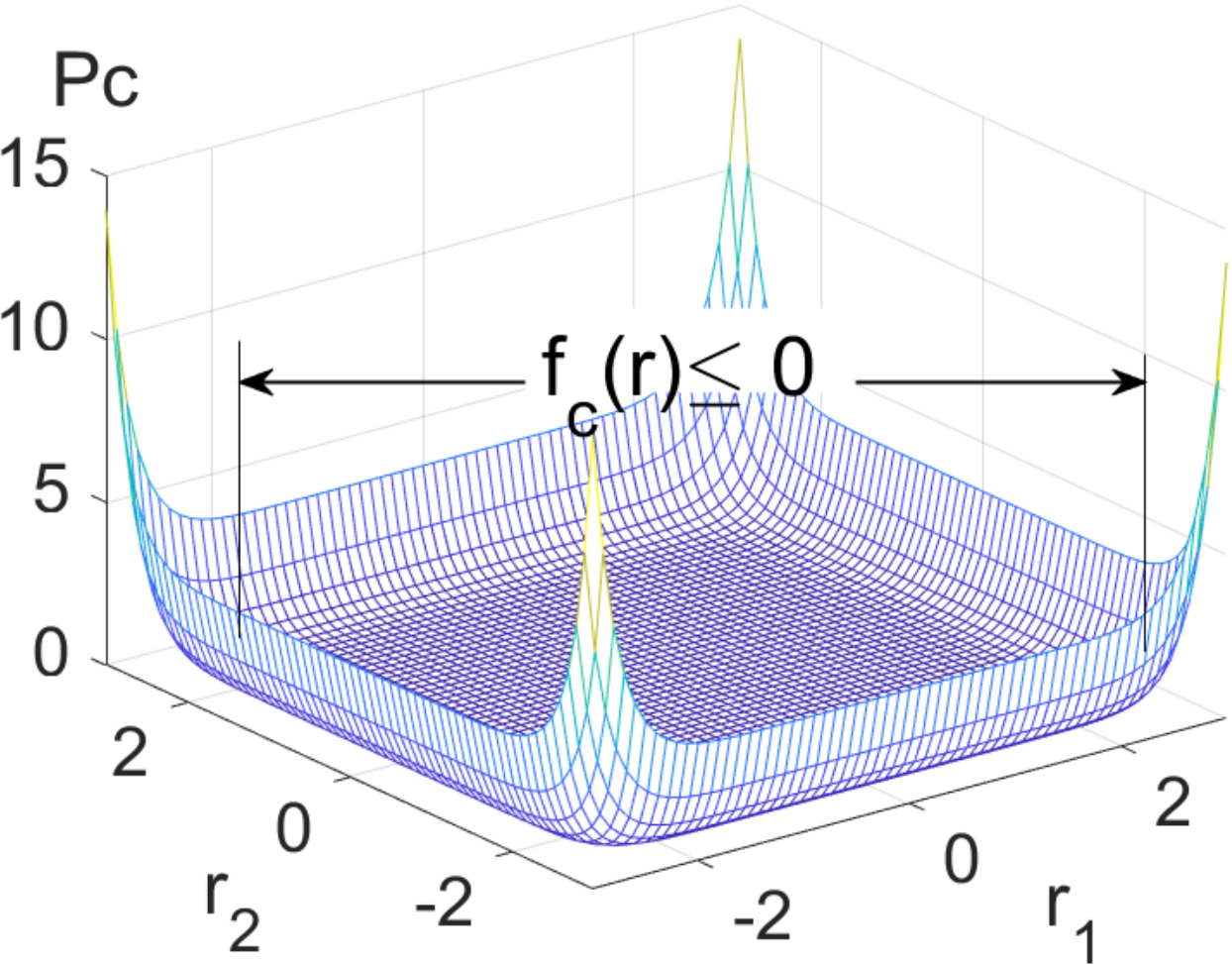}
    \caption{}
    \label{CaPe} 
  \end{subfigure}
  \begin{subfigure}[b]{.443\linewidth}
    \centering
    \includegraphics[width=\linewidth]{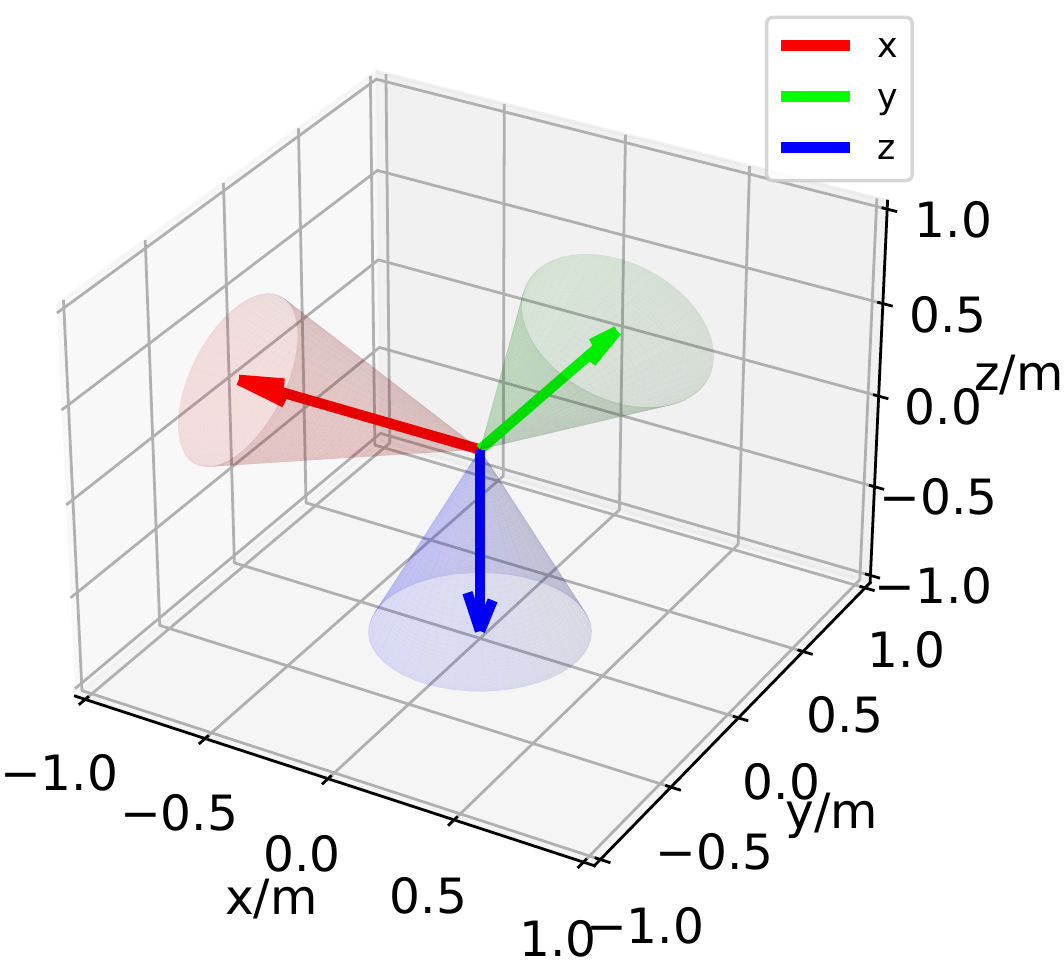}
    \caption{}
    \label{CaOrtRe} 
  \end{subfigure}
  \begin{subfigure}[b]{.543\linewidth}
    \centering
    \includegraphics[width=3.2cm]{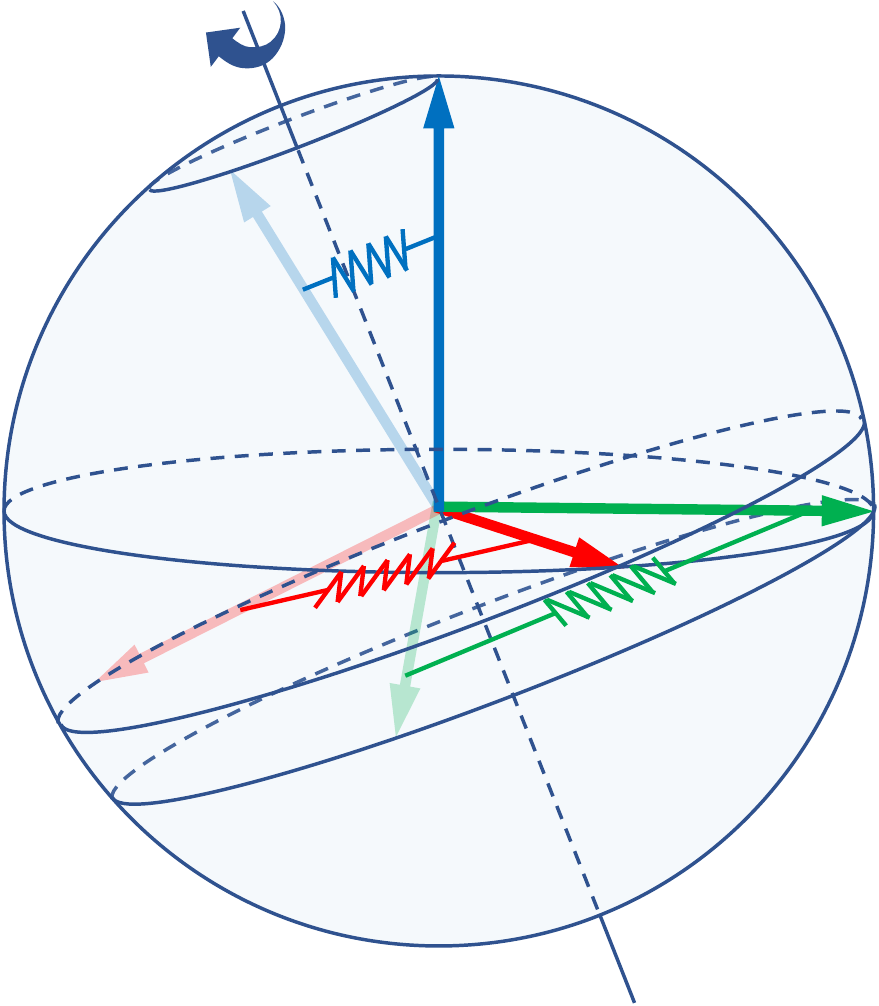}
    \caption{}
    \label{CaOriPo} 
  \end{subfigure}
  \caption{An illustration of the Cartesian-space region and its potential energy. (a) The Cartesian-space position region is formulated as a rectangular block; (b) The potential energy function in 2D space; (c) The Cartesian-space orientation region is formulated as a ``cone''; (d) The potential energy corresponding to the Cartesian orientation region function is intuitively regarded as being stored in the virtual spring system, which is only activated outside the region.}
  \label{exp4ss} 
\end{figure}

\noindent\textbf{Cartesian-Space Feedback}:
When the feature is not within the FOV, the Cartesian-space feedback is employed to drive the robot to move towards the feature, such that the feature can be seen inside the FOV. To match the rectangular FOV, a region is formulated in Cartesian space as
\begin{equation}
\bm f_c(\bm r)= \begin{bmatrix}
f_{c1}(r_1)\\
f_{c2}(r_2)\\
f_{c3}(r_3)
\end{bmatrix}
= \begin{bmatrix}
(\frac{r_1-r_{c1}}{c_1})^2-1\\
(\frac{r_2-r_{c2}}{c_2})^2-1\\
(\frac{r_3-r_{c3}}{c_3})^2-1
\end{bmatrix}
\leq\bm 0,\label{CaFun}
\end{equation}
where $\bm r_c\hspace{-0.1cm}=\hspace{-0.1cm}[r_{c1}, r_{c2}, r_{c3}]^T\hspace{-0.1cm}\in\hspace{-0.1cm}\Re^3$ denotes a reference position in the Cartesian-space position region (also within the FOV when it is projected to the vision space), and $c_1, c_2, c_3$ are positive constants. The region $\bm f_c(\bm r)\hspace{-0.1cm}\leq\hspace{-0.1cm}\bm 0$ describes a cube in Cartesian space (see Fig.~\ref{CaRe}), which matches the FOV in vision space. Note that the region is only used to regulate the position of robot end effector, such that the end effector is visible after it enters the Cartesian-space region.  

Then, the corresponding potential energy function for the above region is proposed as
\begin{equation}
P_t(\bm r)\hspace{-0.05cm}=\hspace{-0.05cm}\sum\limits_{i=1}^{3}\left\{\frac{k_{ci}}{2}[\max(0,f_{ci}(\bm r))]^2\right\},\label{CaPeFun}
\end{equation}
where $k_{ci}$ are positive constants. An illustration of the potential energy function is shown in Fig.~\ref{CaPe}. From (\ref{CaPeFun}) and Fig.~\ref{CaPe}, it can be seen that the potential energy drives the robot end effector to enter the region where $\bm f_c(\bm r)\hspace{-0.1cm}\leq\hspace{-0.1cm}\bm 0$ (which is also inside the FOV) and then reduces to zero.

However, the pose of the robot end effector may not be suitable for grasping if only the position is regulated. To address the problem, another Cartesian-space region is introduced to control the orientation of robot end effector, i.e., 
\begin{equation}
f_o(\bm r)\hspace{-0.05cm}=\hspace{-0.05cm}{\alpha}_o \lVert \log{(\bm p*\bm p_g^{-1})} \rVert_2-1\leq 0, \label{OrtRegion}
\end{equation}
where ${\alpha}_o$ is a positive constant which is related to the tolerance of orientation error, $\bm p_g$ and $\bm p$ are the quaternions representing the goal and the robot end effector, respectively, $(*)$ denotes the Hamilton product, and $\log()$ describes the quaternion logarithm.
The use of quaternions avoids the representation singularity. A simple example for the orientation region (\ref{OrtRegion}) can be given as: 
$\lVert \log(\bm p*\bm p_g^{-1}) \rVert_2$, which describes the distance between $\bm p$ and $\bm p_g$.

Similarly, the corresponding potential energy function is formulated as
\begin{equation}
P_o(\bm r)\hspace{-0.05cm}=\hspace{-0.05cm}\frac{1}{2} k_o[\max(0,f_o(\bm r))]^2,\label{OrtFun1}
\end{equation}
where $k_o$ is a positive scaling factor. The overall potential energy function in Cartesian space is the sum of $P_t(\bm r)$ and $P_o(\bm r)$, i.e.,
\begin{equation}
P_c(\bm r)\hspace{-0.05cm}=\hspace{-0.05cm}P_t(\bm r)+P_o(\bm r).
\end{equation}

Next, the regional feedback vector is specified in Cartesian space as
\begin{equation}
\bm \xi_r\triangleq\frac{\partial P_c(\bm r)}{\partial \bm r}=\frac{\partial P_t(\bm r)}{\partial \bm r}+\frac{\partial P_o(\bm r)}{\partial \bm r},\label{kesi_r_origin}
\end{equation}
which can be treated as an attractive force that drives the robot end effector to enter the Cartesian-space regions, such that the end effector becomes visible and its orientation can be adjusted to a configuration suitable for grasping. The derivative of the potential energy function $\frac{\partial P_o(\bm r)}{\partial \bm r}$ 
depends on whether analytical Jacobian or geometric Jacobian is applied in (\ref{closed}), which is detailed in the appendix. Note that the choice of Jacobian in the experiment had a slight impact on the control performance.

\noindent\textbf{Vision Feedback}: 
The vision feedback is employed by the robot end effector to grasp the target object. First, a region function is specified in vision space as
\begin{equation}
f_v(\bm x)=\left(\frac{x_1-x_{d1}}{b_1}\right)^2 + \left(\frac{x_2-x_{d2}}{b_2}\right)^2-1\leq 0,\label{eq_vision_re_func}
\end{equation}
where $b_1, b_2>0$ are constants representing the half size of the FOV in the coordinates of $x_1$ and $x_2$, respectively, and $\bm x_d\hspace{-0.05cm}=\hspace{-0.05cm}[x_{d1}, x_{d2}]\hspace{-0.05cm}\in\hspace{-0.05cm}\Re^2$ is the desired position, which is also the position of the target object in vision space.

\begin{center}
\begin{figure}[!h]
\centering
\includegraphics[width=2in]{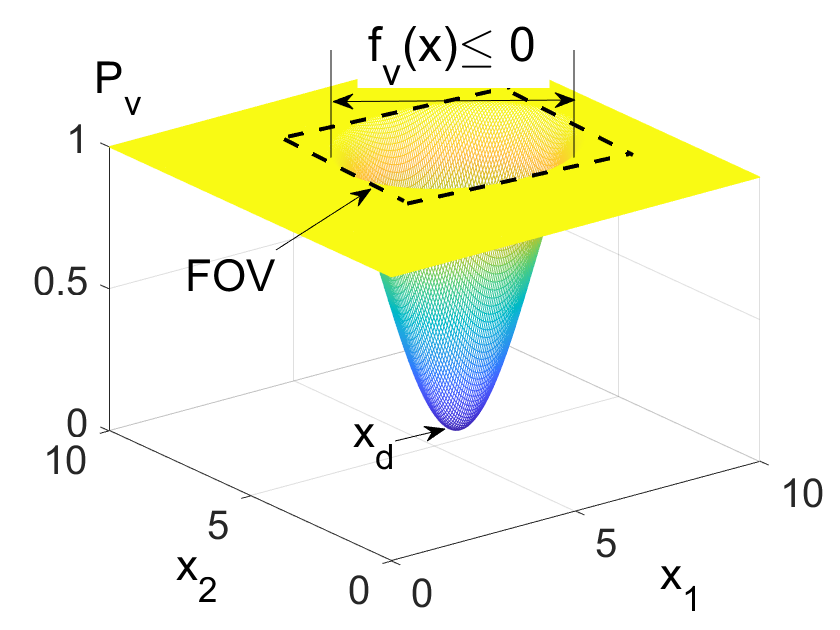}
\vspace{-0.2cm}
\caption{An illustration of the potential energy function $P_v(\bm x)$. Note that the vision-space region $f_v(\bm x)\leq 0$ is inside the FOV.}\label{visionPE}
\end{figure}
\end{center}

Accordingly, the potential energy function in vision space is introduced as
\begin{equation}
P_v(\bm x)=\frac{k_v}{2}\{1-[\min(0, f_v(\bm x))]^2\},\label{eq_vision_p}
\end{equation}
where $k_v$ is a positive constant. The potential energy is shown in Fig.~\ref{visionPE}; its gradient is zero outside the region (i.e., $f_v(\bm x)>0$) and non-zero inside the region, which drives the robot end effector to converge to the desired position for grasping.
\vspace{-0.2cm}
\vspace{-0.5cm}

The regional feedback vector in vision space can now be specified in a similar way as
\begin{equation}
\bm\xi_x\triangleq\frac{\partial P_v(\bm x)}{\partial \bm x},
\end{equation}
which is activated inside the vision-space region where $f_v(\bm x)\leq 0$ (which is also inside the FOV). To ensure that the robot end effector can move from the Cartesian-space region to the vision-space region, the Cartesian-space region can be set smaller than the corresponding FOV, such that there is overlapping between each other.

\section{Vision-Based Global Adaptive Control}
This section presents the global adaptive controller with multiple regional feedback, which drives the robot to interact with humans and grasp the target object in the presence of joint limits, limited FOV and uncalibrated cameras. 
Specifically, the control input is proposed as
\begin{align}
\bm u &= -\bm J^+(\bm q)(\hat{\bm J}_s^T(\bm r)\bm\xi_x+\bm \xi_r) +\bm N(\bm q)c_d^{-1}(\bm d - \bm \xi_q),\label{globalControl}
\end{align}
where $c_d$ is a positive scalar and $\bm d\in\Re^n$ denotes the control efforts exerted by the human expert on robot joints via mixed interfaces (e.g., Microsoft HoloLens 2 in \cite{hololens2}). The first term on the right side of (\ref{globalControl}) is to drive the end effector to carry out the main task in vision space, and the second term is regulate the redundant joints in null space to collaborate with the expert and also avoid joint limits, without affecting the main task.
The objective of null-space control term can also be described as a damping model, i.e., 
\begin{equation}
\bm N(\bm q)(c_d\dot{\bm q})=\bm N(\bm q)(\bm d - \bm \xi_q). \label{nullspace_model}    \end{equation}
where $c_d$ can be considered as the desired damping parameter.

\begin{center}
\begin{figure}[!tb]
\centering
\includegraphics[width=6cm]{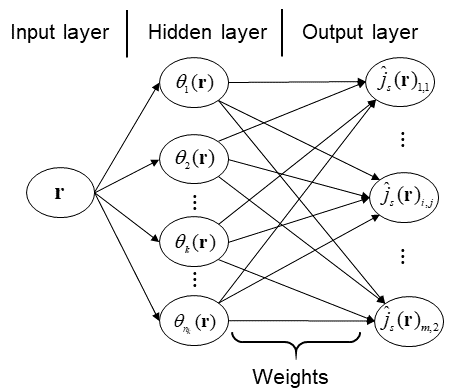}
\vspace{-0.2cm}
\caption{The structure of the radial basis function neural network, where the input layer receives information on the position and orientation of the robot end effector (i.e., the vector of $\bm r$), the hidden layer consists of a series of RBF neurons (i.e., $\theta_1(\bm r), \theta_2(\bm r), \cdots$), and the output layer generates the Jacobian matrix $\bm J_s(\bm r)$. }\label{fig:NN}
\end{figure}
\end{center}

Next, the entry of the unknown image Jacobian transpose is approximated with adaptive neural network (NN) as
\begin{equation}
\hat{j}_s(\bm r)_{i,j}=\hat{\bm w}_{i,j}^T\bm\theta(\bm r),\label{approxNN}
\end{equation}
where $\hat{j}_s(\bm r)_{i,j}$ is the $(i,j)th$ entry of the matrix $\hat{\bm J}_s^T(\bm r)$, $i=1,2,\cdots,m,j = 1,2.$ $\hat{\bm w}_{i,j}\hspace{-0.1cm}\in\hspace{-0.1cm}\Re^{n_k} $ is the corresponding weight, and $\bm\theta(\bm r):\Re^{m}	\rightarrow \Re^{n_k}$ is the nonlinear function of neurons. Radial basis function (RBF) is utilized as the neuron, where the $i$th entry is
\begin{equation}
\theta_{i}(\bm r)=\exp\biggl( -\frac{1}{2 \sigma_{i}^{2}} \lVert\bm r-\bm c_{i} \rVert_2^{2} \biggr),\label{eq_rbf}
\end{equation}
where $\bm c_{i}$ and $\sigma_{i}^2,i=1,\cdots,n_k$ are the centers and the variances, respectively. These parameters are manually predefined. The structure of the NN is shown in Fig.~\ref{fig:NN}.

For simplicity, we rewrite (\ref{approxNN}) in the following vectorized form
\begin{equation}
\vectorize(\hat{\bm J}_{s}^T(\bm r)) = \hat{\bm W}\bm\theta(\bm r),\label{approxNN_vec}
\end{equation}
where $\hat{\bm W}=[\hat{\bm w}^T_{1,1};\cdots;\hat{\bm w}^T_{m,1};\hat{\bm w}^T_{1,2};\cdots;\hat{\bm w}^T_{m,2}]\hspace{-0.1cm}\in\hspace{-0.1cm}\Re^{2m\times n_k}$.


Thus, the weight of NN is updated with the following online adaptation law:
\begin{equation}
\dot{\hat{\bm W}}=-\Bigl[\bm L\bm\theta(\bm r)
(\hat{\bm J}_s^T(\bm r)\bm\xi_x+\bm \xi_r)^T\bm\xi_x^{\prime} \Bigr]^T,\label{update}
\end{equation}
where $\bm L\in\Re^{n_k\times n_k}$ is a positive-definite matrix, and $\bm\xi_{x}^{\prime}$ is a matrix that reformulates the entries of $\bm\xi_x = [\xi_{x1},\xi_{x2}]^T$ as
\begin{equation}
    \bm\xi_{x}^{\prime}=[\xi_{x1}\bm I_{m},\xi_{x2}\bm I_{m}]\in\Re^{m\times2m}.
\end{equation}
which has the following property:
\begin{equation}
    \bm\xi_{x}^{\prime} \vectorize(\hat{\bm J}_{s}^T(\bm r)) = \hat{\bm J}_{s}^T(\bm r)\bm\xi_{x}. \label{reformulation}
\end{equation}

\vspace{-0.2cm}
\vspace{-0.7cm}

The advantages of the proposed control scheme (\ref{globalControl}) are summarized as follows.
\begin{enumerate}
\item[-] When the robot nears the joint limits, the regional feedback vector $\bm \xi_q$ is activated to drive the robot away. 

\item[-] The regional feedback vector $\bm \xi_r$ is used to drive the robot end effector to approach the desired position, such that both the feature and the desired position can be seen by the camera.

\item[-] The regional feedback vector $\bm \xi_x$ is activated only when both the feature and the target object are visible, such that the robot can grasp the target object in the presence of uncalibrated cameras.

\item[-] The online adaptation is driven by the regional feedback to deal with the unknown parameters concurrently.
\end{enumerate}

By substituting (\ref{globalControl}) into (\ref{controlInput}), the closed-loop equation is obtained as
\begin{equation}
\dot{\bm q} = -\bm J^+(\bm q)(\hat{\bm J}_s^T(\bm r)\bm\xi_x+\bm \xi_r)+\bm N(\bm q)c_d^{-1}(\bm d - \bm \xi_q).\label{closed}
\end{equation}
We are now in a position to state the following theorem.\\

\noindent \textbf{Theorem}: {\em When the proposed vision-based adaptive control scheme described by (\ref{globalControl}) and (\ref{update}) is applied to a collaborative robotic system, both the global stability of the closed-loop system and the convergence of task errors to zero are guaranteed, in the presence of joint limits, limited FOV, uncalibrated cameras and human's interaction.}\\

\noindent \textbf{Proof}: See the appendix.

\section{Robot Learning from Demonstration}
After the robot grasps the object, it transfers the object to the desired position in an isolated environment, as illustrated in Fig.~\ref{weight}. 
The desired position and the trajectory to this position are learnt from human's demonstration via the DMP approach. Such a proposed formulation has the following advantages.
\begin{enumerate}
    \item[1)] Learning from demonstration can effectively exploit the expert's knowledge to set the desired position (e.g., a specific grasping pose) and suit the constrained space (e.g., path planning in the existence of the cabinet in Fig.~\ref{mixedScenario});
    \item[2)] DMP allows the learnt trajectory to be conveniently adjusted in response to the task generalization (e.g., modified goal positions on other shelves of the cabinet, or modified reaching speeds).
\end{enumerate}


Basically, the DMP model for learning the trajectory of a single joint can be described as
\begin{align}
\tau^{2} \ddot{q} &= \alpha_{q}\left[\beta_{q}(g-q)-\tau \dot{q}\right]+ \zeta (z), \label{eq_tf} \\
\tau \dot{z} &= -\alpha_{z} z, \label{eq_cs}
\end{align}
where (\ref{eq_tf}) specifies a transformation system; (\ref{eq_cs}) describes a canonical system; $\tau$ is a positive time constant, $q$, $\ddot{q}$, $\ddot{q}$ represent the angle, angular velocity and acceleration of the joint, respectively; $\alpha_{q}, \beta_{q}, \alpha_z$ are positive gain constants. In addition, $\zeta (z)$ is the forcing term, which is formulated as a linear combination of nonlinear basis functions, i.e., 
\begin{equation}
\zeta (z)=\frac{\sum_{i=1}^{N} \psi_{i}(z) \omega_{i}}{\sum_{i=1}^{N} \psi_{i}(z)} z\left(g-q_{0}\right),
\end{equation}
where $\omega_{i}$ is the weight of the $i$th basis function, $N$ is the total number of basis functions, $q_{0}$ is the initial position at $t=0$, and $\psi_{i}(z)$ is the $i$th basis function. The latter is chosen as a radial basis function to allow discrete movement, i.e., 
\begin{equation}
\psi_{i}(z)=\exp\biggl[-\frac{1}{2 \sigma_{i}^{2}}\left(z-c_{i}\right)^{2}\biggr],
\end{equation}
where $c_{i}$ and $\sigma_{i}^2$ denote the center and the variance, respectively. The generation of a learnt trajectory consists of two steps: a \emph{learning phase} and a \emph{reproducing phase}. 

\noindent\textbf{Learning Phase}: First, the human expert demonstrates a trajectory in terms of
$\{q_{\text{demo}}(t),\dot{q}_{\text{demo}}(t), \ddot{q}_{\text{demo}}(t)\}_{t=0}^\mathbb{T}$. By referring to this demonstration, a desired forcing term can be calculated by transposing (\ref{eq_tf}) as
\begin{equation}
\zeta_{d}=\tau^{2} \ddot{q}_{\text{demo}}-\alpha_{q}\left[\beta_{q}\left(g-q_{\text{demo}}\right)-\tau \dot{q}_{\text{demo}}\right].
\end{equation}

Next, the locally weighted quadratic error is defined as the optimization target, i.e., 
\begin{equation}
\text{Cost}_{i}=\sum_{t=1}^{\mathbb{T}} \psi_{i}(z(t))\left[\zeta_d(t)-\omega_{i} z(t)(g-q_0)\right]^{2},
\end{equation}
which forms a standard weighted linear regression problem, with the solution as
\begin{equation}
\omega_{i} = \frac{\bm s^T\Gamma_i \bm \zeta_d}{\bm s^T\Gamma_i \bm s},
\end{equation}
where $i=1,\cdots, N$, and 
\begin{align}
\bm s &= [z(1)(g-q_0), z(2)(g-q_0), \cdots, z(\mathbb{T})(g-q_0)]^T,\\
\Gamma_i &= \diag(\psi_i(1), \psi_i(2), \cdots, \psi_i(\mathbb{T})),\\
\bm \zeta_d &= [\zeta_d(1), \zeta_d(2), \cdots, \zeta_d(\mathbb{T})]^T.
\end{align}

\noindent\textbf{Reproducing Phase}: 
After the weights in the forcing term are learnt, a new trajectory can now be generated by running (\ref{eq_tf}) and (\ref{eq_cs}). 
The learnt trajectory can be modulated according to the given scenario. Specifically, 
$\tau$ can be adjusted to speed up or slow down the trajectory execution, and $g$ can be changed to set a new goal position while maintaining a similar transient movement to that position.

In this paper, the human expert demonstrates the trajectory in both the task space of the robot end effector and the null space of it in a bimanual and intuitive way via the mixed interface, as illustrated in Fig.~\ref{mixedScenario} and Fig.~\ref{weight}. The demonstration teaches the robot to not only find the correct desired pose in task space but also shape the overall body to avoid collisions in a constrained environment.


\section{Experiment}
\begin{figure} [!htb]
    \begin{subfigure}{.495\linewidth}
        \centering
        \includegraphics[width=\linewidth]{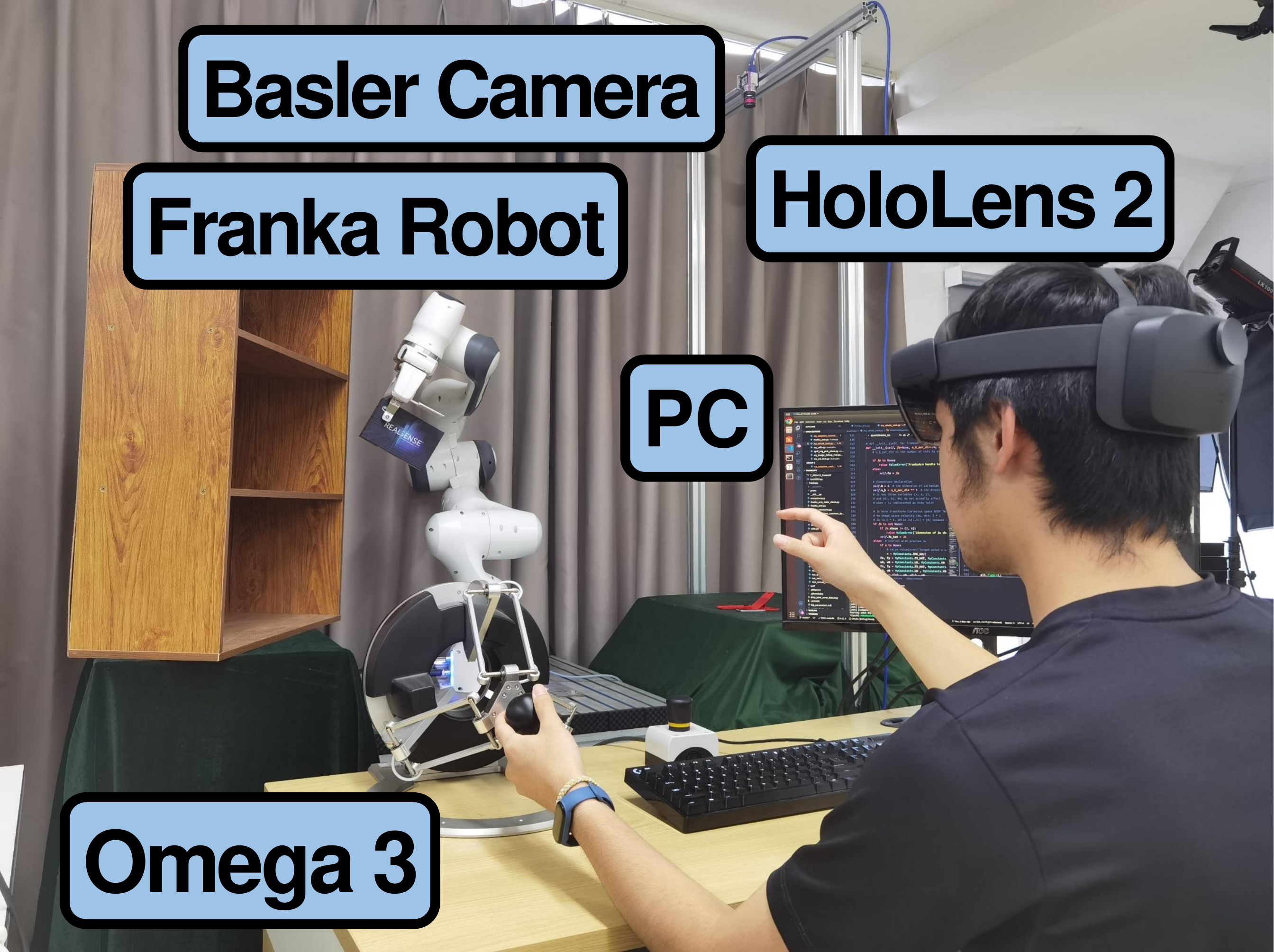}
        \caption{}\label{fig_system}
    \end{subfigure}
    \hfill
    \begin{subfigure}{.495\linewidth}
        \centering
        \includegraphics[width=\linewidth]{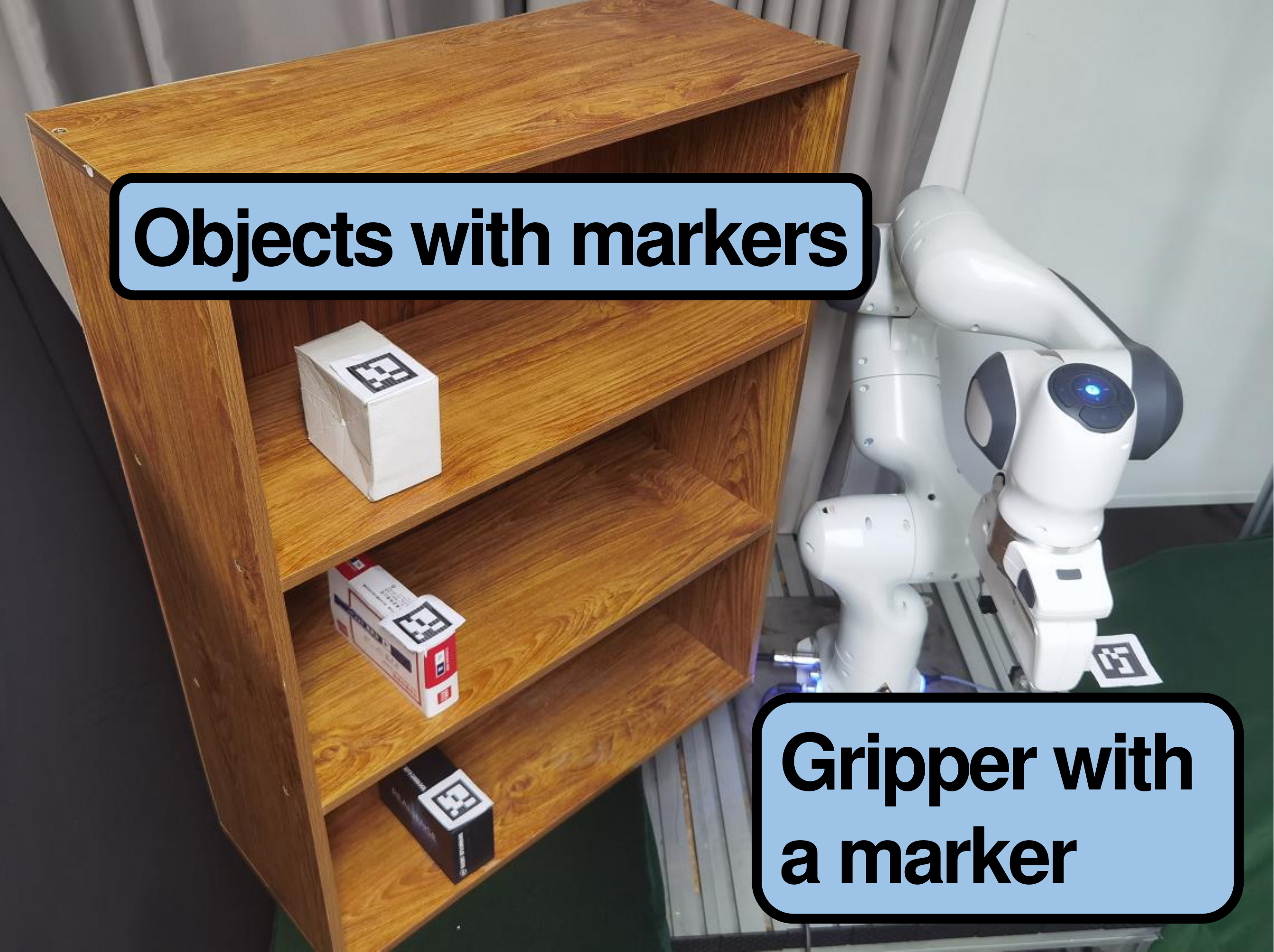}
        \caption{}\label{fig_marker}
    \end{subfigure}
    
    \medskip
    \begin{subfigure}{.32\linewidth}
        \centering
        \includegraphics[width=\linewidth]{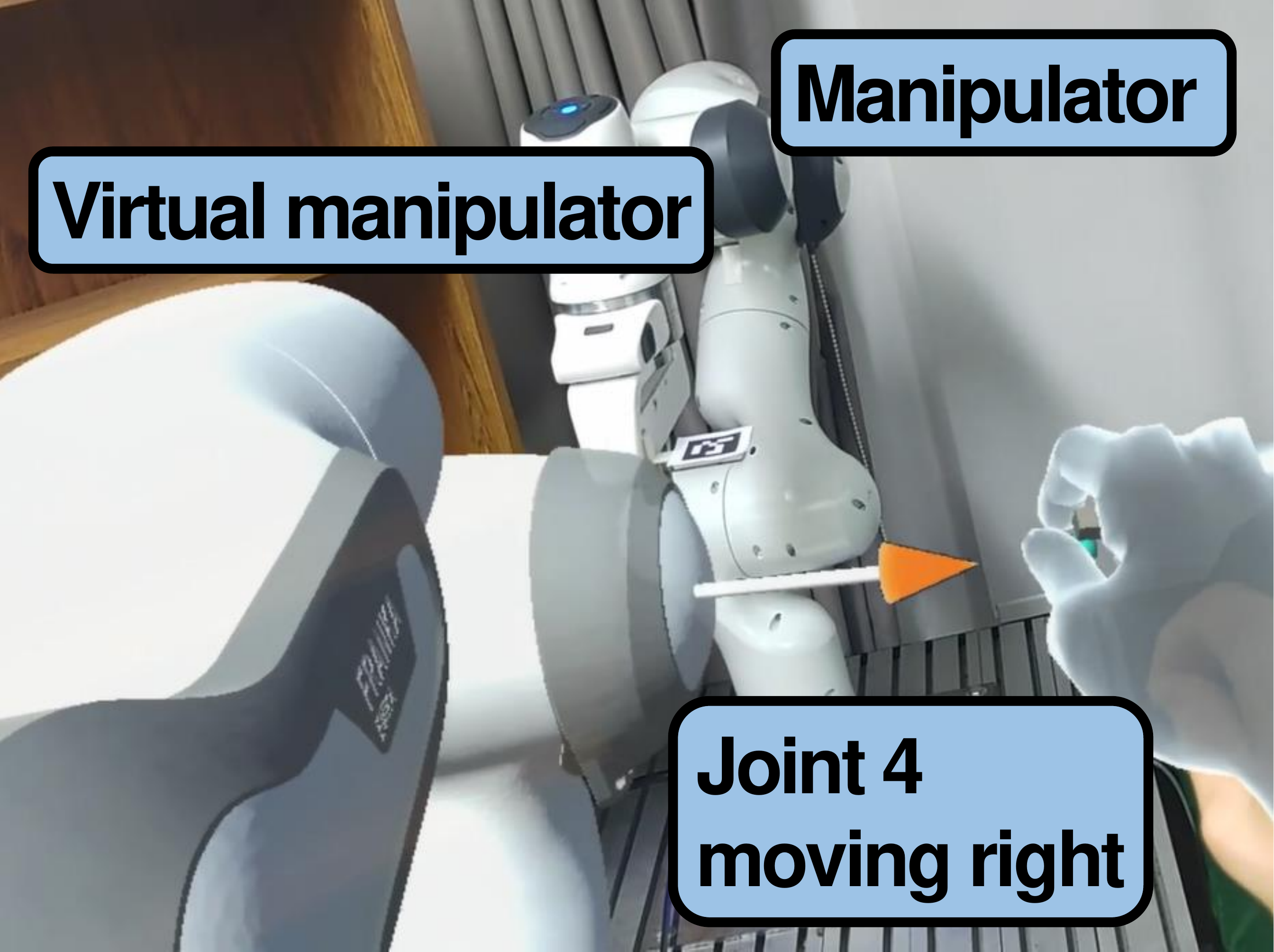}
        \caption{}\label{fig_close_manipulating}
    \end{subfigure}
    \begin{subfigure}{.32\linewidth}
        \centering
        \includegraphics[width=\linewidth]{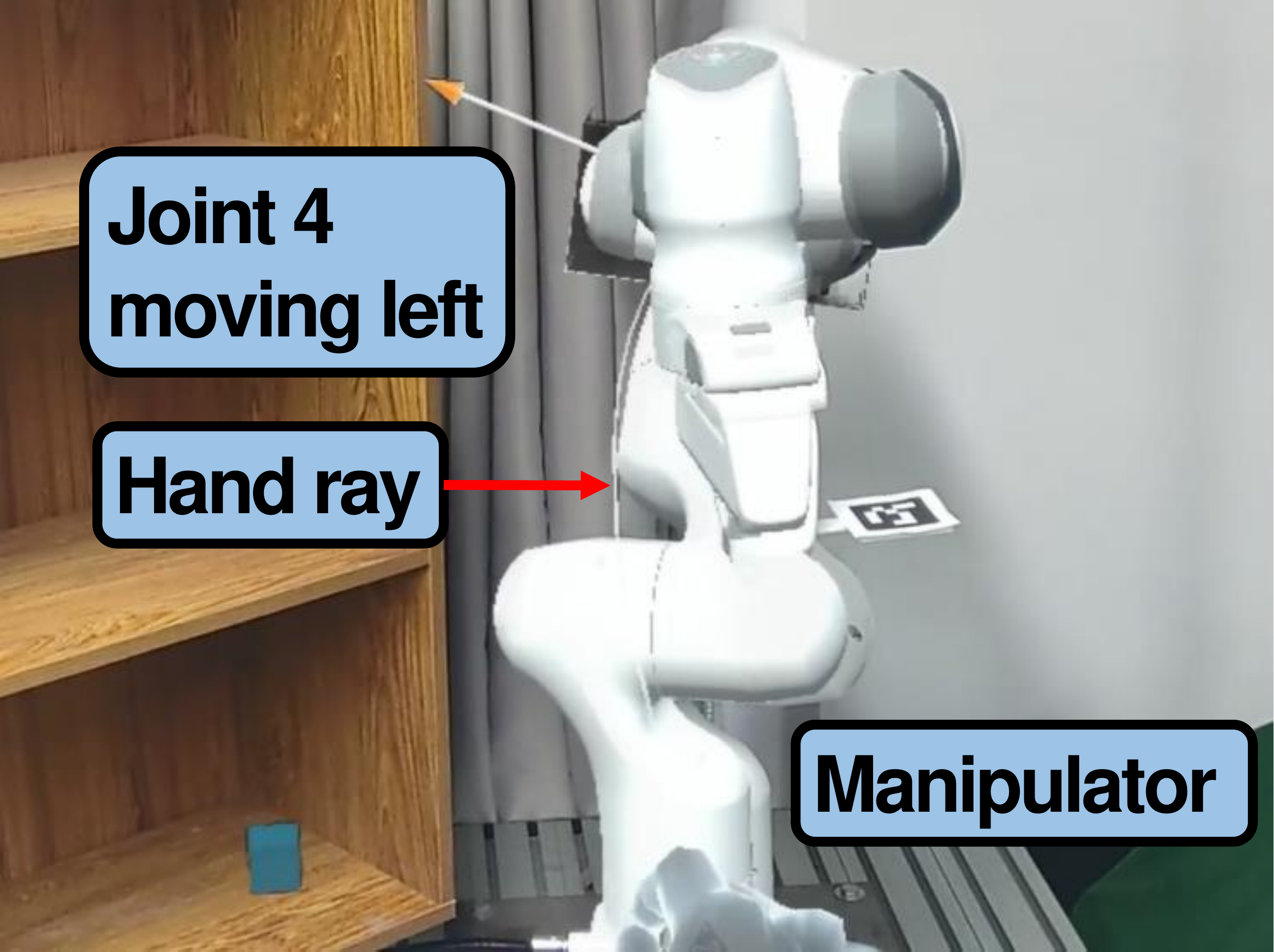}
        \caption{}\label{fig_remote_manipulating}
    \end{subfigure}
    \begin{subfigure}{.32\linewidth}
        \centering
        \includegraphics[width=\linewidth]{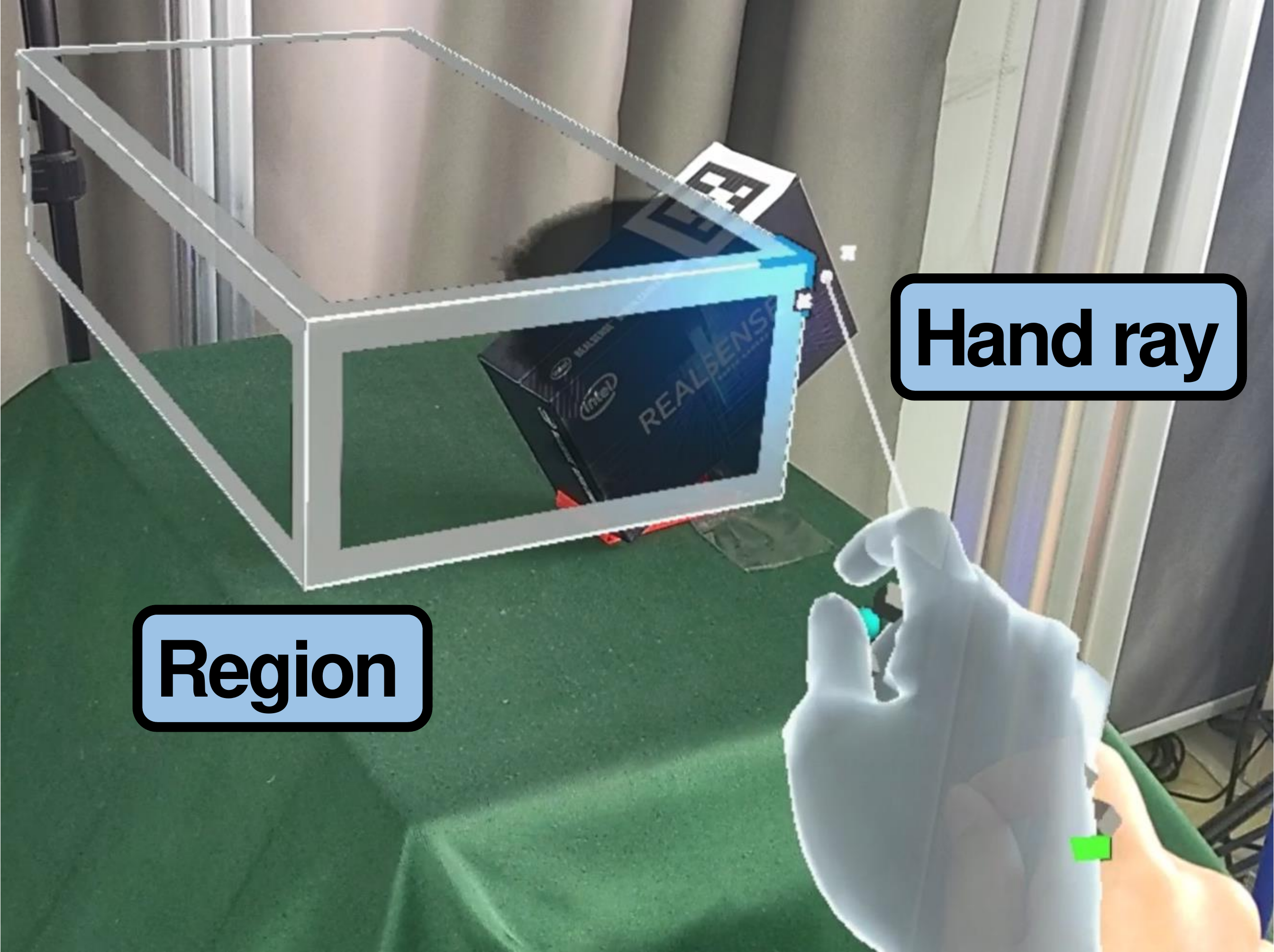}
        \caption{}\label{fig_cube}
    \end{subfigure}
  \caption{The experimental setup: (a) The overall system consists of five modules: a haptic device, an AR device, a robot manipulator, a camera, and a PC. The expert is interacting with the robot in a bi-manual way. (b) The human expert performs the demonstration task in both the task space and the null space of the robot manipulator via the mixed interaction interface (i.e., the Omega 3 and HoloLens 2). (c) Direct dragging in the AR display. (d) Hand rays and air tapping in the AR display. (e) The human expert specifies the position and the size of the Cartesian-space region in the AR display.}\label{fig_experimental_setup}
\end{figure}

\begin{figure} [!htb]
\centering
\includegraphics[width=0.8\linewidth]{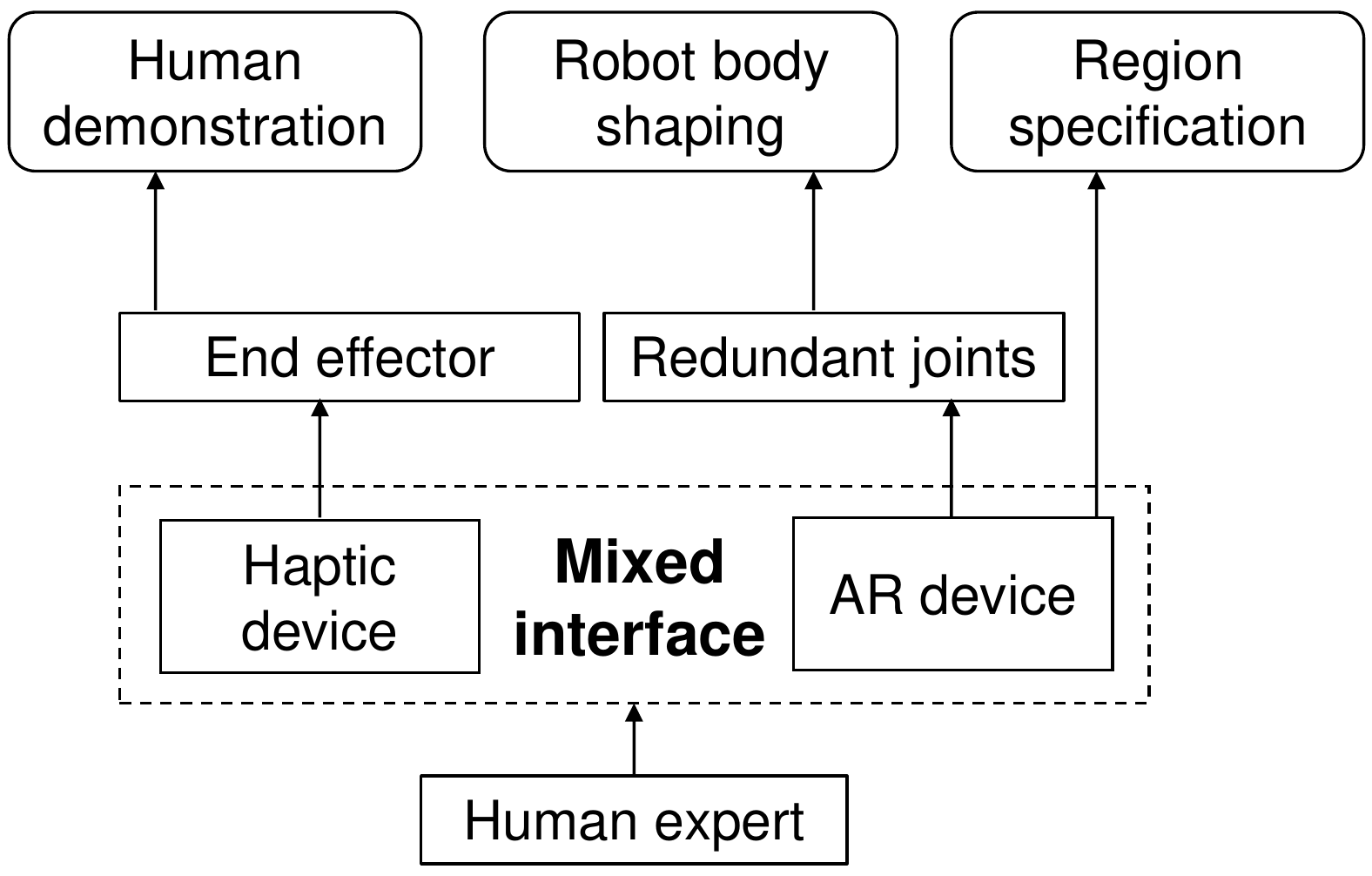}
\caption{The functions of the mixed interface, where the haptic device is used to control the robot end effector to follow human demonstration in task space, and the AR device is used to control redundant joints to follow the demonstration in joint space and also used to specify the Cartesian-space region.}\label{fig_mixed_interface}\vspace{-3mm}
\end{figure}

Experiments were conducted on a vision-based robotic manipulation system to validate the proposed method, as shown in Fig.~\ref{fig_system}. The overall system consisted of five modules: (\romannumeral1) a PC with Robot Operating System (ROS) and Ubuntu 18.04 LTS, in which the algorithm was implemented;  (\romannumeral2) a 7-DOF Franka robot with a two-fingered gripper, with ArUco markers attached to the gripper and the objects to aid perception (Fig. \ref{fig_marker}); 
(\romannumeral3) a Basler ace acA1440-220uc camera with \(1440 \times 1080\) resolution, which was fixed in the workspace of the robot but not calibrated;
(\romannumeral4) an Omega 3, which is a haptic interface developed by Force Dimension; and (\romannumeral5) a HoloLens 2, which is a head-mounted AR device. Items (\romannumeral4) and (\romannumeral5) comprised a mixed interface, which enabled the human expert to interact with the robot in both task space and null space in a bi-manual way. The functions of the AR-haptic mixed interface are described in Fig. \ref{fig_mixed_interface}.


The AR device allowed the human expert to exert control efforts on the robot manipulator in two ways:
by pulling the virtual robot closer and directly manipulating a specific joint (Fig.~\ref{fig_close_manipulating}) or by making the virtual robot overlap with the real robot and then using hand-ray and air-tap gestures to control the real robot remotely (Fig.~\ref{fig_remote_manipulating}).
These control efforts are represented as a vector (visualized in Fig. \ref{fig_close_manipulating} and Fig. \ref{fig_remote_manipulating}) that is converted to a command velocity proportionally. Then, the control efforts \(\bm d\) are injected by projecting the velocity back to the joint space using the pseudo-inverse of the Jacobian matrix of the selected joint. The method used to calculate \(\bm d\) is the same as that used in \cite{icra_22}.
In addition, the AR device allowed the human expert to conveniently specify the Cartesian-space region (Fig.~\ref{fig_cube}) by simply drawing a virtual region.

The following three experimental tasks were carried out to illustrate the performance of the proposed method. The purposes of experiments are detailed as follows.
\begin{enumerate}
    \item[1)] {\em Placing Task} - to demonstrate how the cobot learned the desired trajectory in a complex unstructured environment. 
    
    \item[2)] {\em Grasping Task} - to validate the effectiveness of the global adaptive controller, in the presence of a large-scale transition, an uncalibrated camera, and joint limits. 
    
    \item[3)] {\em Collaboration Task} - to illustrate the entire pipeline, in which the robot transferred an object from an interactive environment (i.e., humans coexist) to an isolated environment. Such a task is commonly performed in many factories, e.g., during the transer of hazardous chemicals.
\end{enumerate}

Note that the Franka robot accepts both torque input and velocity input. In this paper, the proposed control scheme is implemented at the kinematic level (i.e., the velocity input) as the robot moved at a relatively low velocity for human-robot collaboration; The control algorithm is developed based on Franka-interface and Frankapy control stack \cite{zhang2020modular}.

\subsection{Placing Task}
In Experiment 1, the robot was already grasping an object and it was manipulated to place the object at a goal position on a shelf. The trajectory to the goal position was learnt from human demonstration via the DMP method and the mixed interface (i.e., Omega 3 and HoloLens 2).

During the demonstration, the translation of the robot end effector was defined by using Omega 3 and the orientation was defined simultaneously according to the translation as follows:
\begin{subnumcases}{\label{eq:coarse_orientation} r_5=}
  \frac{r_2}{Y_0}\times \beta_0\,, & \(\lvert r_2\rvert\leq Y_0\)\,, \\
  -\beta_0\,, & \(r_2<-Y_0\)\,, \\
  \beta_0\,, & \(r_2>Y_0\)\,,
\end{subnumcases}
where \(\bm r=[r_1,r_2,\cdots,r_6]^T\) is the pose of the end effector in the base frame, 
\(Y_0=
0.6 m\) 
and 
\(\beta_0=
\ang{90}\) 
are the maximum reachable translation range in y-axis and the orientation range along the y-axis, respectively.


Note that the orientation defined in (\ref{eq:coarse_orientation}) is relatively coarse, which would be further adjusted with HoloLens 2; The demonstration via HoloLens 2 was thus used to define the motion of redundant joints in null space, not affecting the position of the robot end effector but shaping its orientation for fine-tuning.  

\begin{figure} [!tb]
  \begin{subfigure}{.475\linewidth}
    \centering
    \includegraphics[width=\linewidth]{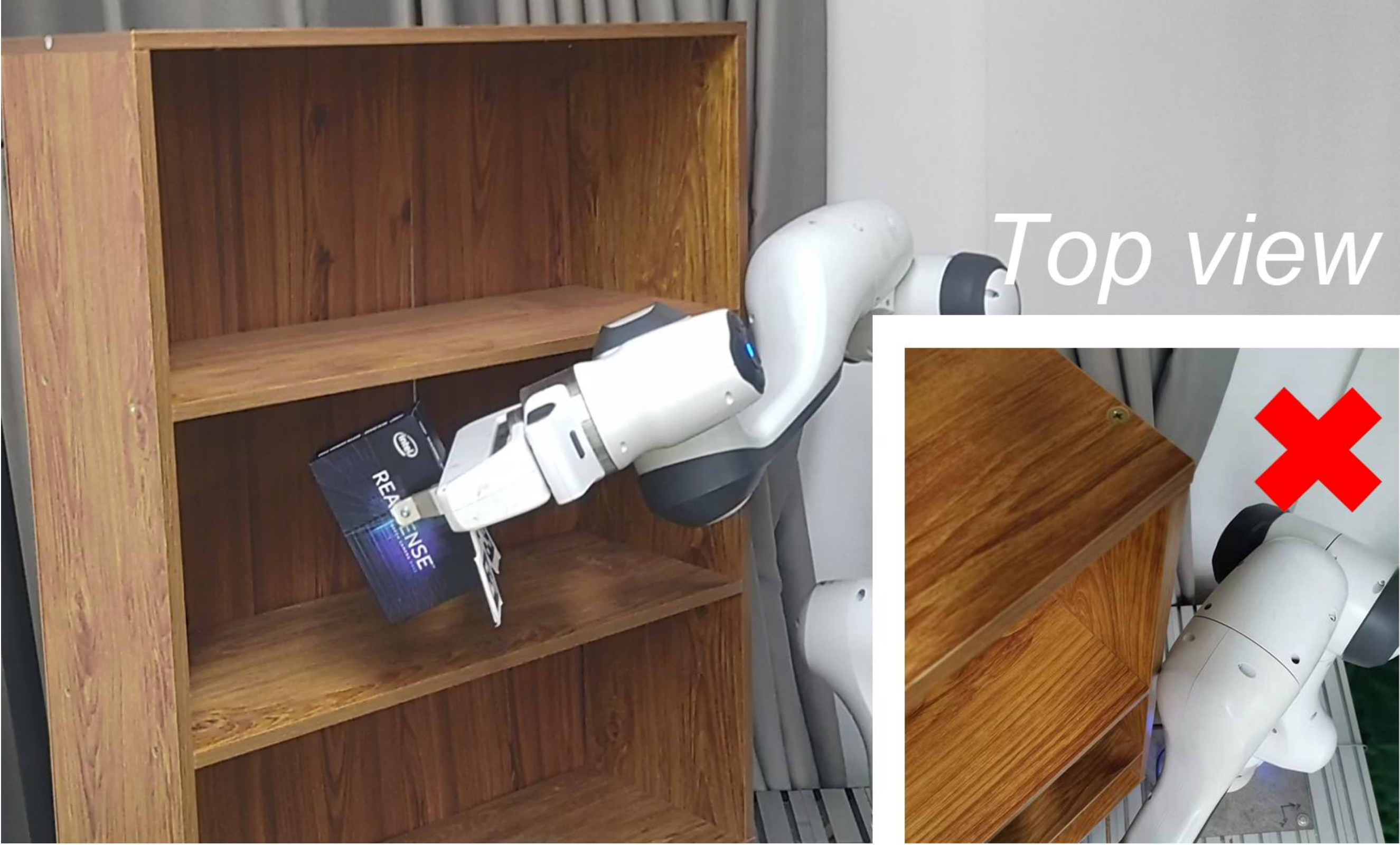}
    \caption{}
  \end{subfigure}
  \hfill
  \begin{subfigure}{.475\linewidth}
    \centering
    \includegraphics[width=\linewidth]{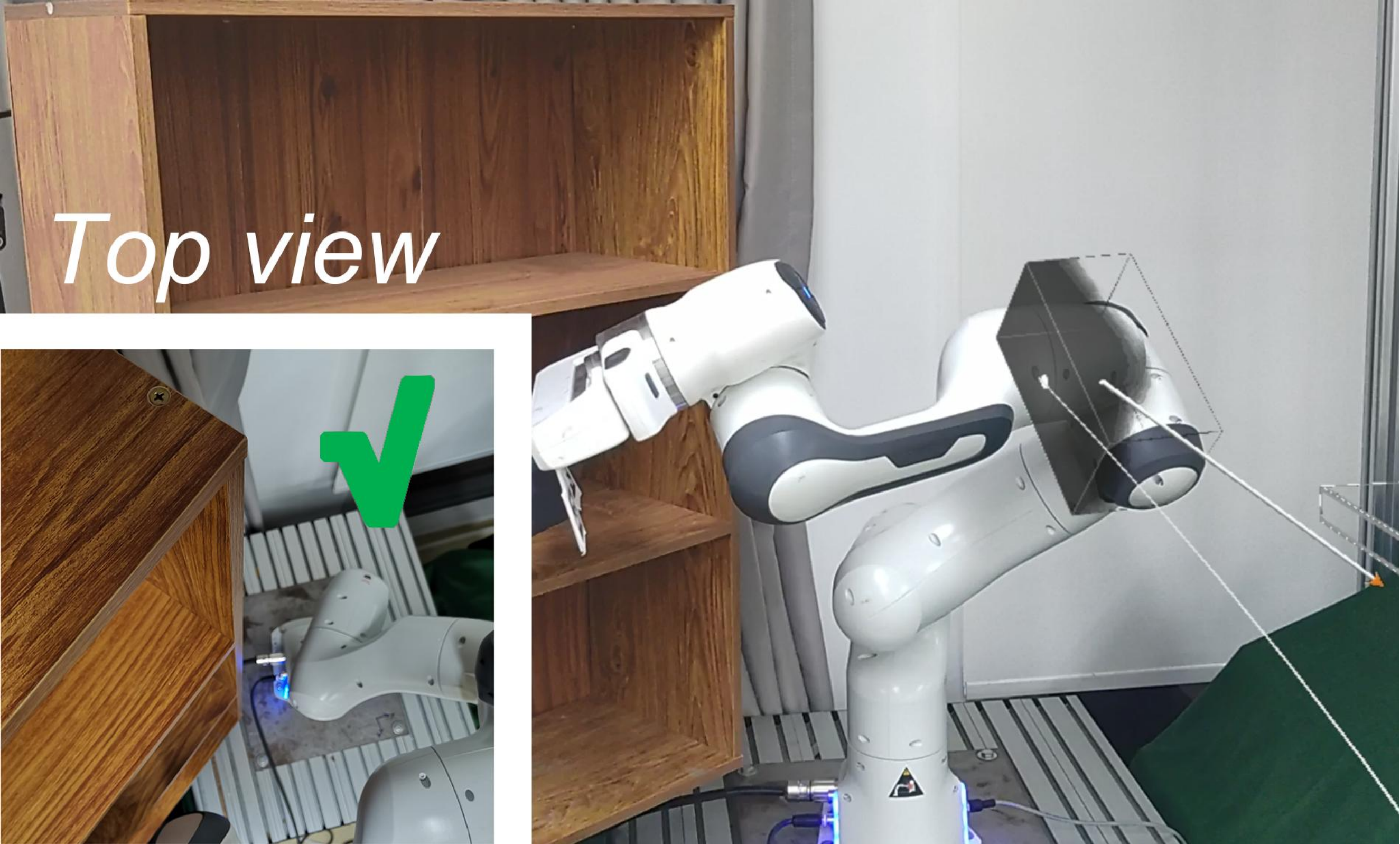}
    \caption{}
  \end{subfigure}
  \caption{Experiment 1 - Goal position inside the cabinet:
  (a) The \(4\)th link collides with the cabinet due to inappropriate body shape; (b) The collision is avoided after the body shape is adjusted.}
  \label{fig_e1_comparison}
  \vspace{-3mm}
\end{figure}

\begin{figure} [!htb]
  \begin{subfigure}{.495\linewidth}
    \centering
    \includegraphics[width=\linewidth]{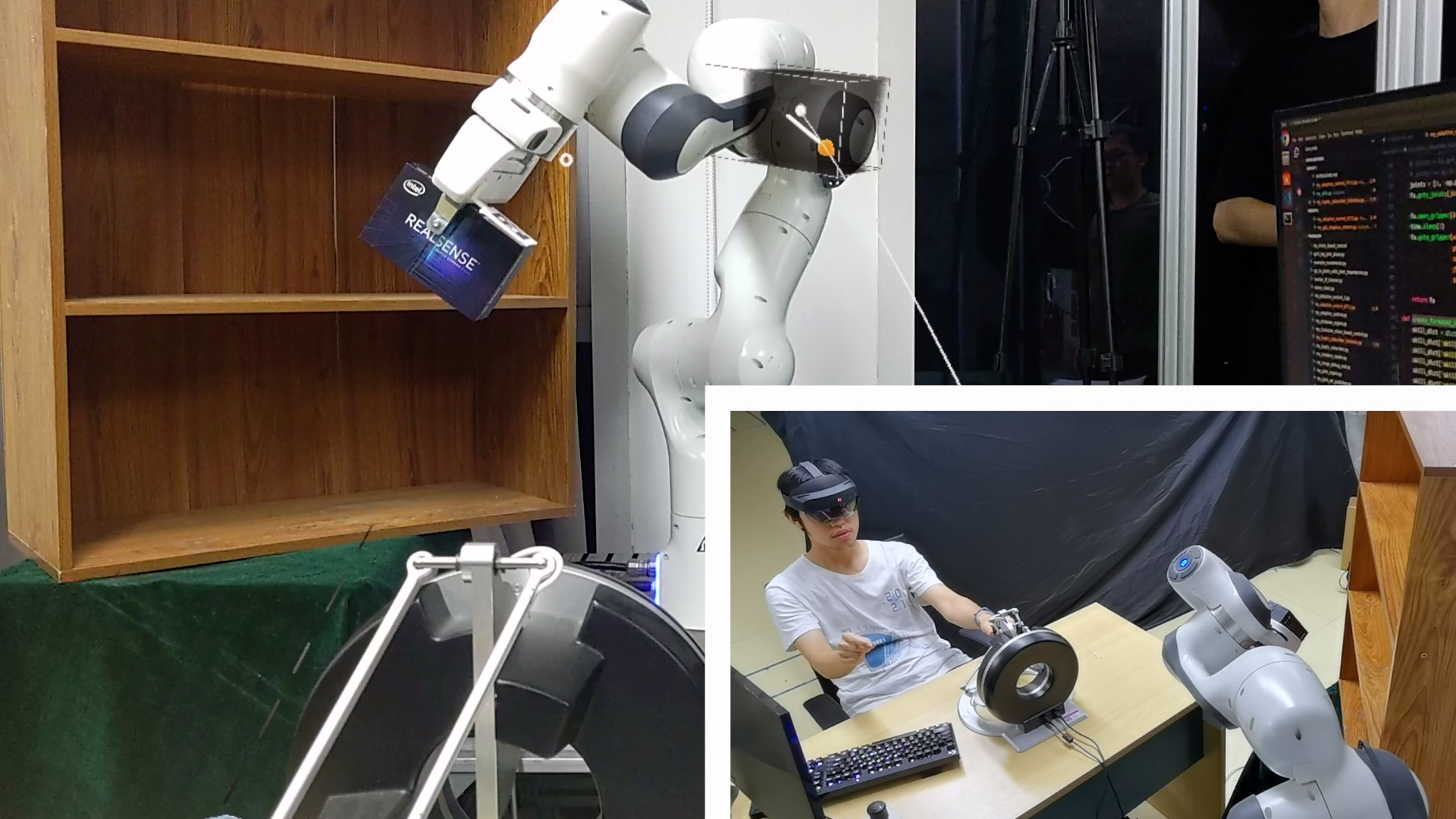}
    \caption{}
  \end{subfigure}
  \hfill
  \begin{subfigure}{.495\linewidth}
    \centering
    \includegraphics[width=\linewidth]{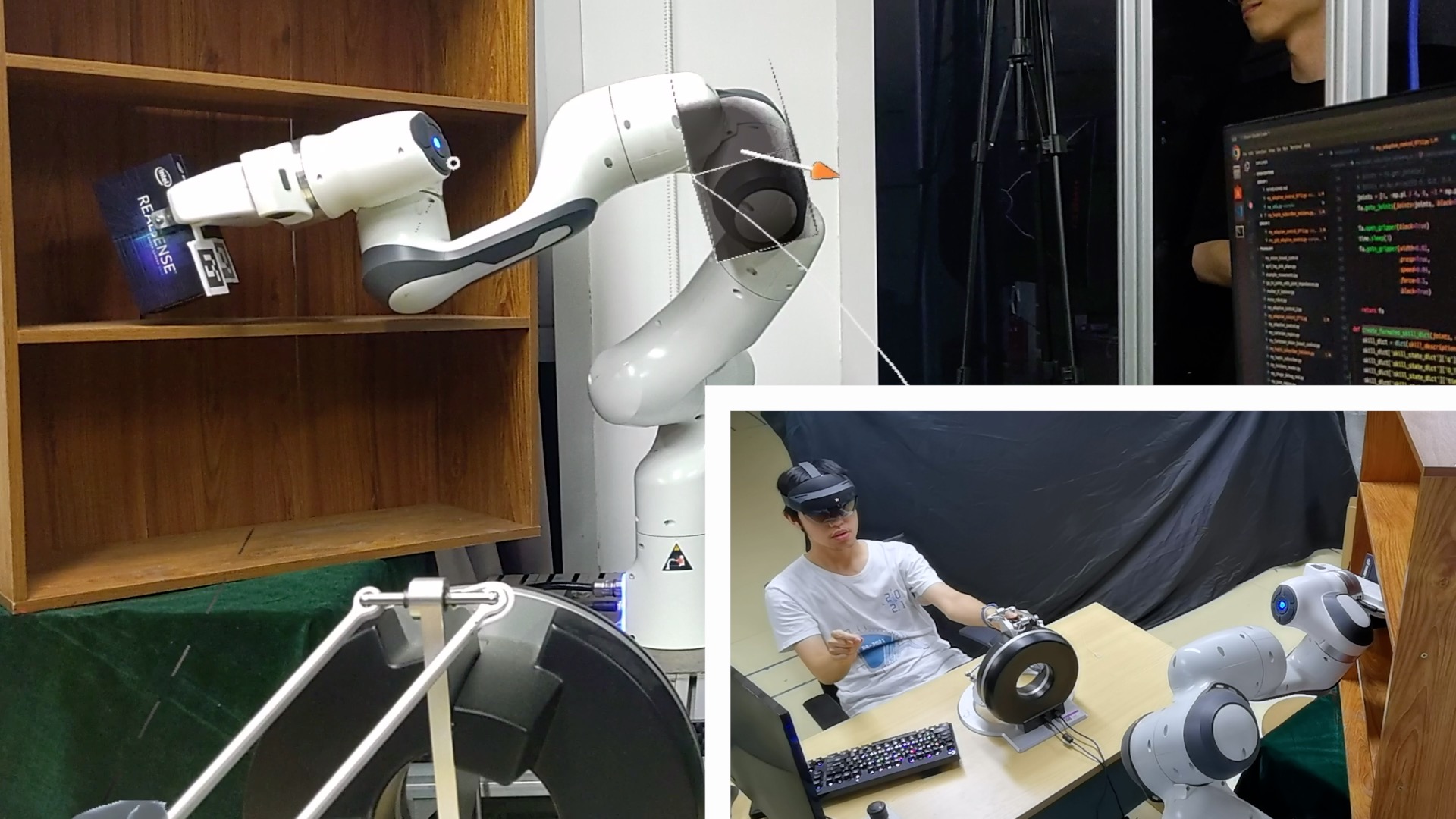}
    \caption{}
  \end{subfigure}
  
  \medskip
  \begin{subfigure}{.495\linewidth}
    \centering
    \includegraphics[width=\linewidth]{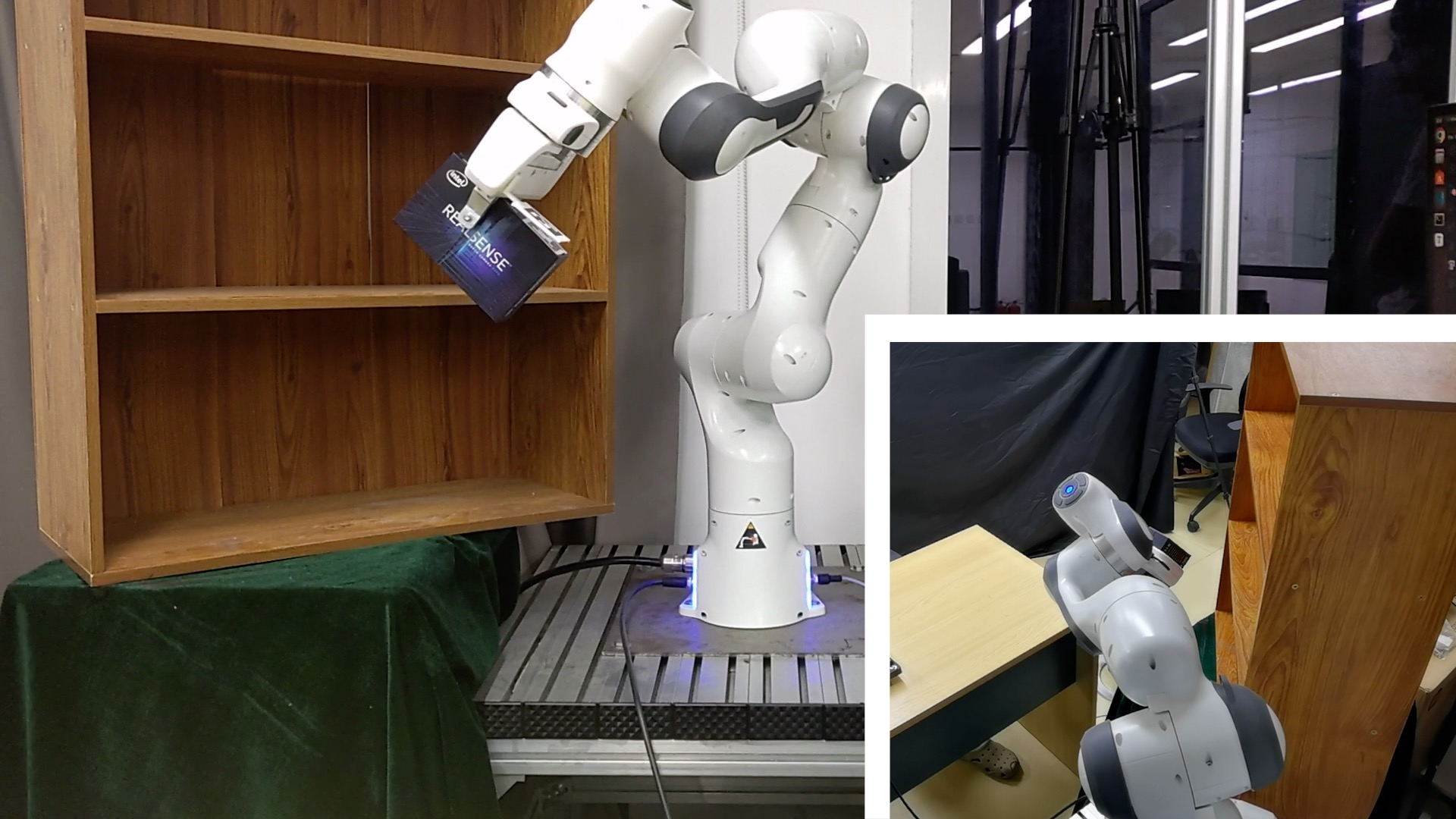}
    \caption{}
  \end{subfigure}
  \hfill
  \begin{subfigure}{.495\linewidth}
    \centering
    \includegraphics[width=\linewidth]{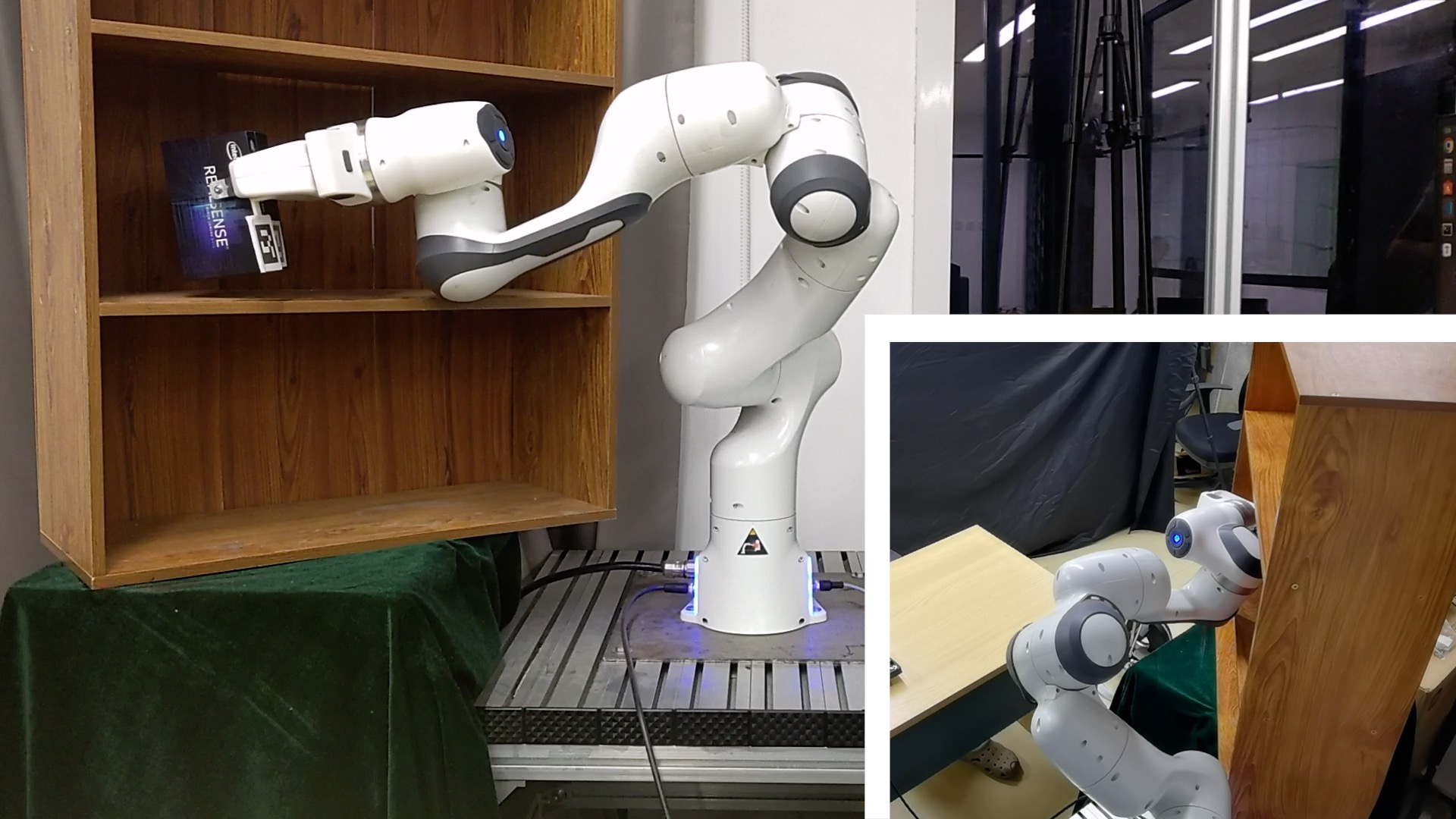}
    \caption{}
  \end{subfigure}
  \caption{Experiment 1 - Snapshots: the human expert performed the demonstration via the mixed interface, and the robot learnt and reproduced the trajectory. (a) {\em Learning phase} \(t=10.3\mathrm{s}\): The human expert pulled the joint \(4\) of a virtual robot via HoloLens 2; (b) {\em learning phase} \(t=24.7\mathrm{s}\): the real robot adjusted its body shape accordingly, without affecting its end effector; (c) {\em reproducing phase}  \(t=6.8\mathrm{s}\): the robot followed the learnt trajectory but at a higher velocity; (d) {\em reproducing phase} \(t=16.7\mathrm{s}\): the robot placed the object at the goal position, without colliding with the cabinet.}
  \label{fig_e1_snapshots}
  \vspace{-4mm}
\end{figure}

The goal position on the shelf was shown in Fig.~\ref{fig_e1_comparison}.
It was insufficient to define only the position and orientation of the robot end effector, as an inappropriate body shape would result in the robot colliding with the cabinet (e.g., the collision of the \(4\)th joint with the cabinet in Fig.~\ref{fig_e1_comparison}). To solve this problem, the human expert used HoloLens 2 to define the motion of redundant joint and thus ``pulled'' the joint away from the cabinet. During the demonstration, the goal position was on the second shelf of the cabinet. The reproduction results shown in Fig.~\ref{fig_e1_snapshots} indicated that the robot reached the goal position by following the learnt trajectory at a higher speed.
\begin{figure} [!htb]
  \centering 
    \includegraphics[width=0.8\linewidth]{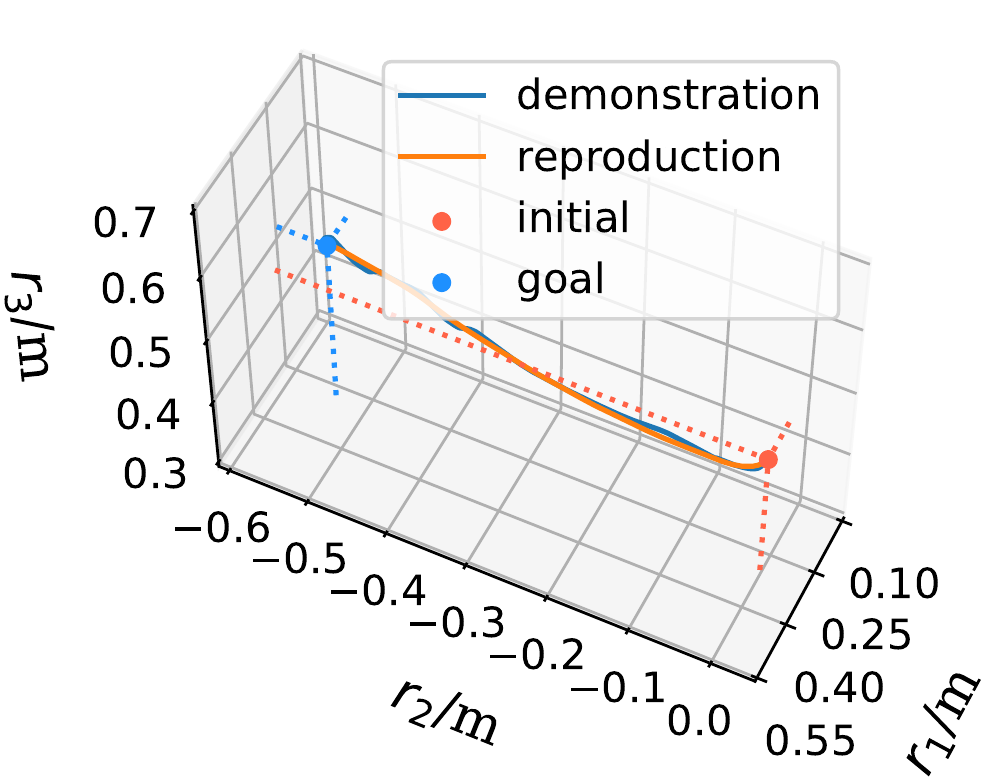}
  \vspace{-1mm}
  \caption{Experiment 1 - Trajectories of the demonstration and reproduction in 3D space}
  \label{fig:e1_traj_3d}
  \vspace{-1mm}
\end{figure}
The trajectories of the robot end effector during the demonstration and the reproduction in 3D space are plotted in Fig.~\ref{fig:e1_traj_3d}, showing that the DMP-based learning method successfully captured the demonstration trajectory while removing several unnecessary jerks as well. The goal position of the learnt trajectory can also be set to positions on other shelves such that the grasped object is placed there accordingly.
\begin{figure} [htb]
  \begin{subfigure}{.24\linewidth}
    \includegraphics[height=2cm]{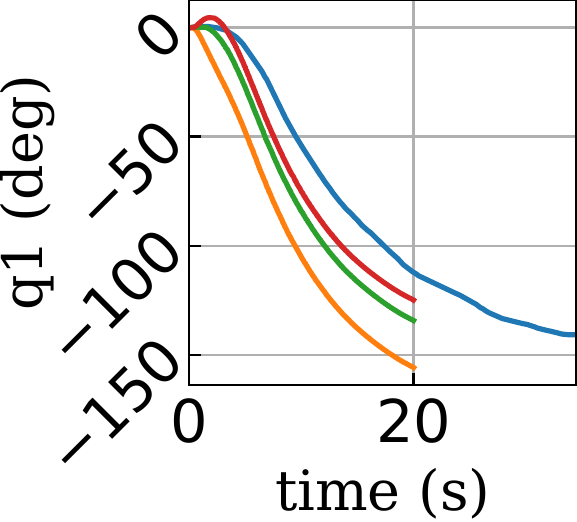}
  \caption{}\label{joint_motion1}
  \end{subfigure}
  \begin{subfigure}{.24\linewidth}
    \includegraphics[height=2cm]{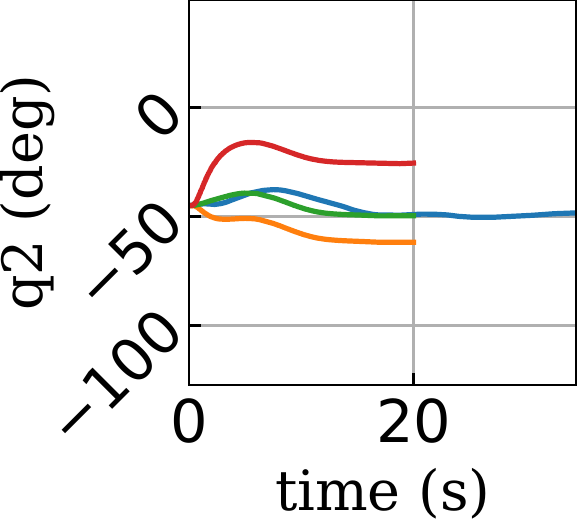}
  \caption{}\label{joint_motion2}
  \end{subfigure}
  \begin{subfigure}{.24\linewidth}
    \includegraphics[height=2cm]{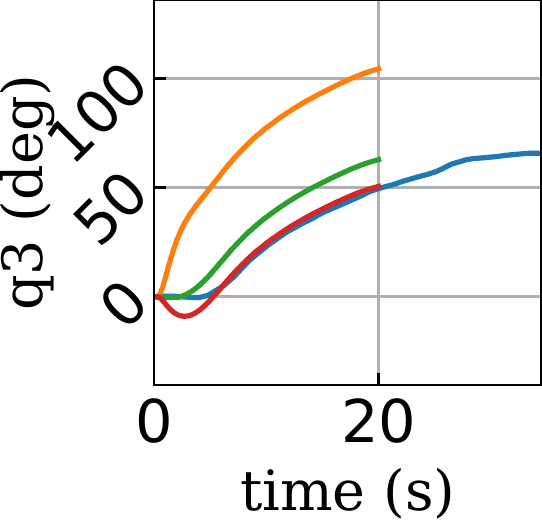}
  \caption{}\label{joint_motion3}
  \end{subfigure}
  \begin{subfigure}{.24\linewidth}
    \includegraphics[height=2cm]{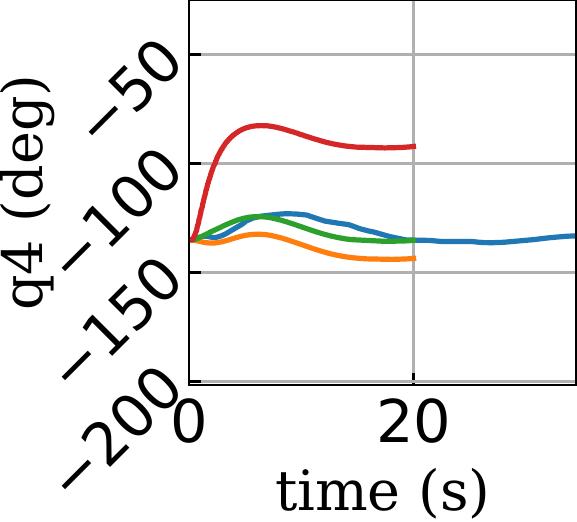}
  \caption{}\label{joint_motion4}
  \end{subfigure}

  \medskip
  \begin{subfigure}{.24\linewidth}
    \includegraphics[height=2cm]{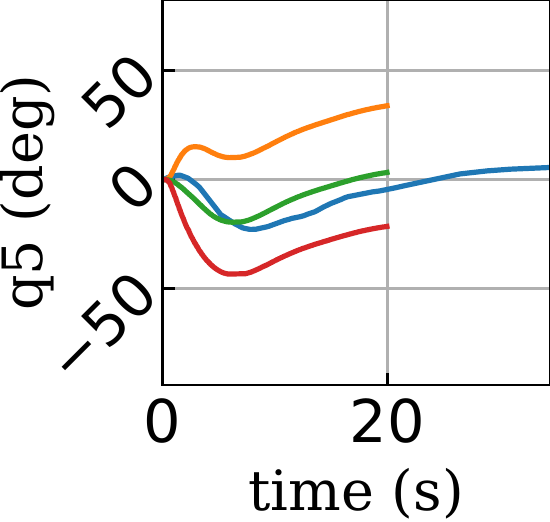}
  \caption{}\label{joint_motion5}
  \end{subfigure}
  \begin{subfigure}{.24\linewidth}
    \includegraphics[height=2cm]{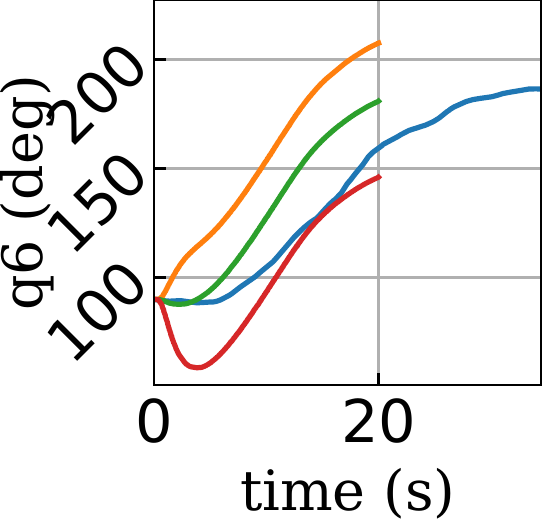}
  \caption{}\label{joint_motion6}
  \end{subfigure}
  \begin{subfigure}{.5\linewidth}
    \includegraphics[height=2cm]{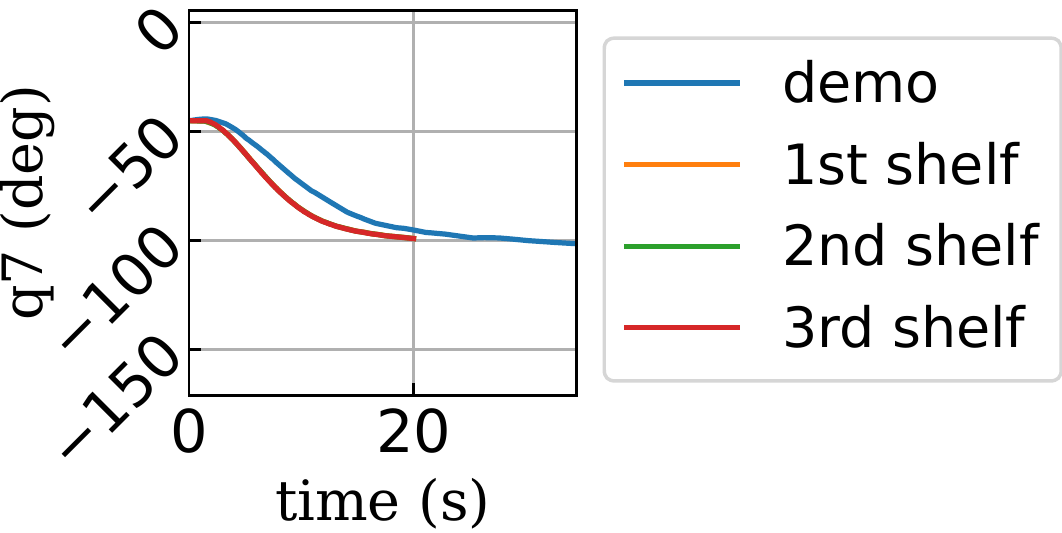}
  \caption{}\label{joint_motion7}
  \end{subfigure}
  \caption{Experiment 1 - Joint motions: the ``demo" denotes the motion in the demonstration phase, and ``1st, 2nd, 3rd'' represent the motion for different goal positions in the reproducing phase. The goal positions were on three different shelves of the cabinet. The ``2nd" shelf goal position was the same as the goal position in the demonstration. The trajectories were reproduced with \(\tau=1.5\), thus they terminated earlier than the demonstration, at \(t=20.0\mathrm{s}\).
    (a) joint 1; 
    (b) joint 2;
    (c) joint 3;
    (d) joint 4;
    (e) joint 5; 
    (f) joint 6;
    (g) joint 7 (the reproduced trajectories overlapped with each other).}
    \label{fig_e1_q}
\end{figure}
Note that the robot's movement was learnt and reproduced in joint space, where the corresponding joint angles were computed based on the analytical inverse kinematics method \cite{HeLiu2021}. In addition, the angle of joint \(7\) was fixed as the redundant parameter, thus the reproduced trajectories with modified goals overlapped with each other (Fig.~\ref{joint_motion7}). 

From the joint motions illustrated in Fig.~\ref{fig_e1_q}, the goal positions of the learnt trajectories were successfully changed to other ones (i.e., the positions on the lower or higher shelves). These new trajectories were implemented in the subsequent task (see Section VII.C) to place multiple objects at different shelves without colliding with the cabinet. 

 \begin{figure*} [!t]
  \begin{subfigure}{.245\linewidth}
    \centering
    \includegraphics[width=\linewidth]{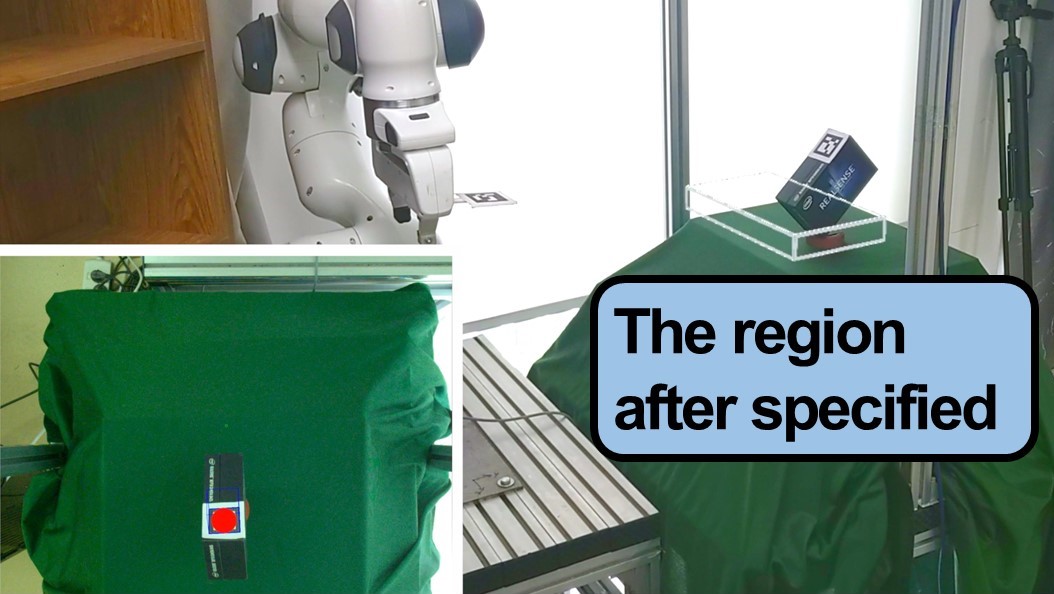}
    \caption{}
  \end{subfigure}
  \hfill
  \begin{subfigure}{.245\linewidth}
    \centering
    \includegraphics[width=\linewidth]{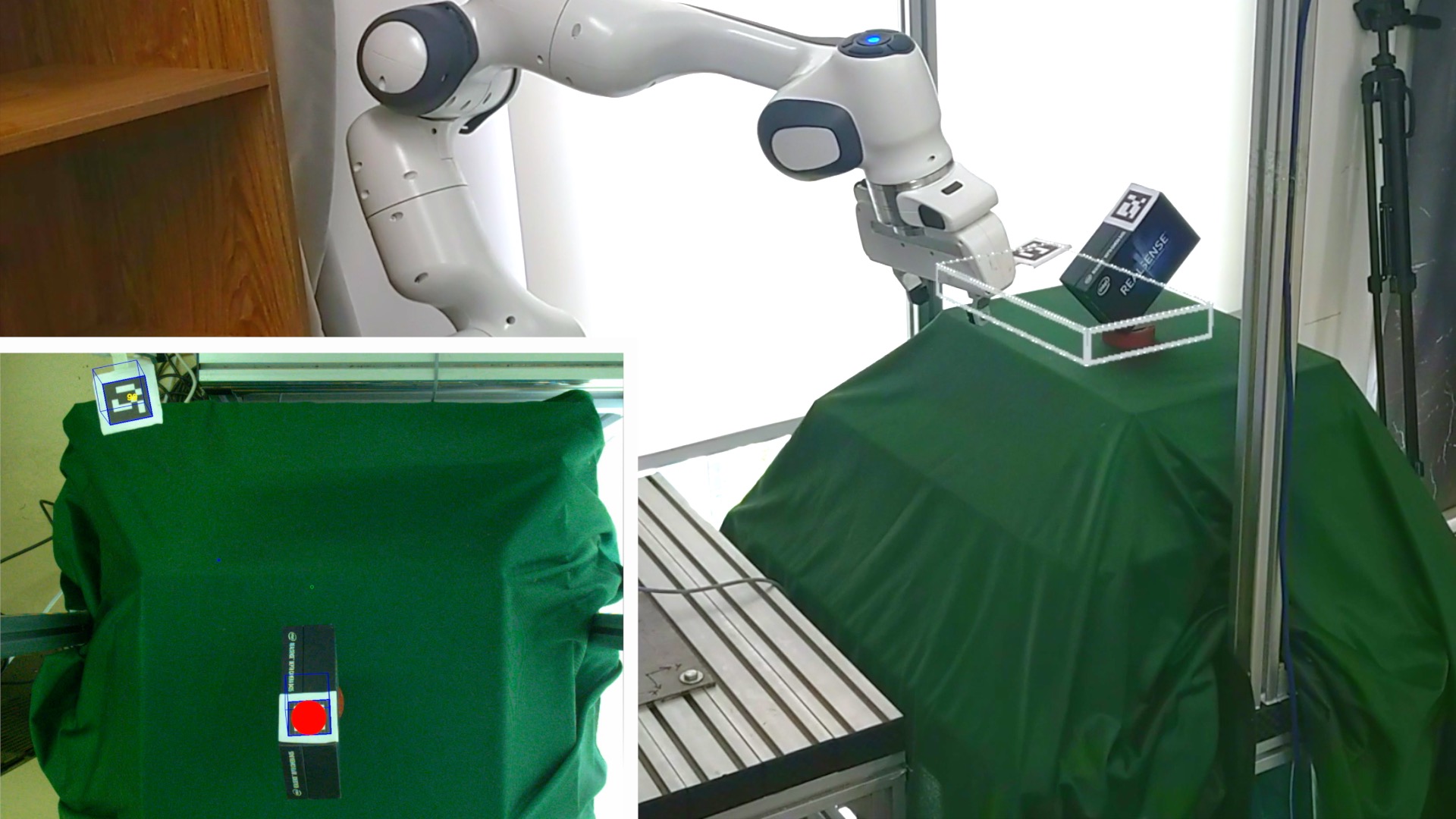}
    \caption{}
  \end{subfigure}
  \hfill
  \begin{subfigure}{.245\linewidth}
    \centering
    \includegraphics[width=\linewidth]{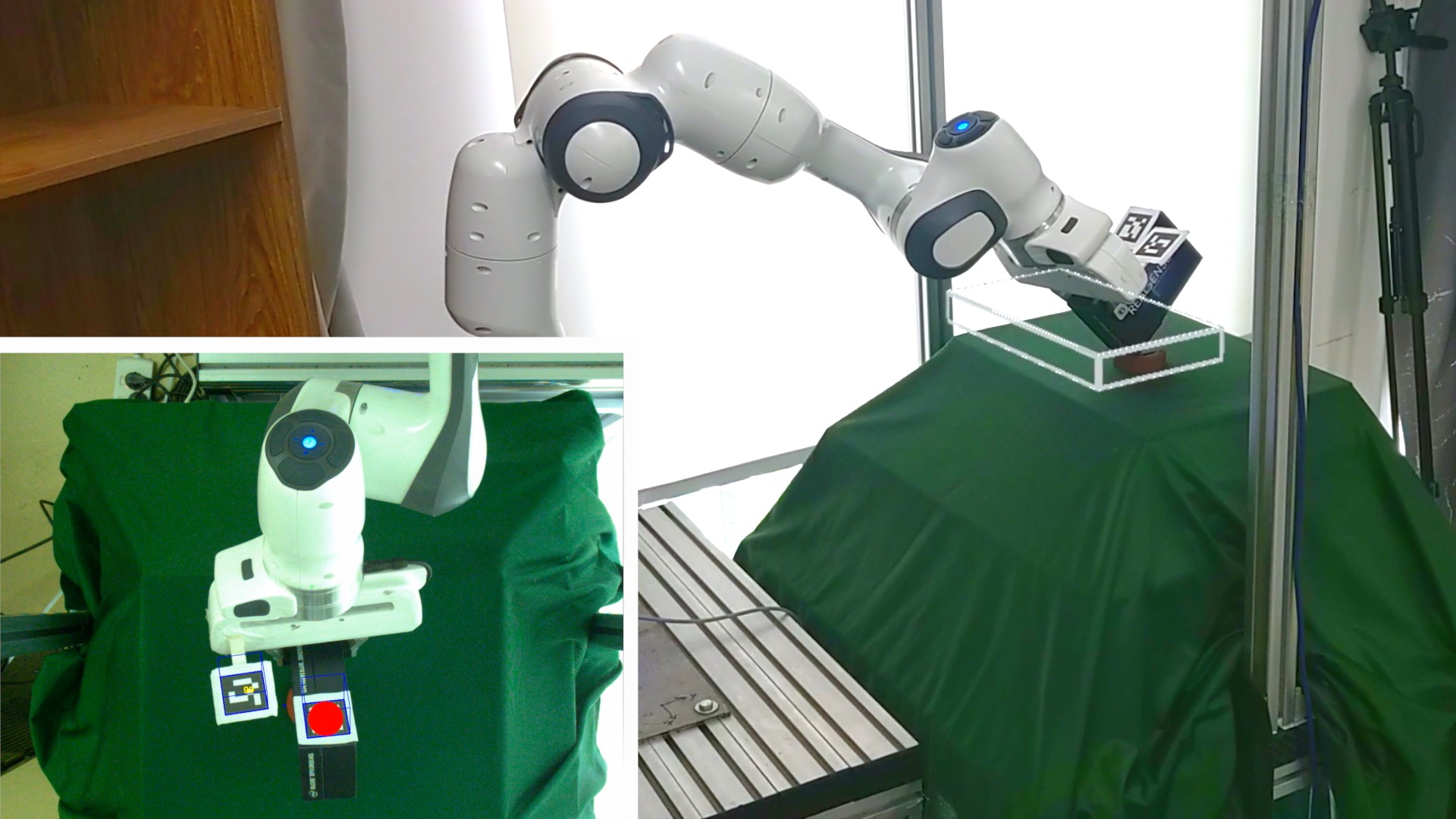}
    \caption{}
  \end{subfigure}
  \hfill
  \begin{subfigure}{.245\linewidth}
    \centering
    \includegraphics[width=\linewidth]{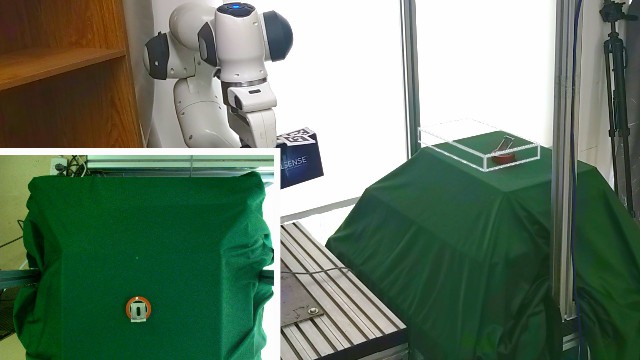}
    \caption{}
  \end{subfigure}
  \caption{Experiment 2 - Snapshots: The robot started from a remote initial position and moved to the target object to grasp it, in the presence of joint limits, uncalibrated camera, and limited FOV. The Cartesian-space region was defined by the expert via the AR interface and represented as a transparent cube with gray edges. 
  (a) \(t=0.0\mathrm{s}\): the initial configuration; (b) \(t=5.8\mathrm{s}\): after the robot entered the Cartesian-space region, the marker appeared in the FOV; (c) \(t=9.2\mathrm{s}\): the robot employed the visual feedback to adjust its pose to aim at the target object; (d) \(t=15.0\mathrm{s}\): the robot grasped the object then returned to the home pose.}
  \label{fig:e2_snapshots}
\end{figure*}
 \begin{figure} [!tb]
  \begin{subfigure}{.415\linewidth}
    \centering
    \includegraphics[height=3.3cm]{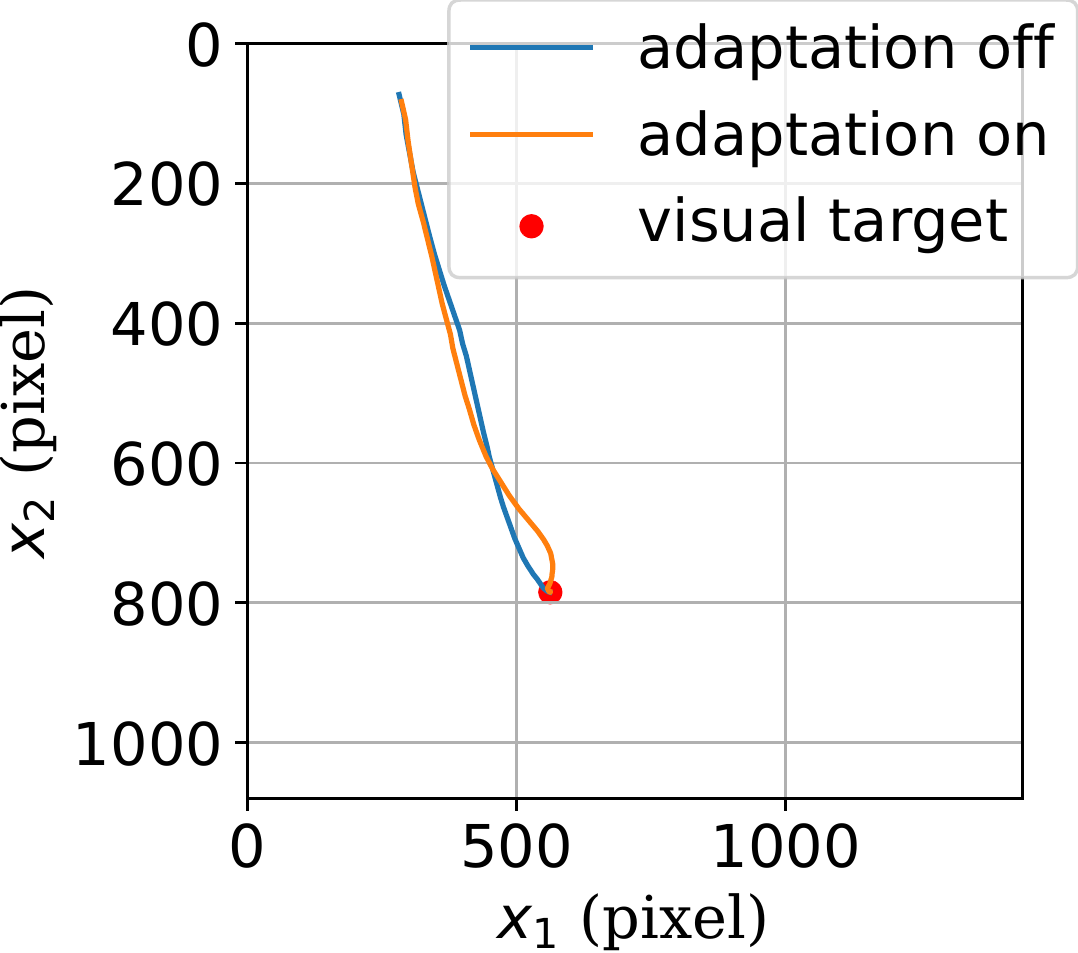}
    \caption{}
  \end{subfigure}
  \hfill
  \begin{subfigure}{.585\linewidth}
    \centering
    \includegraphics[height=3.3cm]{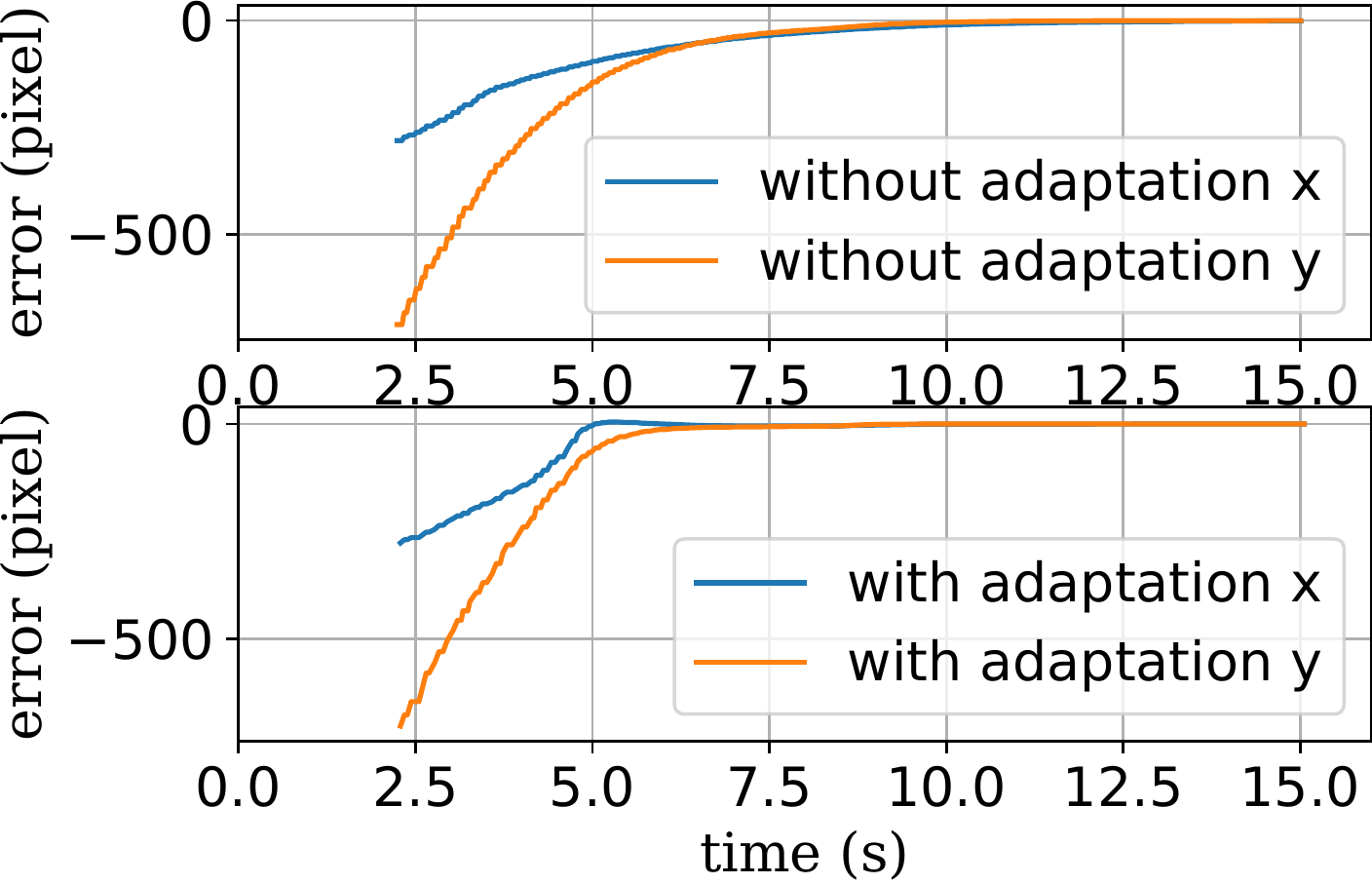}
    \caption{}
  \end{subfigure}
  \caption{Experiment 2 - Results: (a) The path of the robot end effector in vision space; (b) The position errors in vision space.}
  \label{fig_e2_vision_trajectories}
\end{figure}
\begin{figure} [!tb]
\centering
\includegraphics[width=0.6\linewidth]{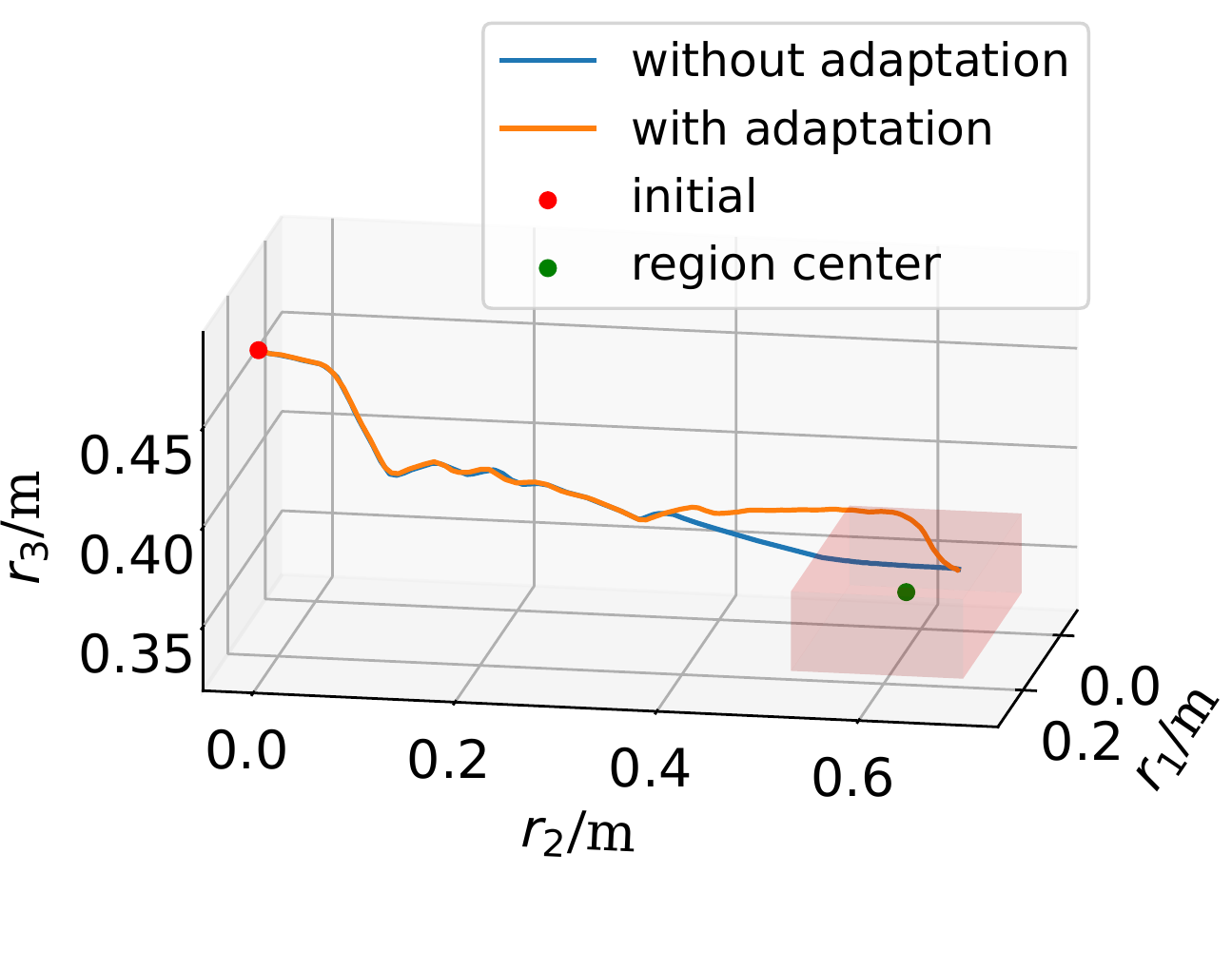}
\vspace{-4mm}
\caption{Experiment 2 - The path of the robot end effector in 3D space after entering the FOV}\label{fig_e2_traj_3d}
\end{figure}

\subsection{Grasping Task}
In Experiment 2, the robot started from a remote initial position to grasp a target object,
as shown in Fig.~\ref{fig:e2_snapshots}a. The proposed global adaptive controller defined by (\ref{globalControl}) and (\ref{update}) was implemented to drive the robot to the object's position for grasping, in the presence of joint limits, uncalibrated camera, and limited FOV. 

Three regions were specified in the workspace of the robot: the joint-space region (\ref{psx2}), the Cartesian-space region (\ref{CaFun}), (\ref{OrtRegion}), and the vision region (\ref{eq_vision_re_func}). 
The joint-space region was introduced to prevent the robot entering joint limits, which could have resulted in an emergency stop.
The vision region was specified to cover the position of the target object, such that the robot was able to employ the visual feedback 
for grasping. The Cartesian-space region was defined to dominate the remaining workspace of the robot, to ensure a smooth transition between different feedback; Specifically, the position region in Cartesian space was defined by the human expert via the AR interface (based on the approximate location of the FOV), and the quaternion region in Cartesian space was pre-defined. The combination of all the region feedback guaranteed the movement of the robot within the whole workspace.

The control parameters are listed in Table \ref{exp_para_table}. The initial image Jacobian \(\hat{\bm J}_{s}(\bm r)_{t=0}\) was set randomly as in (\ref{eq_js0}), and the initial weight \(\hat{\bm W}_{t=t_0}\) was set according to \(\hat{\bm J}_{s}(\bm r)_{t=0}\)  by (\ref{eq_w0}), where \(t_0\) refers to the moment when the marker entered the FOV. In addition, the human expert did not exert additional control efforts on the robot in this experiment, and hence \(\bm d = \bm 0\).
\begin{table*}[!h]
\renewcommand{\arraystretch}{1.3}
\centering
\caption{Control parameters in Experiment 2}\label{exp_para_table}
\begin{tabular}{ccc}
\hline
\multirow{3}{*}{\begin{tabular}[c]{@{}c@{}}\textbf{Joint-Space Region}\\ (\ref{jointxp})\end{tabular}} & \(k_q,k_r\)   & 10, 1   \\
                                                                                             & \(f\)   & $(q_i-q_{imin/max})^2-0.1^2\leq0$ \\
                                                                                             & \(f_r\) & $(q_i-q_{imin/max})^2-0.3^2\leq0$\\ \hline
\multirow{3}{*}{\begin{tabular}[c]{@{}c@{}}\textbf{Position Region in}\\ \textbf{Cartesian Space} (\ref{CaFun})(\ref{CaPeFun})\end{tabular}}    & $r_c$ (center)       & by expert                         \\
                                                                                             & $k_{c1}, k_{c2}, k_{c3}$       & $[4e-4,4e-4,4e-5]^T$                         \\
                                                                                             & $c_1, c_2, c_3$ (size)     & by expert                        \\ \hline
\multirow{3}{*}{\begin{tabular}[c]{@{}c@{}}\textbf{Orientation Region in}\\ \textbf{Cartesian Space} (\ref{OrtRegion})(\ref{OrtFun1})\end{tabular}} & ${\alpha}_o$       & $15$                         \\
                                                                                             & $\bm p_g$       & $[-0.28, 0.63, 0.66, 0.28]^T$                         \\
                                                                                            & $k_o$ & $1$ \\ \hline
\multirow{3}{*}{\begin{tabular}[c]{@{}c@{}}\textbf{Vision Region}\\ (\ref{eq_vision_re_func})(\ref{eq_vision_p})\end{tabular}}                & $\bm x_d$ (target object)  & by detection                      \\
                                                                                             & $b_1, b_2$     & $[1440,1080]$                         \\
                                                                                             & $k_v$       & $0.3$                         \\ \hline
\begin{tabular}[c]{@{}c@{}}\textbf{Human Control Input}\\ (\ref{globalControl})\end{tabular}   & $c_{d}$ & $3$                         \\ \hline
\multirow{4}{*}{\begin{tabular}[c]{@{}c@{}}\textbf{Adaptive NN}\\ (\ref{eq_rbf}),(\ref{approxNN_vec}),(\ref{update})\end{tabular}} & $\bm c_i$   &$[0.05,0.35]^T+[0.15j,0.15j]^T$ for $i,j=0,1,2$ \\
                                                                                             & $\sigma$    & 0.1                         \\
                                                                                             & $\bm L$  & $0.25\bm I_9$ \\
                                                                                             & $\hat{\bm W}_{t=t_0}$       & calculated according to $\hat{\bm J}_s(\bm r)_{t=0}$          \\ \hline
\end{tabular}
\end{table*}
\begin{align}
&\hat{\bm J}_s(\bm r)_{t=0}\nonumber\\
&\qquad=\begin{bmatrix}
-1500&0&0&0&-200&60\\
0&2400&170&-200&0&180
\end{bmatrix},
\label{eq_js0}
\end{align}
\begin{equation}
\begin{aligned}
\hat{\bm W}(i,j)_{t=t_0}= \frac{\vectorize(\hat{\bm{J}}_{s}(\bm{r})_{t=0})(i)}{\sum^{n_k}_{k=1}\theta(\bm r_{t=t_0})}, \\
i=1,\cdots,2m; j=1,\cdots,n_k.
\end{aligned}
\label{eq_w0}
\end{equation}

The experimental results are shown in Fig.~\ref{fig:e2_snapshots}. At the beginning, the human expert specified the size and the position of the Cartesian-space region (\ref{CaFun}) via the mixed interface, to ensure that it was within the FOV (see Fig.~\ref{fig:e2_snapshots}a); Then, the robot employed the Cartesian-space feedback to transit from outside to inside the FOV; Subsequently, the vision feedback became available (see Fig.~\ref{fig:e2_snapshots}b) and was used to drive the robot end effector to aim at the target object (see Fig.~\ref{fig:e2_snapshots}c); Finally, the robot grasped the object and moved back 
%
%
to its home position to complete the task (see Fig.~\ref{fig:e2_snapshots}d).

In this experiment, the camera was not calibrated beforehand and hence the exact information about the image Jacobian matrix was unknown. Thus, the adaptive NN was implemented to estimate the image Jacobian matrix via the online update law (\ref{update}). The results with NN adaptation (where $\bm L\not =\bm 0$) or without NN adaptation (by setting $\bm L=\bm 0$) are shown in Fig.~\ref{fig_e2_vision_trajectories}. While both drove the robot to move to the desired position in vision space (see Fig.~\ref{fig_e2_vision_trajectories}a), the control input with NN adaption achieved a faster convergence. The trajectory of the robot end effector in 3D space was shown in Fig. \ref{fig_e2_traj_3d}, proving the smooth transition of the robot among different regions.



\begin{figure} [!tb]
\begin{subfigure}{.525\linewidth}
    \centering
    \includegraphics[height=3.5cm]{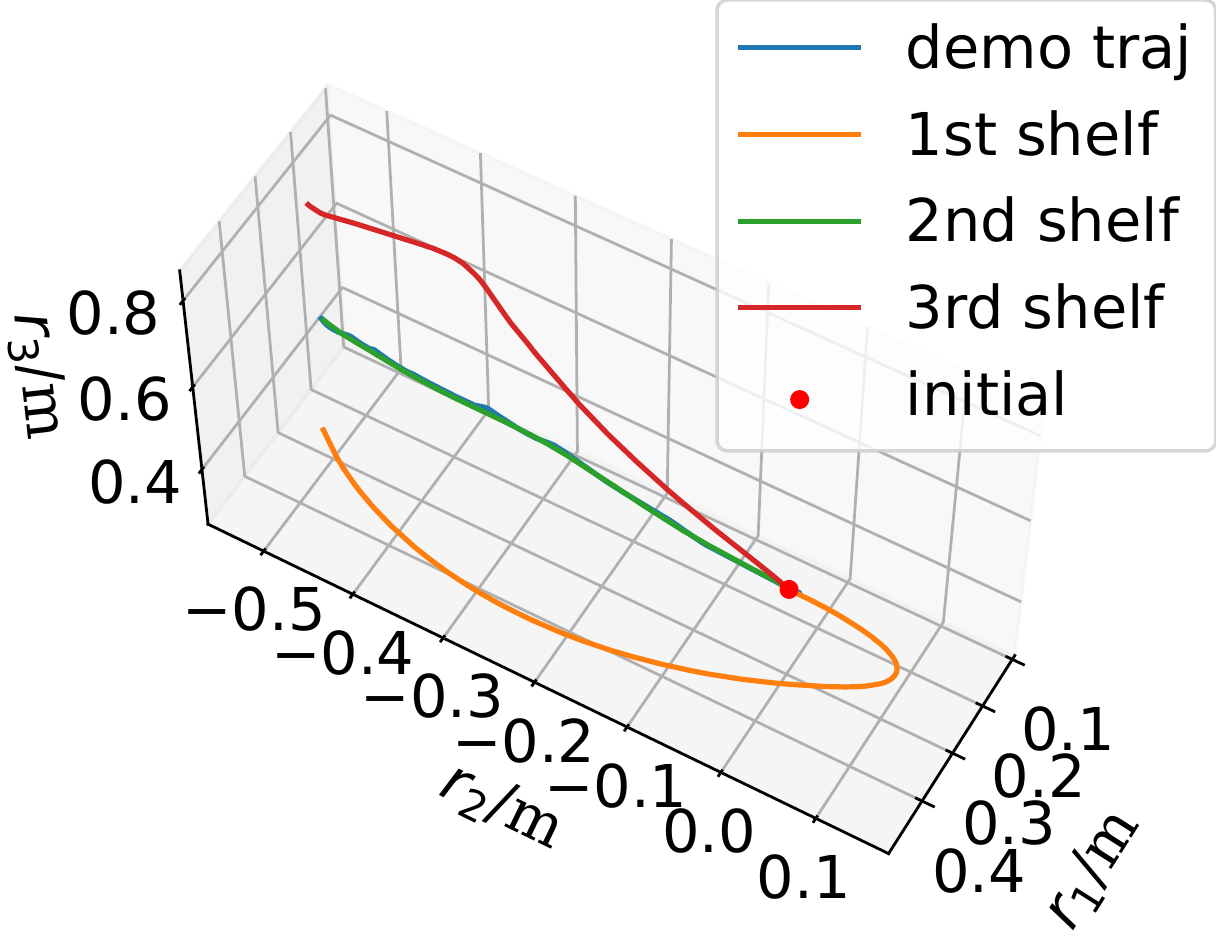}
    \caption{}\label{e3_demo_repro_traj}
  \end{subfigure}
  \hfill
  \begin{subfigure}{.465\linewidth}
    \centering
    \includegraphics[height=3.5cm]{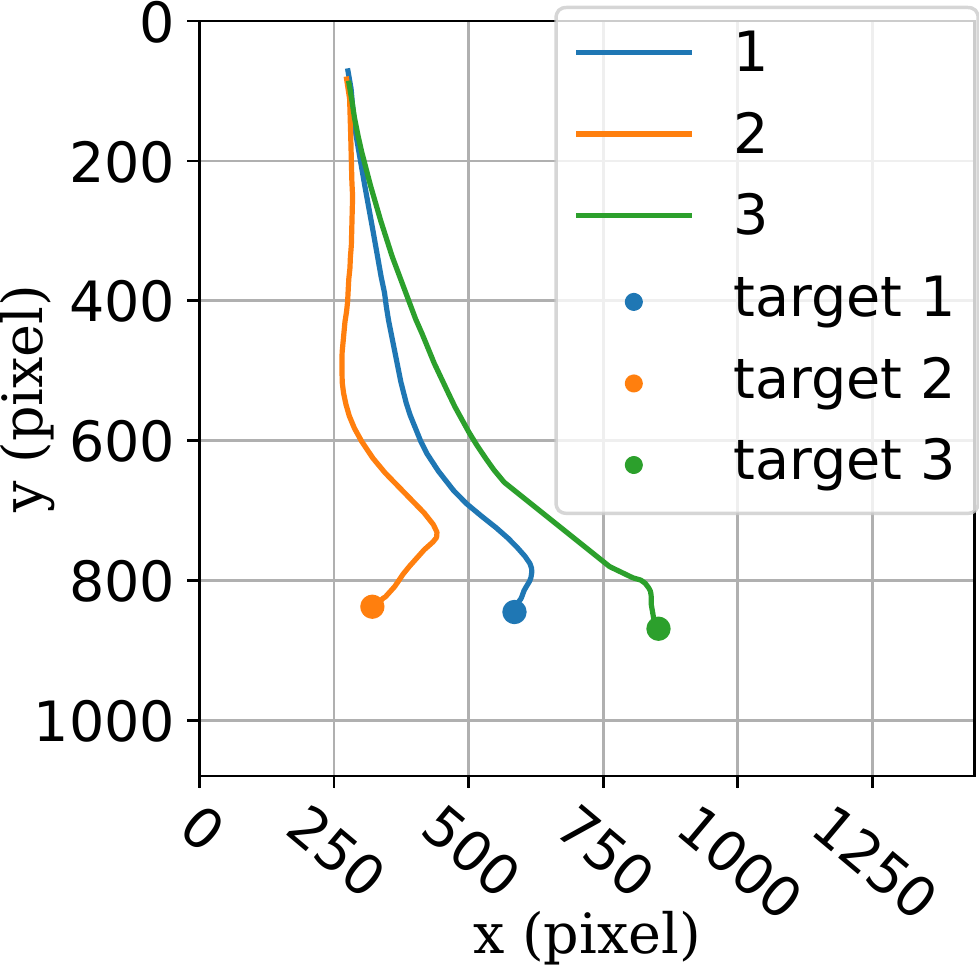}
    \caption{}\label{e3_vision_traj}
  \end{subfigure}
  \caption{Experiment 3 - Results: (a) The trajectories demonstrated by human experts and reproduced with DMP. (b) The path of the ArUco marker (i.e., the feature of target object) in vision space. The initial position was at $x_2>0$ as the marker was occluded at the beginning.}
  \label{fig:e3_vision_trajectory}
\end{figure}
\begin{figure} [!tb]
  \begin{subfigure}{.495\linewidth}
    \centering
    \includegraphics[width=\linewidth]{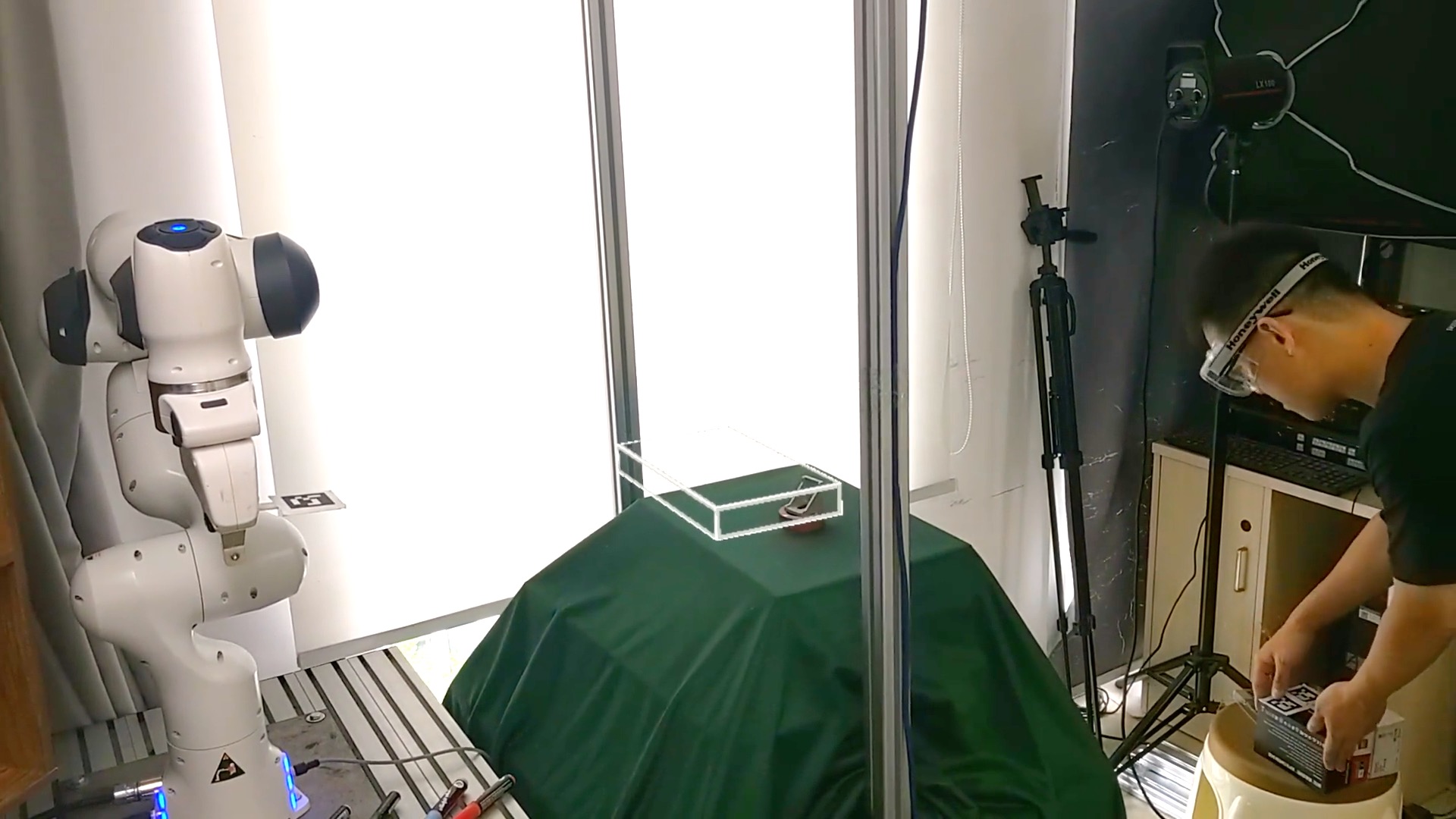}
    \caption{}
  \end{subfigure}
  \hfill
  \begin{subfigure}{.495\linewidth}
    \centering
    \includegraphics[width=\linewidth]{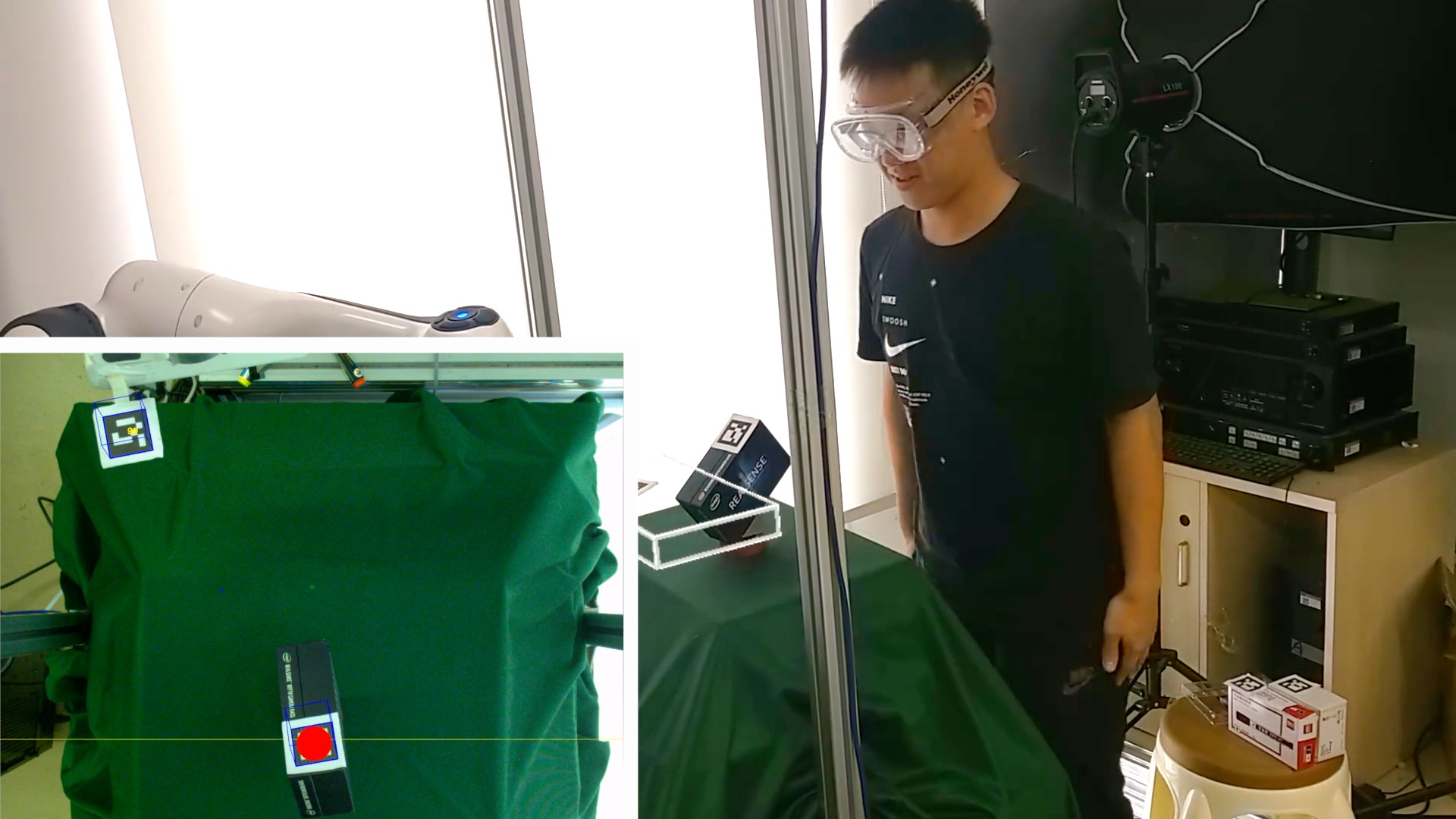}
    \caption{}
  \end{subfigure}
  
  \medskip
  \begin{subfigure}{.495\linewidth}
    \centering
    \includegraphics[width=\linewidth]{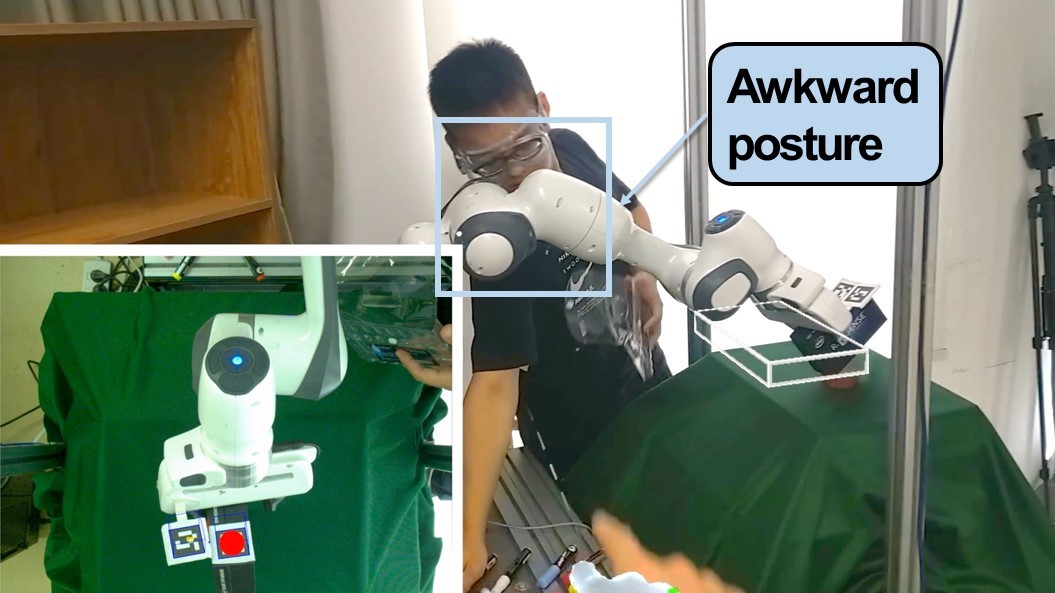}
    \caption{}
  \end{subfigure}
  \hfill
  \begin{subfigure}{.495\linewidth}
    \centering
    \includegraphics[width=\linewidth]{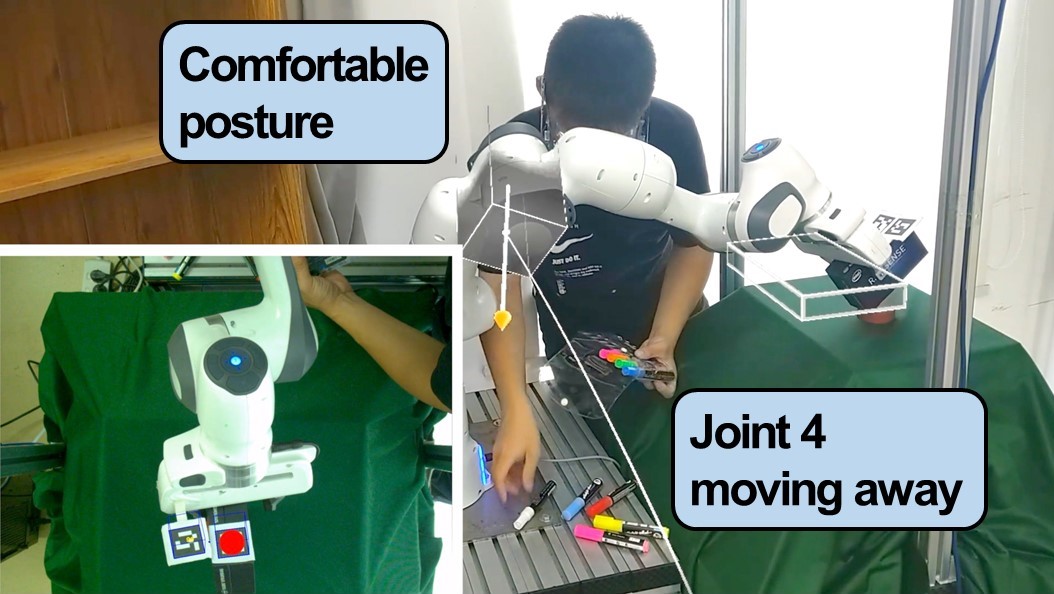}
    \caption{}
  \end{subfigure}
  
  \medskip
  \begin{subfigure}{.495\linewidth}
    \centering
    \includegraphics[width=\linewidth]{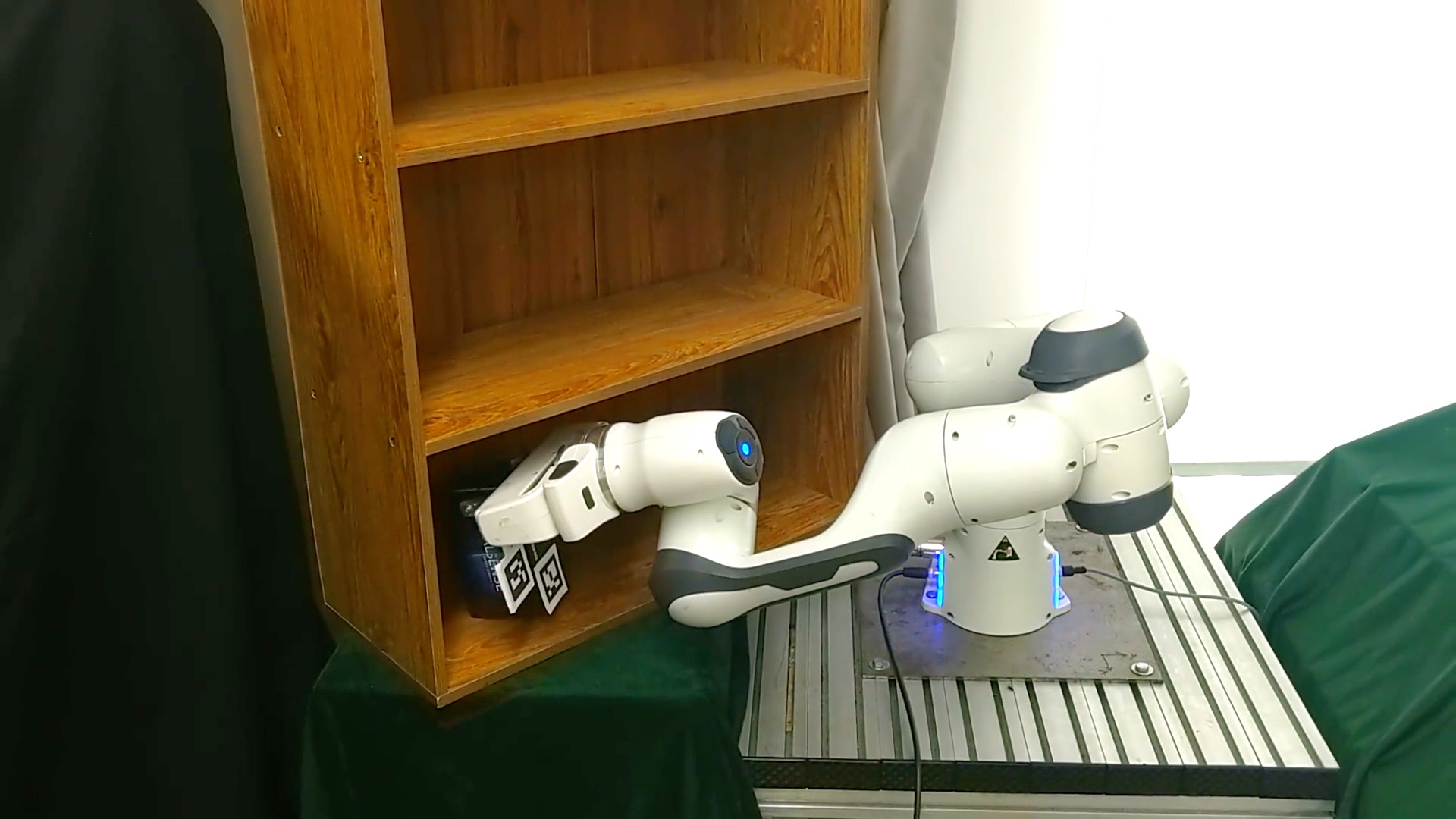}
    \caption{}
  \end{subfigure}
  \hfill
  \begin{subfigure}{.495\linewidth}
    \centering
    \includegraphics[width=\linewidth]{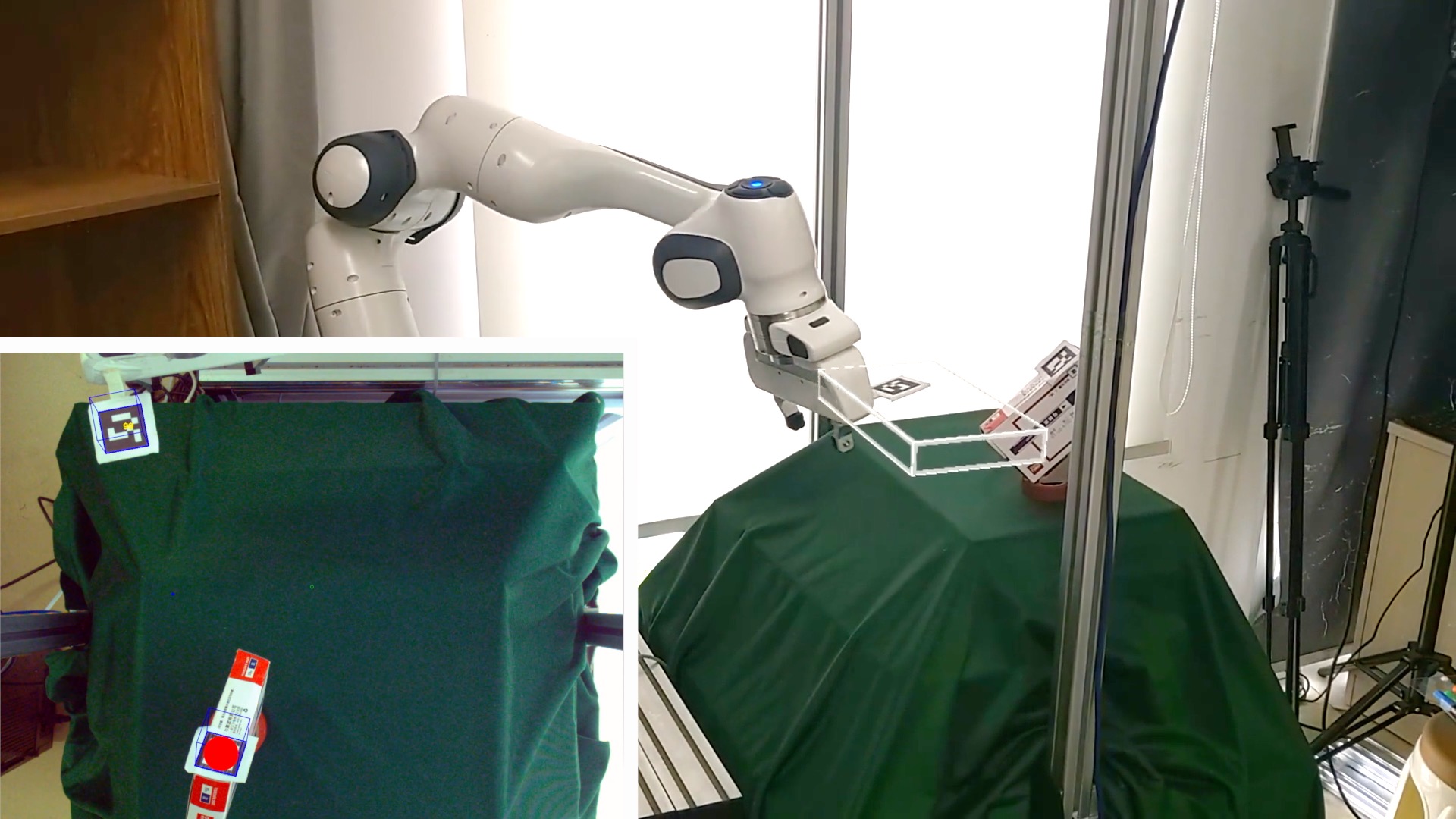}
    \caption{}
  \end{subfigure}
  \caption{Experiment 3 - Snapshots: (a) \(t=0.0\mathrm{s}\): the initial configuration; (b) \(t=3.4\mathrm{s}\): the worker placed the object on the workbench and the robot moved to grasp the object; (c) \(t=15.8\mathrm{s}\): the worker had to collect markers from the table in an awkward position due to obstruction by the robot; (d) \(t=26.3\mathrm{s}\): the expert used the mixed interface to drag the 4th joint of the robot away from the worker without affecting the main task of the robot end effector, enabling the worker to adopt a confortable position; (e) \(t=68.4\mathrm{s}\): the robot transferred the object and placed it on a shelf; (f) \(t=80.1\mathrm{s}\): the worker placed another object at a different position on the workbench, and the robot moved to grasp and then place the object again.}
  \label{fig:e3_snapshots}
\end{figure}
\begin{figure} [!tb]
  \centering 
    \includegraphics[width=\linewidth]{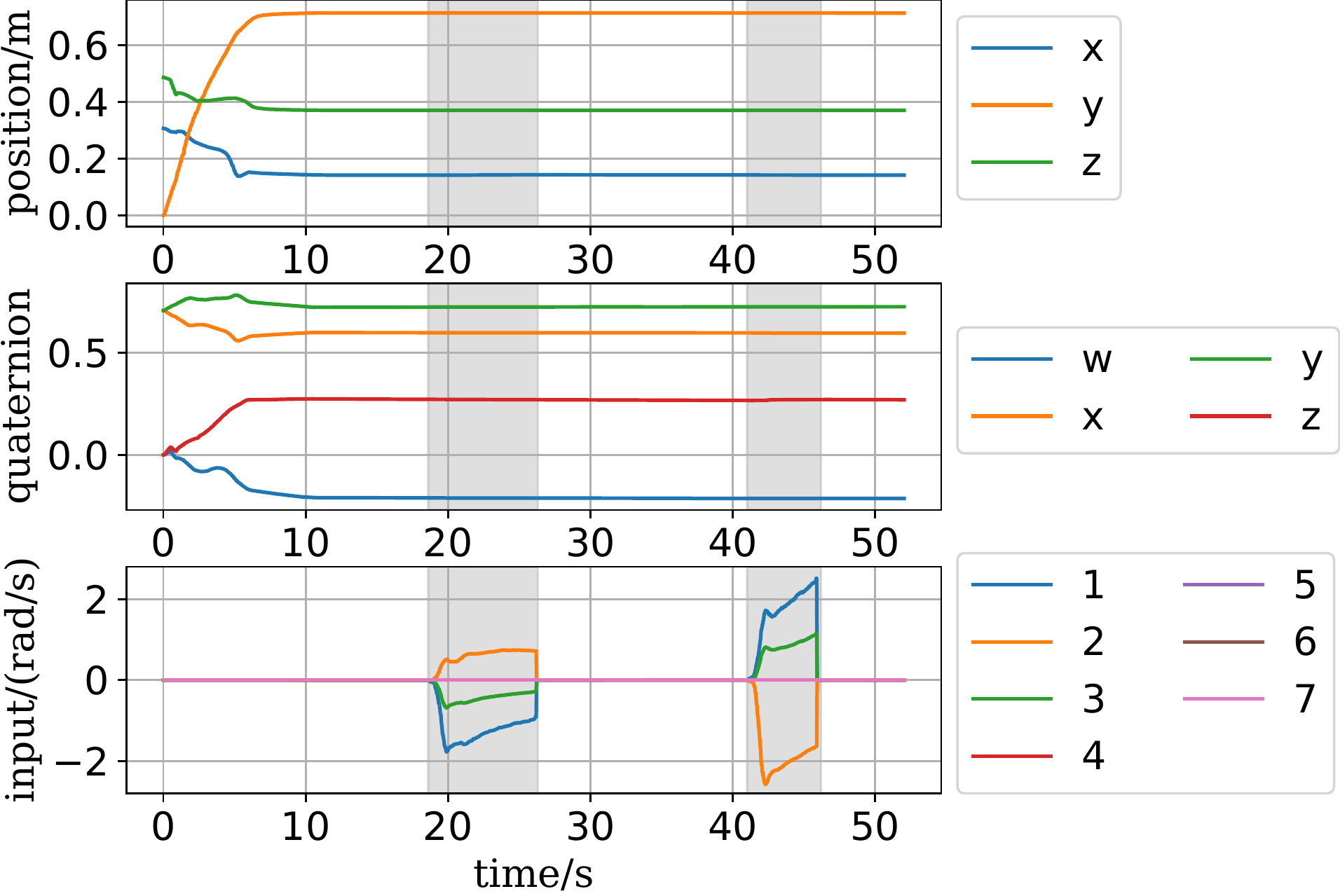}
  \caption{Experiment 3 - The translation (top) and orientation (middle) of the robot end effector, and the human control efforts (bottom). The shaded area denotes the period when the human expert was exerting control efforts.}
  \label{fig:e3_log_d}
\end{figure}

\subsection{Collaboration Task}
In Experiment 3, the robot conducted both grasping and placing tasks. The robot transferred an object from an interactive environment to an isolated environment while also collaborating with the human expert in the interactive environment. Such a mixed scenario is common in factories. For example, a worker hands over objects to a robot, which then transfers the objects to an unmanned laboratory where potentially hazardous experiments are conducted. 

To fulfil the requirements, both the proposed learning scheme and the global adaptive controller were implemented in the robot and activated in different environment (i.e., the isolated and interactive environments, as illustrated in Fig.~\ref{mixedScenario}).  
%
%
The whole task of grasping-placing was performed three times in succession, i.e., three objects (with random initial positions within the FOV) were transferred to different shelves in the cabinet (see Fig.~\ref{fig_marker}). 

For grasping task, the positions of target objects were detected in the vision space and then set as the desired positions for the controller, as shown in Fig.~\ref{e3_vision_traj}. 
In addition, the desired orientation could be approximated by referring to the tangential direction of each trajectory in the neighborhood of the target object. Every time a new object was to be grasped, the image Jacobian matrix was varied but well estimated with the adaptive NN 
(\ref{update}).  

For placing task, the robot followed the learnt trajectories in Experiment 1. The trajectories had different goals corresponding to multiple desired positions on different shelves. The trajectories also consisted of similar paths towards the goal positions, which avoids collisions between the robot body and the shelves whenever a new object was being placed.  
The experimental results are shown in Fig.~\ref{fig:e3_vision_trajectory}, which confirms that the consecutive grasping-placing was successfully realized. 

Snapshots of the experiment are shown in Fig.~\ref{fig:e3_snapshots}. At $t\hspace{-0.05cm}=\hspace{-0.05cm}0\sim 10~
\mathrm{s}$, the robot moved to the desired pose for grasping. Meanwhile, the worker approached the robot and attempted to collect the scattered markers. As the markers were under the robot, the worker had to conduct the task 
in an awkward position.  
At $t\hspace{-0.05cm}=\hspace{-0.05cm}19\sim 26~\mathrm{s}$ and $t\hspace{-0.05cm}=\hspace{-0.05cm}42\sim 46~
\mathrm{s}$, the human expert observed the situations and dragged the \(4\)th joint of the virtual robot forward and backward to adjust the body shape of the real robot, which guaranteed the worker's safety and comfort. The control efforts $\bm d$ exerted by the expert are shown in Fig.~\ref{fig:e3_log_d}, which were projected into the null space; thus, they did not affect the main task (translation and orientation) of the robot end effector. Hence, the collaboration between the human expert and the robot was efficient. During the process, human expertise of the unstructured environment powerfully complemented the robot's large-scale transition ability, which was implemented with the vision-based global adaptive controller.

All the experimental results can be found at \url{https://youtu.be/JCZwo0fCbeg}. Specifically, the uploaded video also shows both the robot and the worker can perform tasks simultaneously in the shared space, without affecting each other. 

\section{Conclusions}
This paper develops a new framework for human-robot collaboration, where the main novelty is its complementary feature. This enables a human expert and a robot to collaborate in a more efficient way. Specifically, a new vision-based adaptive controller is proposed for the robot to ensure the global convergence of the end effector, in the presence of joint limits, uncalibrated camera, and limited FOV; A mixed AR-haptic interface is developed to allow the expert to perform demonstration in both task space and redundant joint space and perform collaboration to deal with unforeseen changes (e.g., suddenly appearing human walkers), without affecting the main task. Therefore, the proposed framework enables the robot to safely interact with other co-existing workers, in parallel to its ongoing works, and it also provides a natural and intuitive way for the expert to deliver his/her knowledge and smart decision.
The global stability of closed-loop system is rigorously proved with Lyapunov methods, and the performance of the proposed scheme is validated in a series of transferring tasks in the hybrid environment (i.e., interactive and isolated). 
Future works will be devoted to the marker-free perception and the field application in factories.


\section*{Appendix} 
\subsection{Stability Analysis}
Multiplying both sides of (\ref{closed}) by $\bm N(\bm q)$ and noting that $\bm N(\bm q)\bm J^+(\bm q)=\bm 0$ and $\bm N^2(\bm q)=\bm N(\bm q)$, we have
\begin{equation}
\bm N(\bm q)\dot{\bm q}=\bm N(\bm q)c_d^{-1}(\bm d - \bm \xi_q),
\end{equation}
that is,
\begin{equation}
\bm N(\bm q)(c_d\dot{\bm q}-\bm d-\bm \xi_q)=\bm 0,\label{temp77}
\end{equation}
which maps the desired damping model $c_d\dot{\bm q}=\bm d - \bm \xi_q$ into the null space of the Jacobian matrix, such that both the expert's control efforts $\bm d$ and the joint-space regional feedback $\bm \xi_q$ work without affecting the robot end effector.

Note that $\bm d$ and $\bm \xi_q$ do not usually work at the same time:
\begin{enumerate}
    \item[-] When the robot leaves the joint-space region and hence stays away from joint limits or singularity, \(\bm \xi_q=\bm 0\), (\ref{temp77}) becomes \(\bm N(\bm q)(c_d\dot{\bm q})=\bm N(\bm q)\bm d\), such that the motion of redundant joints is solely determined by the expert.
    
    \item[-] When the robot is inside the joint-space region and the expert does not input control efforts, \(\bm d=\bm 0\), (\ref{temp77}) becomes \(\bm N(\bm q)(c_d\dot{\bm q})=\bm N(\bm q)\bm \xi_q\). 
    Although it is possible that \(\bm \xi_q\neq\bm 0\) but \(\bm N(\bm q)\bm \xi_q=\bm 0 \), those cases are very rare;
    Because when the robot is not exactly located at the singular configuration, \(rank(\bm N(\bm q)) =1\), \( \bm N(\bm q)= \bm a \bm b^T\), where \(\bm a, \bm b \in\Re^{7} \) are vectors; That is, the aforementioned cases occur only when \(\bm \xi_q \) is exactly orthogonal to \(\bm b\), which is very rare in actual implementation; Hence, the robot will not stay inside the joint-space region (i.e., \(\bm \xi_q\neq\bm 0\) and \(\dot{\bm q}=\bm 0\)) and will leave it by the end, i.e., \(\bm \xi_q=\bm 0\). 
\end{enumerate}
Hence, the control objective in null space is realized.

Multiplying both sides of (\ref{closed}) by $\bm J(\bm q)$ and noting that $\bm J(\bm q)\dot{\bm q}=\dot{\bm r}$ and $\bm J(\bm q)\bm N(\bm q)=\bm 0$, it is obtained that 
\begin{equation}
\dot{\bm r}=-\hat{\bm J}_s^T(\bm r)\bm\xi_x-\bm \xi_r.\label{taskClosed}
\end{equation}

To prove the stability of the closed-loop system in task space,
a Lyapunov-like candidate is proposed as
\begin{equation}
V = P_v(\bm x)+P_c(\bm r)+\tr(\Tilde{\bm W}\bm L^{-1}\Tilde{\bm W}^T), \label{vt}    
\end{equation}
where \(\Tilde{\bm W}\hspace{-0.05cm}=\hspace{-0.05cm}\bm W-\hat{\bm W}\) is the approximation error.

Differentiating (\ref{vt}) with respect to time yields
\begin{align}
\dot V &= \dot{\bm x}^T\frac{\partial P_v(\bm x)}{\partial \bm x}+\dot{\bm r}^T\frac{\partial P_c(\bm r)}{\partial \bm r}-\tr(\Tilde{\bm W}\bm L^{-1}\dot{\hat{\bm W}}^T) \nonumber\\
&= \dot{\bm x}^T\bm\xi_x+\dot{\bm r}^T\bm\xi_r - \tr(\Tilde{\bm W}\bm L^{-1}\dot{\hat{\bm W}}^T) \nonumber\\
&= \dot{\bm r}^T(\bm J_s^T(\bm r)\bm\xi_x+\bm\xi_r)-\tr(\Tilde{\bm W}\bm L^{-1}\dot{\hat{\bm W}}^T).\label{adaLya}
\end{align}

Substituting (\ref{taskClosed}) into (\ref{adaLya}), it is obtained that
\begin{align}
\dot V &= -(\hat{\bm J}_s^T(\bm r)\bm\xi_x+\bm \xi_r)^T \times(\bm J_s^T(\bm r)\bm\xi_x+\bm\xi_r)\nonumber\\
&\phantom{{}={}}\negmedspace- \tr(\Tilde{\bm W}\bm L^{-1}\dot{\hat{\bm W}}^T)\nonumber\\
&= -(\hat{\bm J}_s^T(\bm r)\bm\xi_x+\bm \xi_r)^T\times((\hat{\bm J}_s(\bm r)+\Tilde{\bm J}_s(\bm r))^T\bm\xi_x+\bm\xi_r)\nonumber\\
&\phantom{{}={}}\negmedspace -\tr(\Tilde{\bm W}\bm L^{-1}\dot{\hat{\bm W}}^T)\nonumber\\
&= -(\hat{\bm J}_s^T(\bm r)\bm\xi_x+\bm \xi_r)^T\times(\hat{\bm J}_s^T(\bm r)\bm\xi_x+\bm\xi_r)\nonumber\\
&\phantom{{}={}}\negmedspace -(\hat{\bm J}_s^T(\bm r)\bm\xi_x+\bm \xi_r)^T\Tilde{\bm J}_s^T(\bm r)\bm\xi_x-\tr(\Tilde{\bm W}\bm L^{-1}\dot{\hat{\bm W}}^T), \label{adaLya1}
\end{align}
where $\Tilde{\bm J}_s^T(\bm r)\triangleq\bm J_s^T(\bm r)-\hat{\bm J}_s^T(\bm r)$. 
Making use of (\ref{approxNN_vec}) and (\ref{reformulation}), it is clear that
\begin{align}
&-(\hat{\bm J}_s^T(\bm r)\bm\xi_x+\bm \xi_r)^T\Tilde{\bm J}_s^T(\bm r)\bm\xi_x\nonumber\\
&\qquad = -\tr[\Tilde{\bm J}_s^T(\bm r)\bm\xi_x(\hat{\bm J}_s^T(\bm r)\bm\xi_x+\bm \xi_r)^T]\nonumber\\
&\qquad = -\tr[\bm\xi_{x}^{\prime} \vectorize(\Tilde{\bm J}_{s}^T(\bm r)) (\hat{\bm J}_s^T(\bm r)\bm\xi_x+\bm \xi_r)^T]\nonumber\\
&\qquad = -\tr[\bm\xi_x^{\prime}\Tilde{\bm W}\bm\theta(\bm r)
(\hat{\bm J}_s^T(\bm r)\bm\xi_x+\bm \xi_r)^T]. \label{last2}
\end{align}  

Substituting the update law (\ref{update}) into the last term of (\ref{adaLya1}), it is obtained that
\begin{align}
&-\tr(\Tilde{\bm W}\bm L^{-1}\dot{\hat{\bm W}}^T) \nonumber\\
&\qquad = \tr[\Tilde{\bm W}\bm\theta(\bm r)
(\hat{\bm J}_s^T(\bm r)\bm\xi_x+\bm \xi_r)^T\bm\xi_x^{\prime}] \nonumber\\
&\qquad = \tr[\bm\xi_x^{\prime}\Tilde{\bm W}\bm\theta(\bm r)
(\hat{\bm J}_s^T(\bm r)\bm\xi_x+\bm \xi_r)^T]. \label{last1}
\end{align}

With (\ref{last2}) and (\ref{last1}), the last two terms in (\ref{adaLya1}) can be cancelled such that
\begin{align}
\dot V &= -(\hat{\bm J}_s^T(\bm r)\bm\xi_x+\bm \xi_r)^T\times(\hat{\bm J}_s^T(\bm r)\bm\xi_x+\bm\xi_r)\leq 0 .\label{adaLya2}
\end{align}

Since $V\hspace{-0.05cm}>\hspace{-0.05cm}0$ and $\dot V\hspace{-0.05cm}\leq\hspace{-0.05cm}0$, $V$ is bounded, and the closed-loop system is stable. The boundedness of $V$ ensures the boundedness of \(P_v(\bm x),\ P_c(\bm r)\), and \(\Tilde{\bm W}\). Hence, all the regional feedback vectors \(\bm\xi_x, \bm\xi_r\) are bounded. From (\ref{taskClosed}), it can be seen that $\dot{\bm r}$ is bounded, which also ensures the boundedness of $\dot{\bm x}$ and $\dot{\bm q}$. Hence, the term $(\hat{\bm J}_s^T(\bm r)\bm\xi_x+\bm\xi_r)$ is uniformly continuous. From (\ref{adaLya2}), it follows that $(\hat{\bm J}_s^T(\bm r)\bm\xi_x\hspace{-0.05cm}+\hspace{-0.05cm}\bm\xi_r\hspace{-0.05cm})\hspace{-0.05cm}\in\hspace{-0.05cm}L_2(0, +\infty)$. Therefore, we have $(\hat{\bm J}_s^T(\bm r)\bm\xi_x\hspace{-0.05cm}+\hspace{-0.05cm}\bm\xi_r\hspace{-0.05cm})\hspace{-0.05cm}\rightarrow\hspace{-0.05cm}\bm 0$.

When the regional feedback vector $\bm\xi_q$ keeps the robot away from the joint limits, $\bm \xi_q=\bm 0$. Then, the regional feedback vector $\bm\xi_r$ drives the robot to move from outside to inside the FOV. After it enters the FOV, the regional feedback vector $\bm\xi_x$ is activated and it reduces to zero only when the robot has reached the desired position (to grasp the target object). Hence, the convergence of $(\hat{\bm J}_s^T(\bm r)\bm\xi_x\hspace{-0.05cm}+\hspace{-0.05cm}\bm\xi_r\hspace{-0.05cm})\hspace{-0.05cm}\rightarrow\hspace{-0.05cm}\bm 0$ actually implies that the grasping task is realized, in the presence of limited FOV and uncalibrated camera. 

\subsection{Orientation Region}
As the analytical Jacobian is derived by differentiating the forward kinematic equations, $\bm{\xi}_r$ can be obtained through pure differential operations. Representing the Cartesian configuration $\bm{r}$ as $\begin{bmatrix}\bm{r}_t^T\ \bm{r}_o^T\end{bmatrix}^T$, which consists of the translation part $\bm{r}_t=\begin{bmatrix}x\ y\ z\end{bmatrix}^T$ and the orientation part $\bm{r}_o=\begin{bmatrix}r_{ox}\ r_{oy}\ r_{oz}\end{bmatrix}^T=\begin{bmatrix}{\phi\hat{n}_x}\ {\phi\hat{n}_y}\ {\phi\hat{n}_z}\end{bmatrix}^T=\phi\hat{\bm{n}}$. Note that $\bm{r}_o$ is given in rotation vector form, which represents the axis of rotation $\hat{\bm{n}}=\begin{bmatrix}{\hat{n}_x}\ {\hat{n}_y}\ {\hat{n}_z}\end{bmatrix}^T$ and the angle of rotation $\phi$. Then, the regional feedback vector in Cartesian space from (\ref{kesi_r_origin}) is rewritten as
\begin{equation}
\bm \xi_r\triangleq
\begin{bmatrix}
    \dfrac{\partial P_t(\bm r)}{\partial \bm {r}_t}\\
    \dfrac{\partial P_o(\bm r)}{\partial \bm {r}_o}
\end{bmatrix}.\label{kesi_r_redefine}
\end{equation}
Applying the chain rule and we have
\begin{equation}
\frac{\partial P_o(\bm r)}{\partial \bm {r}_o^T}=\frac{\partial P_o(\bm r)}{\partial \bm {p}^T}\cdot\frac{\partial \bm p}{\partial \bm {r}_o^T}.\label{chainRule}
\end{equation}

Assuming that the unit quaternions $\bm p=(v_o,\bm{u}_o)$, $\bm p_g^{-1}=(v_g,\bm{u}_g)$, and their product $\bm p\ast{\bm p_g}^{-1}=(v_{e},\bm{u}_{e})$ in the orientation region function (\ref{OrtRegion}). According to the quaternion product rule, we can compute the real part of the orientation error as  
\begin{equation}
v_{e}=v_o\cdot{v_g}-\bm{u}_o^T\bm{u}_g.\label{quatProduct}
\end{equation}
Considering the constraint $v_o^2+\lVert{\bm{u}_o}\rVert^2\equiv{1}$ for the unit quaternion $\bm p=(v_o,\bm{u}_o)$, where $\bm{u}_o=(u_{ox},u_{oy},u_{oz})$, the partial derivative of $v_{e}$ with respect to $\bm p$ is derived as
\begin{align}
\frac{\partial v_{e}}{\partial v_o} &= v_g+v_o\left(\frac{u_{gx}}{u_{ox}}+\frac{u_{gy}}{u_{oy}}+\frac{u_{gz}}{u_{oz}}\right)\nonumber \\
\frac{\partial v_{e}}{\partial u_{ox}} &= -u_{gx}+u_{ox}\left(-\frac{v_{g}}{v_{o}}+\frac{u_{gy}}{u_{oy}}+\frac{u_{gz}}{u_{oz}}\right)\nonumber \\
\frac{\partial v_{e}}{\partial u_{oy}} &= -u_{gy}+u_{oy}\left(-\frac{v_{g}}{v_{o}}+\frac{u_{gx}}{u_{ox}}+\frac{u_{gz}}{u_{oz}}\right)\nonumber \\
\frac{\partial v_{e}}{\partial u_{oz}} &= -u_{gz}+u_{oz}\left(-\frac{v_{g}}{v_{o}}+\frac{u_{gx}}{u_{ox}}+\frac{u_{gy}}{u_{oy}}\right).\label{PartialV2Quat}
\end{align}

To obtain a more specific expression of (\ref{OrtRegion}), the unit quaternion logarithmic is defined as
\begin{subnumcases}{\label{quatLog} \log{(\bm p*\bm p_g^{-1})} = }
  \arccos(v_{e})\frac{\bm{u}_{e}}{\lVert{\bm{u}_{e}}\rVert}\,, & \(\bm{u}_{e}\neq 0\)\,, \\
  \left[0,0,0\right]^T\,, & otherwise.
\end{subnumcases}
Substituting (\ref{quatLog}) into (\ref{OrtRegion}), it is obtained that
\begin{equation}
{f}_o(\bm{r})={\alpha}_o\cdot\arccos(v_{e})\cdot{\mathbb{I}_{\lVert{\bm{u}_{e}}\rVert>0}}-1.\label{OrtRegion2}
\end{equation}
where ${\mathbb{I}_{\lVert{\bm{u}_{e}}\rVert>0}}$ is the indicator function introduced to prevent dividing by zero in (\ref{quatLog}).
\begin{subnumcases}{\mathbb{I}_{\lVert{\bm{u}_{e}}\rVert>0}=}
  1\,, & \(\lVert{\bm{u}_{e}}\rVert>0\)\,, \\
  0\,, & otherwise.
\end{subnumcases}
Computing the partial derivative of (\ref{OrtRegion2}) with respect to quaternion $\bm p$ and we have
\begin{align}
    \frac{\partial{{f}_o(\bm{r})}}{\partial{\bm p^T}}
    &= -\frac{{\alpha}_o}{\sqrt{1-(v_{e})^2}}\cdot{\frac{\partial{v_{e}}}{\partial{\bm p^T}}}\cdot{\mathbb{I}_{\lVert{\bm{u}_{e}}\rVert>0}}\nonumber \\
    &= -\frac{{\alpha}_o}{\lVert{\bm{u}_{e}}\rVert}\cdot{\frac{\partial{v_{e}}}{\partial{\bm p^T}}}\cdot{\mathbb{I}_{\lVert{\bm{u}_{e}}\rVert>0}},\label{PartialFoPartialQ}
\end{align}
where $\frac{\partial{v_{e}}}{\partial{\bm p^T}}=(\frac{\partial v_{e}}{\partial v_o},\frac{\partial v_{e}}{\partial u_{ox}},\frac{\partial v_{e}}{\partial u_{oy}},\frac{\partial v_{e}}{\partial u_{oz}})$ is derived from (\ref{PartialV2Quat}). Then according to (\ref{OrtFun1}), we have
\begin{align}
    \frac{\partial P_o(\bm{r})}{\partial\bm p^T} &= {k}_o\max(0,{f}_o(\bm{r}))\cdot\frac{\partial{{f}_o(\bm{r})}}{\partial{\bm p^T}}\nonumber \\
    &= -\frac{{\alpha}_o{k}_o}{\lVert{\bm{u}_{e}}\rVert}\cdot{\max(0,{f}_o(\bm{r}))}\cdot{\frac{\partial{v_{e}}}{\partial{\bm p^T}}}\cdot{\mathbb{I}_{\lVert{\bm{u}_{e}}\rVert>0}}.\label{PartialPoPartialQ}
\end{align}

The conversion from the unit quaternion to the rotation vector form which represents the same orientation is defined as
\begin{align}
    \bm p &= (\underbrace{p_w}_{v_e},\underbrace{(p_x,p_y,p_z)}_{\bm{u}_e}) = \biggl(\cos\frac{\phi}{2},\sin\frac{\phi}{2}\cdot\underbrace{(\hat{n}_x,\hat{n}_y,\hat{n}_z)}_{\hat{\bm{n}}}\biggr), \label{CvtQuat2AngAxis}
\end{align}
where $\phi=\lVert{\bm{r}_o}\rVert$ and $\hat{\bm{n}}=\frac{\bm{r}_o}{\phi}$. By collecting first partial derivatives of the quaternion $\bm p$ with respect to the rotation vector $\bm{r}_o$, we obtain the Jacobian matrix as $\bm{J}_{rot}=\begin{bmatrix}\bm {J}_{rot, ij}\end{bmatrix}=\begin{bmatrix}\frac{\partial {p}_i}{\partial {r}_{oj}}\end{bmatrix}\in{\Re^{4 \times 3}}$. The Jacobian matrix $\bm {J}_{rot}$ is defined as
\begin{equation}
\bm {J}_{rot}\triangleq
\begin{bmatrix}
    A_x & A_y & A_z\\
    B_{xyz} & C_{xy} & C_{xz}\\
    C_{yx} & B_{yzx} & C_{yz}\\
    C_{zx} & C_{zy} & B_{zxy}    
\end{bmatrix}
= \frac{\partial \bm p}{\partial \bm {r}_o^T}.\label{JacobianQuatAngAxis}
\end{equation}

There are three types of elements in the matrix above, which are represented as
\begin{eqnarray}
\left\{
\begin{aligned}
    &A_i = -\frac{r_{oi}}{2\lVert{\bm{r}_o}\rVert}\sin\frac{\lVert{\bm{r}_o}\rVert}{2} \\
    &B_{ijk} = \frac{r_{oi}^2}{2\lVert{\bm{r}_o}\rVert^2}\cos\frac{\lVert{\bm{r}_o}\rVert}{2}+\frac{r_{oj}^2+r_{ok}^2}{\lVert{\bm{r}_o}\rVert^3}\sin\frac{\lVert{\bm{r}_o}\rVert}{2} \\
    &C_{ij} = \frac{r_{oi}r_{oj}}{2\lVert{\bm{r}_o}\rVert^2}\cos\frac{\lVert{\bm{r}_o}\rVert}{2}-\frac{r_{oi}r_{oj}}{\lVert{\bm{r}_o}\rVert^3}\sin\frac{\lVert{\bm{r}_o}\rVert}{2}
\end{aligned}
\right.,\label{JacobianQuatAngAxisDetailed}
\end{eqnarray}
where the subscripts are selected as follows
\begin{eqnarray}
\left\{
\begin{aligned}
    &A_i,\ \ {i}\in\{x,y,z\} \\
    &B_{ijk},\ \ {ijk}\in\{xyz,yzx,zxy\} \\
    &C_{ij},\ \ {ij}\in\{xy,yx,xz,zx,yz,zy\}
\end{aligned}
\right..\label{SubscriptSelection}
\end{eqnarray}

With (\ref{PartialPoPartialQ}), (\ref{JacobianQuatAngAxis}) and (\ref{JacobianQuatAngAxisDetailed}), we can obtain the regional feedback vector through (\ref{chainRule}). Note that if other orientation representations are chosen as $\bm{r}_o$, such as the Euler angles, $\bm{\xi}_r$ can be derived in a similar manner as (\ref{kesi_r_redefine}) and (\ref{chainRule}).

Nevertheless, when the geometric Jacobian is used in lieu of the analytic Jacobian, the differential operations mentioned above become invalid. Refering to the translation part of the differential potential energy
\begin{equation}
    \frac{\partial P_t(\bm r)}{\partial \bm {r}_t}=2\bm{k}_c\odot\max(0,\bm{f}_c(\bm{r}_t))\odot(\Delta\bm{r}_t\oslash\bm{c}^2),\label{trans_diff_potential}
\end{equation}
where $\Delta\bm{r}_t={\bm{r}_t-\bm{r}_c}$, $\bm{c}=\begin{bmatrix}c_1\ c_2\ c_3\end{bmatrix}^T$, and $\bm{k}=\begin{bmatrix}k_{c1}\ k_{c2}\ k_{c3}\end{bmatrix}^T$ $\in{\Re^{3}}$ are column vectors, and $\odot$, $\oslash$ represent the element-wise multiplication and division, respectively. We formally define the orientation part of differential potential energy as
\begin{equation}
    \frac{\partial P_o(\bm{r})}{\partial \bm{r}_o}=\alpha_{o}k_o\max(0,f_o(\bm{r}_o))\bm{r}_e.\label{orient_diff_potential}
\end{equation}
In (\ref{orient_diff_potential}), $\bm{r}_e$ denotes the rotation vector form of orientation error $\bm p\ast{\bm p_g}^{-1}$ and resembles the term $\Delta\bm{r}_t$ in (\ref{trans_diff_potential}), which are, respectively, equivalent to the angular and linear velocity required to align the desired frame with the end effector frame in unit time. The differences between (\ref{trans_diff_potential}) and (\ref{orient_diff_potential}) are as follows:
\begin{enumerate}
    \item [-] The region function is a vector $\bm{f}_c$ in (\ref{trans_diff_potential}), while it is a scalar $f_o$ in (\ref{orient_diff_potential}). This is due to the coupling effect of orientation representation.
    \item [-] In (\ref{orient_diff_potential}), the scaling factor $\alpha_{o}k_o$ is consistent for each dimension. However, in (\ref{trans_diff_potential}), the vectorized term $2\bm{k}_c\oslash\bm{c}^2$ may lead to different convergence speeds in different dimensions of $\bm{r}_t$. Thus, $\bm{k}_c$ must be chosen carefully.
\end{enumerate}

Generally, calculating $\bm{\xi}_r$ through (\ref{orient_diff_potential}) is less computationally intensive than through (\ref{chainRule}). In addition, the introduction of geometric Jacobian prevents the occurrence of representation singularities.

\bibliography{main}

\begin{thebibliography}{10}

\bibitem{weiss_21}
A.~Weiss, A.-K. Wortmeier, and B.~Kubicek, ``Cobots in industry 4.0: A roadmap
  for future practice studies on human--robot collaboration,'' {\em IEEE
  Transactions on Human-Machine Systems}, vol.~51, no.~4, pp.~335--345, 2021.

\bibitem{tcst16_li}
X.~Li, G.~Chi, S.~Vidas, and C.~C. Cheah, ``Human-guided robotic
  comanipulation: Two illustrative scenarios,'' {\em IEEE Transactions on
  Control Systems Technology}, vol.~24, no.~5, pp.~1751--1763, 2016.

\bibitem{hentout2019human}
A.~Hentout, M.~Aouache, A.~Maoudj, and I.~Akli, ``Human--robot interaction in
  industrial collaborative robotics: a literature review of the decade
  2008--2017,'' {\em Advanced Robotics}, vol.~33, no.~15-16, pp.~764--799,
  2019.

\bibitem{iqbal2017prospects}
J.~Iqbal, Z.~H. Khan, and A.~Khalid, ``Prospects of robotics in food
  industry,'' {\em Food Science and Technology}, vol.~37, pp.~159--165, 2017.

\bibitem{sanderson2019intelligent}
A.~Sanderson, ``Intelligent robotic recycling of flat panel displays,''
  Master's thesis, University of Waterloo, 2019.

\bibitem{icra_22}
X.~Yan, C.~Chen, and X.~Li, ``Adaptive vision-based control of redundant robots
  with null-space interaction for human-robot collaboration,'' in {\em IEEE
  International Conference on Robotics and Automation (ICRA)}, IEEE, 2022.

\bibitem{handbook}
S.~Chiaverini, ``Kinematically redundant manipulators,'' {\em Handbook of
  Robotics}, pp.~245--268, 2008.

\bibitem{tro16_Gosselin}
C.~Gosselin and L.-T. Schreiber, ``Kinematically redundant spatial parallel
  mechanisms for singularity avoidance and large orientational workspace,''
  {\em IEEE Transactions on Robotics}, vol.~32, no.~2, pp.~286--300, 2016.

\bibitem{ijrr17_Carmichael}
M.~G. Carmichael, D.~Liu, and K.~J. Waldron, ``A framework for
  singularity-robust manipulator control during physical human-robot
  interaction,'' {\em The International Journal of Robotics Research}, vol.~36,
  no.~5-7, pp.~861--876, 2017.

\bibitem{tro04_chesi}
G.~Chesi, K.~Hashimoto, D.~Prattichizzo, and A.~Vicino, ``Keeping features in
  the field of view in eye-in-hand visual servoing: A switching approach,''
  {\em IEEE Transactions on Robotics}, vol.~20, no.~5, pp.~908--914, 2004.

\bibitem{tro05_garcia-aracil}
N.~Garc{\'\i}a-Aracil, E.~Malis, R.~Aracil-Santonja, and C.~P{\'e}rez-Vidal,
  ``Continuous visual servoing despite the changes of visibility in image
  features,'' {\em IEEE Transactions on Robotics}, vol.~21, no.~6,
  pp.~1214--1220, 2005.

\bibitem{tro11_gans}
N.~R. Gans, G.~Hu, K.~Nagarajan, and W.~E. Dixon, ``Keeping multiple moving
  targets in the field of view of a mobile camera,'' {\em IEEE Transactions on
  Robotics}, vol.~27, no.~4, pp.~822--828, 2011.

\bibitem{tro19_Bechlioulis}
C.~P. Bechlioulis, S.~Heshmati-Alamdari, G.~C. Karras, and K.~J. Kyriakopoulos,
  ``Robust image-based visual servoing with prescribed performance under field
  of view constraints,'' {\em IEEE Transactions on Robotics}, vol.~35, no.~4,
  pp.~1063--1070, 2019.

\bibitem{li2022hybrid}
T.~Li, J.~Yu, Q.~Qiu, and C.~Zhao, ``Hybrid uncalibrated visual servoing
  control of harvesting robots with rgb-d cameras,'' {\em IEEE Transactions on
  Industrial Electronics}, 2022.

\bibitem{liang2020purely}
X.~Liang, H.~Wang, Y.-H. Liu, Z.~Liu, B.~You, Z.~Jing, and W.~Chen, ``Purely
  image-based pose stabilization of nonholonomic mobile robots with a truly
  uncalibrated overhead camera,'' {\em IEEE Transactions on Robotics}, vol.~36,
  no.~3, pp.~724--742, 2020.

\bibitem{automatica13_li}
X.~Li and C.~C. Cheah, ``Global task-space adaptive control of robot,'' {\em
  Automatica}, vol.~49, no.~1, pp.~58--69, 2013.

\bibitem{Hjorth_22}
S.~Hjorth and D.~Chrysostomou, ``Human--robot collaboration in industrial
  environments: A literature review on non-destructive disassembly,'' {\em
  Robotics and Computer-Integrated Manufacturing}, vol.~73, p.~102208, 2022.

\bibitem{Inkulu_21}
A.~K. Inkulu, M.~R. Bahubalendruni, A.~Dara, and K.~SankaranarayanaSamy,
  ``Challenges and opportunities in human robot collaboration context of
  industry 4.0-a state of the art review,'' {\em Industrial Robot: the
  international journal of robotics research and application}, 2021.

\bibitem{simoes_22}
A.~C. Sim{\~o}es, A.~Pinto, J.~Santos, S.~Pinheiro, and D.~Romero, ``Designing
  human-robot collaboration (hrc) workspaces in industrial settings: A
  systematic literature review,'' {\em Journal of Manufacturing Systems},
  vol.~62, pp.~28--43, 2022.

\bibitem{Kanazawa_21}
A.~Kanazawa, J.~Kinugawa, and K.~Kosuge, ``Motion planning for human--robot
  collaboration using an objective-switching strategy,'' {\em IEEE Transactions
  on Human-Machine Systems}, vol.~51, no.~6, pp.~590--600, 2021.

\bibitem{Palleschi_21}
A.~Palleschi, M.~Hamad, S.~Abdolshah, M.~Garabini, S.~Haddadin, and
  L.~Pallottino, ``Fast and safe trajectory planning: Solving the cobot
  performance/safety trade-off in human-robot shared environments,'' {\em IEEE
  Robotics and Automation Letters}, vol.~6, no.~3, pp.~5445--5452, 2021.

\bibitem{noohi_16}
E.~Noohi, M.~{\v{Z}}efran, and J.~L. Patton, ``A model for human--human
  collaborative object manipulation and its application to human--robot
  interaction,'' {\em IEEE transactions on robotics}, vol.~32, no.~4,
  pp.~880--896, 2016.

\bibitem{wang_18}
W.~Wang, R.~Li, Z.~M. Diekel, Y.~Chen, Z.~Zhang, and Y.~Jia, ``Controlling
  object hand-over in human--robot collaboration via natural wearable
  sensing,'' {\em IEEE Transactions on Human-Machine Systems}, vol.~49, no.~1,
  pp.~59--71, 2018.

\bibitem{makrini_17}
I.~El~Makrini, K.~Merckaert, D.~Lefeber, and B.~Vanderborght, ``Design of a
  collaborative architecture for human-robot assembly tasks.,'' in {\em IROS},
  pp.~1624--1629, 2017.

\bibitem{hoffman_19}
G.~Hoffman, ``Evaluating fluency in human--robot collaboration,'' {\em IEEE
  Transactions on Human-Machine Systems}, vol.~49, no.~3, pp.~209--218, 2019.

\bibitem{bogue_16}
R.~Bogue, ``Europe continues to lead the way in the collaborative robot
  business,'' {\em Industrial Robot: An International Journal}, 2016.

\bibitem{argall2009survey}
B.~D. Argall, S.~Chernova, M.~Veloso, and B.~Browning, ``A survey of robot
  learning from demonstration,'' {\em Robotics and autonomous systems},
  vol.~57, no.~5, pp.~469--483, 2009.

\bibitem{ravichandar2020recent}
H.~Ravichandar, A.~S. Polydoros, S.~Chernova, and A.~Billard, ``Recent advances
  in robot learning from demonstration,'' {\em Annual Review of Control,
  Robotics, and Autonomous Systems}, vol.~3, pp.~297--330, 2020.

\bibitem{saveriano2021dynamic}
M.~Saveriano, F.~J. Abu-Dakka, A.~Kramberger, and L.~Peternel, ``Dynamic
  movement primitives in robotics: A tutorial survey,'' {\em arXiv preprint
  arXiv:2102.03861}, 2021.

\bibitem{ude2014orientation}
A.~Ude, B.~Nemec, T.~Petri{\'c}, and J.~Morimoto, ``Orientation in cartesian
  space dynamic movement primitives,'' in {\em 2014 IEEE International
  Conference on Robotics and Automation (ICRA)}, pp.~2997--3004, IEEE, 2014.

\bibitem{gams2014coupling}
A.~Gams, B.~Nemec, A.~J. Ijspeert, and A.~Ude, ``Coupling movement primitives:
  Interaction with the environment and bimanual tasks,'' {\em IEEE Transactions
  on Robotics}, vol.~30, no.~4, pp.~816--830, 2014.

\bibitem{zhou2019learning}
Y.~Zhou, J.~Gao, and T.~Asfour, ``Learning via-point movement primitives with
  inter-and extrapolation capabilities,'' in {\em 2019 IEEE/RSJ International
  Conference on Intelligent Robots and Systems (IROS)}, pp.~4301--4308, IEEE,
  2019.

\bibitem{beik2020model}
H.~Beik-Mohammadi, M.~Kerzel, B.~Pleintinger, T.~Hulin, P.~Reisich, A.~Schmidt,
  A.~Pereira, S.~Wermter, and N.~Y. Lii, ``Model mediated teleoperation with a
  hand-arm exoskeleton in long time delays using reinforcement learning,'' in
  {\em 2020 29th IEEE International Conference on Robot and Human Interactive
  Communication (RO-MAN)}, pp.~713--720, IEEE, 2020.

\bibitem{peternel2018robotic}
L.~Peternel, T.~Petri{\v{c}}, and J.~Babi{\v{c}}, ``Robotic assembly solution
  by human-in-the-loop teaching method based on real-time stiffness
  modulation,'' {\em Autonomous Robots}, vol.~42, no.~1, pp.~1--17, 2018.

\bibitem{sensoryfeedbackbook}
C.~C. Cheah and X.~Li, {\em Task-space sensory feedback control of robot
  manipulators}, vol.~73.
\newblock Springer, 2015.

\bibitem{tro14_sadeghian}
H.~Sadeghian, L.~Villani, M.~Keshmiri, and B.~Siciliano, ``Task-space control
  of robot manipulators with null-space compliance,'' {\em IEEE Transactions on
  Robotics}, vol.~30, no.~2, pp.~493--506, 2013.

\bibitem{slotine}
J.-J.~E. Slotine, W.~Li, {\em et~al.}, {\em Applied nonlinear control},
  vol.~199.
\newblock Prentice hall Englewood Cliffs, NJ, 1991.

\bibitem{hololens2}
Microsoft, ``Microsoft hololens2.''
  https://www.microsoft.com/en-us/hololens/hardware, 2020.
\newblock Accessed: 2022-06-01.

\bibitem{zhang2020modular}
K.~Zhang, M.~Sharma, J.~Liang, and O.~Kroemer, ``A modular robotic arm control
  stack for research: Franka-interface and frankapy,'' {\em arXiv preprint
  arXiv:2011.02398}, 2020.

\bibitem{HeLiu2021}
Y.~He and S.~Liu, ``Analytical inverse kinematics for {F}ranka {E}mika {P}anda
  -- a geometrical solver for 7-{DOF} manipulators with unconventional
  design,'' in {\em 2021 9th International Conference on Control, Mechatronics
  and Automation (ICCMA2021)}, {IEEE}, Nov. 2021.

\end{thebibliography}

\end{document}